\documentclass{article}
\usepackage{pdf14}

\usepackage[final]{neurips_2025}

\usepackage[utf8]{inputenc} % allow utf-8 input
\usepackage[T1]{fontenc}    % use 8-bit T1 fonts
\usepackage{hyperref}       % hyperlinks
\usepackage{url}            % simple URL typesetting
\usepackage{booktabs}       % professional-quality tables
\usepackage{amsfonts}       % blackboard math symbols
\usepackage{nicefrac}       % compact symbols for 1/2, etc.
\usepackage{microtype}      % microtypography
\usepackage{xcolor}         % colors

\usepackage{microtype}
\usepackage{graphicx}
\usepackage{subcaption}
\usepackage{multirow}
\usepackage{makecell}
\usepackage{enumitem}
\usepackage{colortbl}

\usepackage{amsmath}
\usepackage{amssymb}
\usepackage{mathtools}
\usepackage{amsthm}
\usepackage{bbm}
\usepackage{float}
\usepackage{wrapfig}

\theoremstyle{plain}
\newtheorem{theorem}{Theorem}
\newtheorem{proposition}[theorem]{Proposition}

\theoremstyle{definition}
\newtheorem{definition}[theorem]{Definition}

\theoremstyle{remark}

\DeclareMathOperator{\Er}{Er}

\DeclareMathOperator{\PMI}{PMI}
\DeclareMathOperator{\TTE}{TTE}
\DeclareMathOperator{\censoring}{censoring}

\title{The Boundaries of Fair AI in Medical Image Prognosis: A Causal Perspective}

\author{%
  Thai-Hoang Pham\textsuperscript{1,2}, Jiayuan Chen\textsuperscript{1,2}\thanks{Equal contribution authors.}, Seungyeon Lee\textsuperscript{1,2}\footnotemark[1], Yuanlong Wang\textsuperscript{1,2}\footnotemark[1],
  \\
  \textbf{Sayoko Moroi\textsuperscript{3}, Xueru Zhang\textsuperscript{1}, Ping Zhang\textsuperscript{1,2}}\thanks{Corresponding author.} \\
  \textsuperscript{1}Department of Computer Science and Engineering, The Ohio State University \\
  \textsuperscript{2}Department of Biomedical Informatics, The Ohio State University \\
  \textsuperscript{3}Department of Ophthalmology and Visual Sciences, The Ohio State University \\
  \texttt{\{pham.375,chen.12930,lee.10029,wang.16050\}@osu.edu} \\
  \texttt{\{moroi.4,zhang.12807,zhang.10631\}@osu.edu}
}

\begin{document}

\maketitle

% Causal in title
\setcounter{footnote}{0}
\begin{abstract}
As machine learning (ML) algorithms are increasingly used in medical image analysis, concerns have emerged about their potential biases against certain social groups. Although many approaches have been proposed to ensure the fairness of ML models, most existing works focus only on medical image diagnosis tasks, such as image classification and segmentation, and overlooked prognosis scenarios, which involve predicting the likely outcome or progression of a medical condition over time. To address this gap, we introduce FairTTE, the first comprehensive framework for assessing fairness in time-to-event (TTE) prediction in medical imaging. FairTTE encompasses a diverse range of imaging modalities and TTE outcomes, integrating cutting-edge TTE prediction and fairness algorithms to enable systematic and fine-grained analysis of fairness in medical image prognosis. Leveraging causal analysis techniques, FairTTE uncovers and quantifies distinct sources of bias embedded within medical imaging datasets. Our large-scale evaluation reveals that bias is pervasive across different imaging modalities and that current fairness methods offer limited mitigation. We further demonstrate a strong association between underlying bias sources and model disparities, emphasizing the need for holistic approaches that target all forms of bias. Notably, we find that fairness becomes increasingly difficult to maintain under distribution shifts, underscoring the limitations of existing solutions and the pressing need for more robust, equitable prognostic models.
\end{abstract}
\addtocontents{toc}{\protect\setcounter{tocdepth}{0}}
\section{Introduction}
Machine learning (ML) algorithms trained on real-world medical images may inherently exhibit bias, leading to discrimination against certain social groups~\cite{seyyed2021underdiagnosis,larrazabal2020gender}. This is especially concerning in the medical field, where biased algorithms can result in inequitable treatment recommendations, misdiagnoses, or unequal access to care~\cite{panch2019artificial}. Therefore, ensuring the fairness of ML models and identifying hidden biases within medical images is critical for advancing health equity~\cite{ricci2022addressing}. 

Existing research on fairness in ML and medical image analysis has primarily focused on medical image diagnosis tasks, such as image classification and segmentation~\cite{zong2023medfair,tian2024fairseg}. These diagnostic tasks typically aim to determine the presence or absence of a condition. However, prognosis scenarios, which involve predicting the likely outcome or progression of a medical condition over time, have been largely overlooked. In ML, prognosis is framed as time-to-event (TTE) prediction or survival analysis, where the goal is to predict the time until a critical event. Unlike classification, TTE prediction provides a richer and more dynamic approach to modeling medical outcomes by accounting not only for the presence or absence of a condition but also for the timing and progression of key health events. This makes it especially valuable for predicting long-term outcomes such as survival or disease recurrence~\cite{holste2024harnessing}. Despite its importance, there remains a significant gap in the literature regarding fairness in TTE prediction, particularly in the context of medical imaging (see Appendix~\ref{sec:a}).

\begin{figure*}[t]
  \begin{center}
    \includegraphics[width=\textwidth]{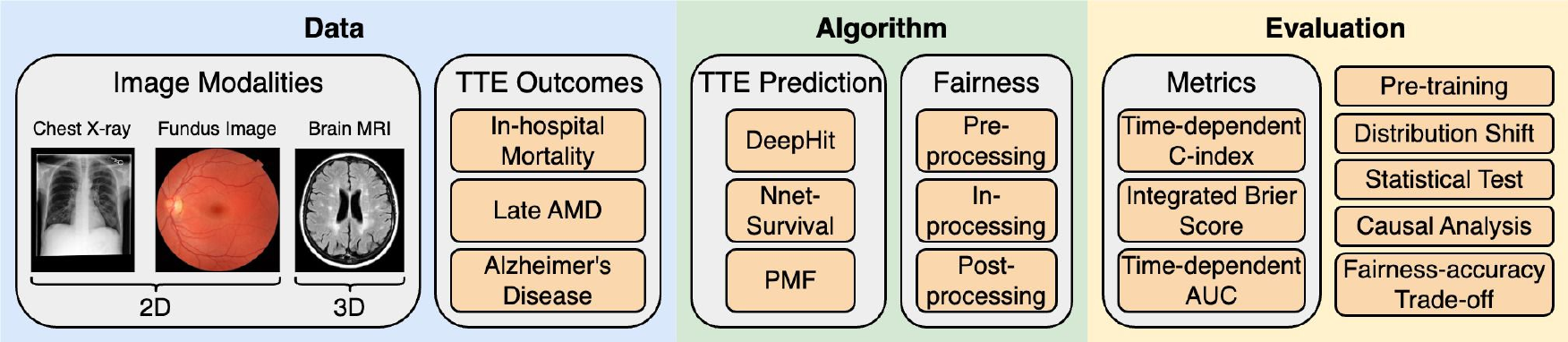}
  \end{center}
  % \vskip -0.1in
  \vspace{-0.15cm}
  \caption{An overview of the FairTTE, a unified framework designed to investigate fairness in TTE prediction for medical image analysis.\label{fig:1}}
  % \vskip -0.1in
\vspace{-0.5cm}
\end{figure*}

Developing fairness methods for TTE prediction in medical imaging presents several unique challenges compared to tasks like image classification or segmentation. First, there is a lack of public datasets that pair medical images with TTE outcome information. While many public datasets include sensitive attributes, they primarily focus on diagnostic tasks and do not provide the temporal data necessary for training TTE prediction models. Second, there is limited understanding of how biases in medical images specifically affect the fairness of TTE prediction models, complicating the development of fair algorithms. Third, the absence of a universally accepted fairness metric for TTE prediction poses a challenge in assessing and comparing fairness across studies~\cite{keya2021equitable,rahman2022fair,zhang2022longitudinal}. Existing fairness metrics are often not well-suited to the complexities of medical applications (see Section~\ref{sec:framework-1}). Furthermore, varying experimental designs and evaluation approaches in the literature hinder direct comparisons between fairness methods, leading to fragmentation in the field and impeding the establishment of best practices for fair TTE prediction in medical imaging.

To address the challenges of understanding fairness in TTE prediction for medical imaging, we introduce FairTTE, the first comprehensive framework specifically designed for this task (Figure~\ref{fig:1}). Our approach begins by applying causal reasoning to develop a general framework for understanding and quantifying bias in TTE prediction. This framework decomposes the data likelihood into distinct components, enabling a fine-grained analysis of various sources of bias, including disparities in image features, censoring rates, TTE labels, mutual information between images and TTE, and mutual information between images and censoring indicators. This principled decomposition allows us to explain why standard TTE prediction methods and fairness algorithms frequently fail to mitigate bias in practice (see Section~\ref{sec:framework-2}). Using this framework, we characterize not only the presence but also the type and extent of bias across different datasets, offering a detailed understanding of fairness challenges in medical image prognosis. To complement the framework, we construct a collection of large-scale, publicly available medical image datasets tailored for fair TTE prediction. These datasets span a wide range of imaging modalities, including fundus images, chest X-rays, and brain MRIs, and cover diverse TTE outcomes, such as mortality, late age-related macular degeneration (AMD), and Alzheimer's disease. Finally, leveraging both the datasets and causal framework, we establish a realistic and comprehensive investigation study for fairness in TTE prediction. This investigation incorporates state-of-the-art (SOTA) algorithms across various learning strategies and, to our knowledge, represents the first attempt to provide a fine-grained analysis of fairness in TTE prediction, making a significant contribution to the growing field of fair ML in medical imaging.

Our work also explores fair TTE prediction methods across various settings, considering factors such as pre-training and distribution shifts that may influence model fairness. Building on our causal framework and through the training of over 20,000 models, we derive several key insights to inform future research in this area, as follows:
\begin{itemize}[itemsep=0pt, topsep=0pt, left=0pt]
    \item Bias is prevalent across TTE prediction models trained on different imaging modalities, with consistent performance disparities observed between demographic groups.
    \item All medical image datasets examined in our study exhibit some sources of bias that adversely affect the fairness of TTE prediction models.
    \item Most SOTA fairness methods struggle to consistently mitigate bias—while they can improve fairness in some settings, these gains are often accompanied by reductions in predictive accuracy.
    \item Pre-training improves model accuracy but has minimal impact on fairness.
    \item Different types of distribution shifts affect fair TTE prediction in distinct ways, underscoring the growing difficulty of maintaining both fairness and utility under realistic clinical scenarios.
\end{itemize}
\section{Unified Framework for Fair TTE Prediction}\label{sec:framework}

\subsection{Fair TTE Prediction Setup}\label{sec:framework-1}

We first introduce the notations used throughout the paper and then formulate the fair TTE prediction task. In this study, we focus on analyzing methods designed for achieving group fairness in right-censored TTE prediction.

\textbf{Notations.} A TTE dataset contains observations for each individual along with their corresponding (right-censored) TTE outcomes. Specifically, each individual's data is represented by a tuple of random variables (RVs) $\left(X, Y, \Delta, A\right)$, where $X$ denotes a set of features, $Y$ represents an observed time, $\Delta$ is an event indicator, and $A$ is a sensitive attribute for the individual. If $\Delta = 1$ (indicating that the event has occurred), then $Y$ represents a true survival time; otherwise, $Y$ is a censoring time. We denote $T$ as a true (possibly unobserved) survival time associated with $X$, and $C$ as a true (possibly unobserved) censoring time associated with $X$. We use lowercase letters $x$, $y$, $\delta$, $a$, $t$, $c$ to denote the respective realizations of these random variables. It is important to note that for each individual, we do not observe both $T$ and $C$; instead, we observe exactly one of them. More precisely, $Y = \min\left\{T, C\right\}$ and $\Delta = \mathbbm{1}\left\{T \leq C\right\}$, where $\mathbbm{1}$ is an indicator function. Let $S(t'|x) = 1 - \int_{0}^{t'}P(t|x)dt$ represent the survival function at time $t$ given the feature $x$. In fair TTE prediction, our goal is to accurately predict $S(t'|x)$ while adhering to fairness constraints.

\textbf{Performance Metrics for TTE Prediction.} Consider a TTE prediction model $h: \mathcal{X} \rightarrow \mathcal{T}$ where $\mathcal{X}$ and $\mathcal{T}$ are the spaces of $x$ and $t$, we denote the performance metric for this model on the dataset $D = \left\{ X_i, Y_i, \Delta_i, A_i \right\}_{i=1}^{|D|}$ as $\Er(f,h,D)$ where $f$ is the ground-truth labeling function (i.e., $f = P(t|x)$). Note that this notation is sufficiently flexible to encompass censoring data and different metrics in TTE prediction. Furthermore, while prediction error (smaller values indicate better performance, \textit{e.g.}, Brier score~\cite{graf1999assessment}) is used as the metric in our theoretical analysis, prediction accuracy (larger values indicate better performance, \textit{e.g.}, concordance index~\cite{harrell1982evaluating}) can also be applied.

\textbf{Fairness Metrics for TTE Prediction.}
Various fairness metrics have been proposed recently for TTE prediction, which can be roughly classified into three categories based on their objectives: (i) ensuring similar predicted TTE outcomes for similar data points~\cite{keya2021equitable,rahman2022fair,zhang2023individual,zhang2023censored,wang2024individual}, (ii) ensuring similar predicted outcomes for data points from different groups~\cite{keya2021equitable,rahman2022fair,zhao2023fairness}, and (iii) ensuring similar predictive performance across different groups~\cite{do2023fair,hu2024fairness,zhang2021fair,zhang2022longitudinal}. Notably, some of these metrics are less applicable in the context of medical imaging. For instance, metrics in the first category require a well-defined similarity measure between data points, which is difficult to establish for medical images. Metrics in the second category may also be inappropriate when sensitive attributes are strong risk factors for the TTE outcome. For example, age is often a key predictor of various TTE outcomes, such as mortality. In this case, asking for similar predicted survival times for young and elderly individuals would be nonsensical. 

Therefore, in this study, we focus on fairness metrics from the third category, which aim to ensure that the model maintains equal predictive performance across different groups. Specifically, given a performance metric $\Er$ for TTE prediction task, we define fairness metric $\mathcal{F}_{\Er}$ as follow:
\begin{align}\label{eq:1}
    \mathcal{F}_{\Er}(h) = \max_{a, a' \in \mathcal{A}} \left | \Er\left(f_a,h,D_a\right) - \Er\left(f_{a'},h,D_{a'}\right) \right |
\end{align}
where $\mathcal{A}$ is the set of groups considered in TTE prediction task, $D_a = \left\{ X_i, Y_i, \Delta_i, A_i | A_i=a\right\}_{i=1}^{|D_a|}$ and $D_{a'} = \left\{ X_i, Y_i, \Delta_i, A_i | A_i=a'\right\}_{i=1}^{|D_{a'}|}$ are the subsets containing data from groups $a$ and $a'$, and $f_a = P(t|x,a)$ and $f_{a'} = P(t|x,a')$ are the ground-truth labeling functions for $D_a$ and $D_{a'}$, respectively. If the labeling function $f$ is independent of the sensitive attribute, then $f=f_a=f_{a'}$. Note that the fairness metric in Eq.~\eqref{eq:1} is defined as the maximum performance gap between any two groups, and as such, it can be applied to any performance metric used in TTE prediction.

\subsection{Causal Structure for Fair TTE Prediction}\label{sec:framework-2}
% \begin{wrapfigure}[17]{r}{0.4\textwidth}\label{fig:example}
%   \begin{center}
%   \vspace{-0.8cm}
%     \includegraphics[width=0.37\textwidth]{example.pdf}
%   \end{center}
%   \vspace{-1.2em}
%   \caption{An example of domain generalization in healthcare: (fair) ML model trained with patient data in CA, NY, etc., can be deployed in other states by maintaining high accuracy/fairness. }
% \end{wrapfigure}
To conduct a fine-grained analysis to understand fairness in TTE prediction, we leverage the structural causal model (SCM)~\cite{pearl2000models} to represent the data generation process underlying TTE data. Specifically, we examine how the sensitive attribute $A$ influences the TTE prediction model, which aims to approximate the data likelihood $P(t|x)$ using the SCM framework. To construct the causal graph, we introduce an unobserved underlying health condition of the patient, denoted as $Z$. In this context, the patient’s feature set $X$ can be viewed as noisy and partially observed information derived from $Z$. 
\begin{wrapfigure}[14]{r}{0.5\textwidth}
\vspace{-0.5cm}
  \begin{center}
    \includegraphics[width=0.5\textwidth]{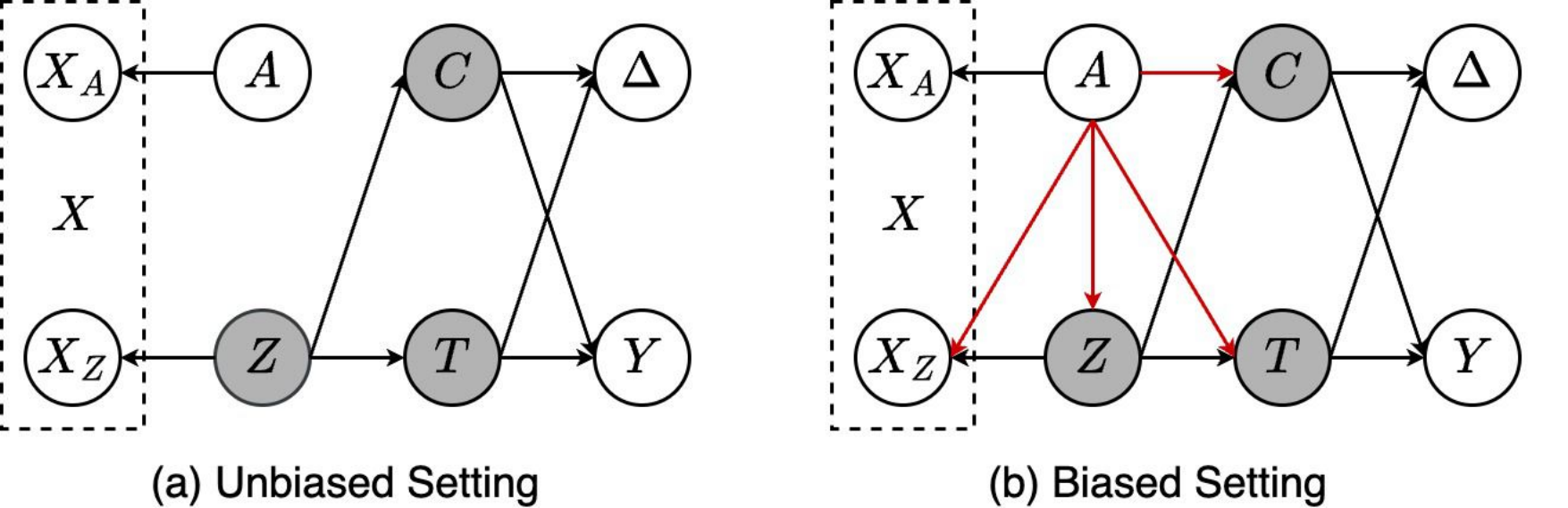}
  \end{center}
  % \vskip -0.1in
  \caption{Causal structure in TTE prediction. Gray circles represent unobserved RVs. (a) Unbiased setting, where the sensitive attribute $A$ affects only $X_A$. (b) Biased setting, where the sensitive attribute $A$ may be correlated (red arrow) with other RVs in causal graph.\label{fig:2}}
  % \vskip -0.2in
\end{wrapfigure}
Following the approach in~\cite{jiang2022invariant,wang2021self}, we partition $X$ into two components: $X_Z$, representing target-related features directly influenced by $Z$, and $X_A$, representing features related to the sensitive attribute, directly influenced by $A$. By construction, $X_A$ encodes sensitive information as it is predictive of the sensitive attribute $A$. Using these definitions, we construct causal graphs for TTE data under two settings: unbiased and biased scenarios, illustrated in Figure~\ref{fig:2}. In the unbiased scenario (Figure~\ref{fig:2}a), the sensitive attribute $A$ is irrelevant to the TTE outcome. It only influences $X_A$, and not any other variable in the graph. Therefore, capturing the invariant feature $X_Z$ across groups is sufficient for learning a fair model. Specifically, we have, $P(t|x_z) = P(t|x_z, a), \; \forall a \in \mathcal{A}$. In contrast, in the biased scenario (Figure~\ref{fig:2}b), the sensitive attribute $A$ influences additional variables in the causal graph beyond $X_A$, leading to dependency between $A$ and TTE outcome. This results in the conditional distribution of the outcome given $X_Z$	varies across groups $P(t|x_z, a) \neq P(t|x_z, a')$, $a,a' \in \mathcal{A}$ and/or  the distribution of $X_Z$ differs across groups $P(x_z|a) \neq P(x_z|a')$. These discrepancies reflect the presence of bias in the data and challenge the assumption that learning invariant features alone is sufficient for fair TTE prediction.\footnote{We acknowledge the possibility of unobserved RVs that may influence both sensitive attribute $A$ and other RVs (i.e., $X,Z,T,C$) in causal graph. However, since our primary objective is to characterize disparities in group-specific data distributions, we omit unobserved RVs from the causal graph for simplicity (see Appendix~\ref{sec:g1}).} 

\begin{wrapfigure}[16]{r}{0.5\textwidth}
\vspace{-0.55cm}
  \begin{center}
    \includegraphics[width=0.5\textwidth]{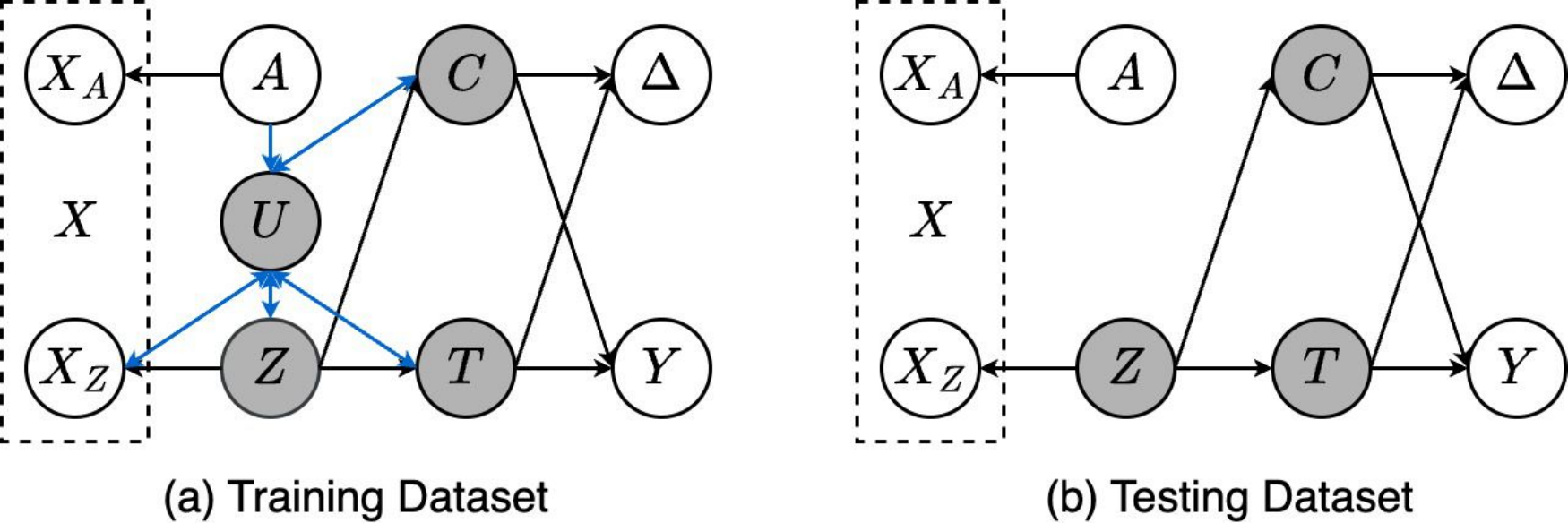}
  \end{center}
  % \vskip -0.1in
  \vspace{-0.1cm}
  \caption{Causal structure in TTE prediction under distribution shift. We illustrate a scenario where an unfair causal pathways (blue arrows), induced by unobserved RV $U$, are present in train data (a) but absent in test data (b), leading to distribution shift. Bidirectional arrows indicate that the causal direction may vary depending on the specific context. Fair causal pathways (appear in both train and test data) may exists but are omitted for simplicity.\label{fig:3}}
  % \vskip -0.2in
\end{wrapfigure}
% We note that, the presence of data distribution disparity across groups in the causal graph does not necessarily imply that the sensitive information captured during model training is spurious or inappropriate in all contexts. In some cases, legitimate biological differences between groups may exist, making the sensitive attribute relevant for disease prediction. In such scenarios, incorporating group-specific information can be helpful, as it enables the model to learn distinct disease mechanisms that reflect true underlying variation. 
It is important to note that disparities in data distributions across groups, as reflected in the causal graph, do not necessarily indicate that the sensitive information captured during model training is spurious or inappropriate in every context. In some cases, genuine biological differences between groups may exist, making the sensitive attribute relevant for disease prediction. In such situations, incorporating group-specific information can be beneficial, as it allows the model to capture distinct disease mechanisms that reflect true underlying biological variation.

% However, observed biases in the training data may stem from spurious correlations introduced by historical inequities in healthcare delivery (blue arrows in Figure~\ref{fig:3}a). In these settings, models risk perpetuating unfairness by reflecting these spurious associations. From this perspective, fairness and distribution shift are related: achieving fairness requires models that generalize to the test distribution where unfair causal pathways are absent. (Figure~\ref{fig:3}b)~\cite{kim2019learning,wang2020towards,tartaglione2021end}.\footnote{Real-world medical examples of fair and unfair causal pathways are provided in Appendix~\ref{sec:g2}.}

However, observed biases in the training data may arise from spurious correlations that reflect historical inequities in healthcare access, diagnosis, and treatment practices (illustrated by the blue arrows in Figure~\ref{fig:3}a). When models are trained on such data, they risk encoding and amplifying these biases, thereby perpetuating unfairness in clinical decision-making. From this perspective, fairness and distribution shift are intrinsically connected: achieving fairness requires models that can generalize to a test distribution in which unfair causal pathways are absent and only fair causal pathways remain (Figure~\ref{fig:3}b)~\cite{kim2019learning,wang2020towards,tartaglione2021end}.\footnote{Real-world medical examples of fair and unfair causal pathways are provided in Appendix~\ref{sec:g2}.}

% \vspace{-0.3cm}
\paragraph{Group-specific Distribution Disparity and Fairness.}
As shown in Figure~\ref{fig:2}b, the effect of $A$ on other RVs can induce distribution disparity across groups. We now formalize how it affect fairness in Theorem~\ref{thm:1} below.

\begin{theorem}\label{thm:1}
Given a performance metric $\Er$ satisfying triangle inequality and symmetry properties, i.e., $\left| \Er(h,h',D) - \Er(h,h'',D) \right| \leq \Er(h',h'',D)$ and $\Er(h,h',D) = \Er(h',h,D)$, then we have:
\begin{align}
    \mathcal{F}_{\Er} (h) \leq \max_{a,a' \in \mathcal{A}}\left( \eta(\mathcal{H},f_a,f_{a'}) + \mathcal{D} (\mathcal{H},D_a,D_{a'})\right), \;\;\;\;  \forall h \in \mathcal{H} \label{eq:4}
\end{align}
% \vspace{-0.2cm}
with
\vspace{-0.1cm}
\begin{align*}
\eta(\mathcal{H}, f_a,f_{a'}) &= \min_{h' \in \mathcal{H}} \left(\Er(f_a,h',D_a) + \Er(f_{a'},h',D_{a'})\right);\\
\mathcal{D}(\mathcal{H},D_a,D_{a'}) &= \max_{h', h'' \in \mathcal{H}}\left| \Er(h',h'',D_a) - \Er(h',h'',D_{a'}) \right|
\end{align*}
\end{theorem}
where $\mathcal{H}$ represents the hypothesis class of the TTE prediction model $h$, %$\eta(\mathcal{H}, f_a,f_{a'}) = \min_{h' \in \mathcal{H}} \Er(f_a,h',D_a) + \Er(f_{a'},h',D_{a'})$ 
$\eta(\mathcal{H}, f_a,f_{a'}) $ denotes the minimum joint prediction error on $D_a$ and $D_{a'}$, and $\mathcal{D}(\mathcal{H},D_a,D_{a'})$ %$\mathcal{D}(\mathcal{H},D_a,D_{a'}) = \max_{h', h'' \in \mathcal{H}}\left| \Er(h',h'',D_a) + \Er(h',h'',D_{a'}) \right|$ 
is the largest distance between two groups $a$ and $a'$, with respect to the hypothesis class $\mathcal{H}$. Note that the requirements on symmetric and triangle inequality properties are relatively mild and can be met by many performance metrics commonly used in practice (see Appendix~\ref{sec:b2}).

\textbf{Discussion of Theorem~\ref{thm:1}.} The term $\mathcal{D}(\mathcal{H}, D_a, D_{a'})$ aligns with the concept of subgroup separability—the ability to predict group membership from image features—in the group fairness literature~\cite{jones2023role}. Low subgroup separability implies that the image feature distributions are similar across groups, which corresponds to a small value of $\mathcal{D}(\mathcal{H}, D_a, D_{a'})$. Prior work has shown that unfairness tends to be less severe in datasets with low subgroup separability~\cite{jones2023role,yang2024limits}. Our results extend these findings by demonstrating that this relationship also holds in the context of TTE prediction, while further revealing that model fairness is influenced by additional factors beyond subgroup separability. We observe that while it is often possible to reduce $\mathcal{D}(\mathcal{H}, D_a, D_{a'})$ during training—for example, by learning fair representations across groups—minimizing the term $\eta(\mathcal{H}, f_a, f_{a'})$ may remain challenging. In particular, when the labeling functions $f_a$ and $f_{a'}$ differ significantly across groups (Figure~\ref{fig:2}b), even the optimal choice of hypothesis class $\mathcal{H}$ may not ensure a small $\eta$. In such cases, the upper bound on fairness error becomes large, implying that $\mathcal{F}_{\Er}(h)$ may also be large. Conversely, when $f_a$ and $f_{a'}$ are similar—as in Figure~\ref{fig:2}a or when the only causal pathway from $A$ is to $X$ in Figure~\ref{fig:2}b—the upper bound becomes small, indicating that $\mathcal{F}_{\Er}(h)$ can also be small. In these settings, fairness can be achieved by learning invariant representations across groups. We formally state this observation as follows.
\begin{proposition}\label{thm:2}
Let $g: \mathcal{X} \rightarrow \mathcal{Z}$ be a mapping from the input space $\mathcal{X}$ to a representation space $\mathcal{Z}$. Assume the following conditions hold:
\begin{itemize}[topsep=0pt, left=8pt]
    \item[(i)] Data distribution disparity across groups is a covariate shift, i.e., $P(t | x, a) = P(t | x), \forall a \in \mathcal{A}$.
    \item[(ii)] Representation $\mathcal{Z}$ is sufficient for an arbitrary group $a \in \mathcal{A}$, i.e., $I_a(Z, T) = I_a(X, T)$, where $ I_a(\cdot, \cdot) $ denotes the mutual information computed over the distribution $D_a$.
\end{itemize}
Then, distribution shift across groups w.r.t. representation $\mathcal{Z}$ is also a covariate shift, that is,
\[
P(t | z, a) = P(t | z), \quad \forall a \in \mathcal{A}.
\]
\end{proposition}
\textbf{Discussion of Proposition~\ref{thm:2}.} This proposition suggests that under covariate shift $(i)$, fair TTE prediction can be achieved by learning a fair representation $Z$ across groups (i.e., $P(z | a) = P(z), \forall a \in \mathcal{A}$). More clearly, $P(z | a) = P(z), \forall a \in \mathcal{A}$ guarantees small $\mathcal{D}(\mathcal{H},D_a,D_{a'})$ while $P(t | z, a) = P(t | z), \forall a \in \mathcal{A}$ indicates small $\eta(\mathcal{H}, f_a,f_{a'})$. We note that the sufficiency condition imposed on $Z$ $(ii)$ is practical: it only needs to hold for one group, and during training we have access to labeled data. Moreover, the dimension of $T$ is often smaller than that of $Z$, making the sufficiency assumption more attainable in practice.

To further investigate the influence of sensitive attributes on the fairness of TTE prediction models, we utilize our proposed causal framework to decompose the group-specific labeling functions. Following Bayes' theorem, we rewrite the labeling function $f_a$ w.r.t. each group $a$ as:
\begin{align}
    P(t|x,a) &= \frac{P(x|t,a)P(t|a)}{P(x|a)} = \frac{P(x_a|x_z,t,a)}{P(x_a|x_z,a)} \cdot \frac{P(x_z|t,a)}{P(x_z|a)} \cdot P(t|a) = \frac{P(x_z|t,a)}{P(x_z|a)} \cdot P(t|a) \label{eq:2}
\end{align}
Note that in practical TTE prediction scenario, we observe $Y$ and $\Delta$ rather than $T$. Therefore, the TTE prediction model estimates $P(y,\delta|x)$ instead. Replacing $t$ with $y$ and $\delta$ in Eq.~\eqref{eq:2}, we obtain:
\begin{align}
    &P(y, \delta|x,a) = \frac{P(x_z|y,\delta,a)}{P(x_z|a)} \cdot P(y,\delta|a) = \underbrace{\frac{P(x_z|y,\delta,a)}{P(x_z|\delta,a)}}_{\PMI(x_z,y)} \cdot \underbrace{\frac{P(x_z|\delta,a)}{P(x_z|a)}}_{\PMI(x_z,\delta)} \cdot \underbrace{P(y|\delta,a)}_{\TTE} \cdot \underbrace{P(\delta|a)}_{\censoring} \label{eq:3}
\end{align}
where the first and second terms represent the pointwise mutual information (PMI) between $X_Z$ and $Y$ (conditioned on $\Delta$), and between $X_Z$ and $\Delta$, respectively. The third term corresponds to the TTE distribution, while the fourth term accounts for the censoring rate. As shown in Figure~\ref{fig:2}b, when $A$ is correlated with other random variables in the causal graph, it alters these four terms, resulting in $P(y, \delta|x,a) \neq P(y, \delta|x,a')$.

\textbf{Sources of Bias\footnote{In our context, sources of bias refer to disparities in the data probability distributions across groups, which can be quantified from observed data. However, identifying causal pathways between the sensitive attribute and other variables in the causal graph (Figure~\ref{fig:2}b) remains challenging and requires deep clinical insight.} in Fair TTE Prediction.}
Based on Theorem~\ref{thm:1} and the decomposition formula in Eq.~\eqref{eq:3}, we can identify five primary sources of bias across groups in fair TTE prediction: 1) disparity in image feature distributions, 2) disparity in mutual information between $X_Z$ and $Y$, 3) disparity in mutual information between $X_Z$ and $\Delta$, 4) disparity in the TTE distributions, and 5) disparity in the censoring rates. In practice, we observe that TTE datasets often contain multiple sources of bias rather than just one. These five cases represent the fundamental sources of bias and are crucial for understanding the complex biases present in real-world TTE data.%\footnote{In our context, sources of bias refer to disparities in the data probability distributions across groups}
% \vspace{-1cm}
\section{Experimental Setup for Fair TTE Prediction}\label{sec:setup}

We propose FairTTE, a reproducible and user-friendly framework for evaluating fairness algorithms in TTE prediction within medical imaging. Our framework includes large-scale experiments conducted on three real-world medical image datasets, covering diverse imaging modalities and TTE outcomes, with up to three sensitive attributes considered for each dataset. Using these datasets, we evaluate three TTE prediction models and five fairness algorithms in the context of fair TTE prediction. % We provide both the source code and comprehensive documentation, enabling other researchers to easily replicate results and extend the framework to incorporate additional datasets and algorithms.

\textbf{Datasets.} FairTTE includes \textbf{MIMIC-CXR}~\cite{johnson2019mimic} for predicting in-hospital mortality from chest X-ray images, \textbf{ADNI}~\cite{petersen2010alzheimer} for predicting Alzheimer’s disease from brain MRI images, and \textbf{AREDS}~\cite{ferris2005simplified} for predicting late AMD from color fundus images. We selected these datasets for our benchmark based on several key criteria: the availability of temporal information to derive TTE outcomes, the diversity of medical imaging modalities, the presence of various potential sources of bias, the availability of sensitive attributes, and the range of dataset sizes. Table~\ref{tab:a1} provides basic information about these datasets, with additional details on data access and TTE label construction provided in Appendix~\ref{sec:c}.

\textbf{Algorithms.} \underline{TTE prediction models:} We employ \textbf{DeepHit}~\cite{lee2018deephit}, \textbf{Nnet-survival}~\cite{gensheimer2019scalable}, and \textbf{PMF}~\cite{kvamme2021continuous} as the base models for TTE prediction to investigate the fairness issue in medical image prognosis. \underline{Fairness algorithms:} We incorporate five SOTA fairness algorithms and adapt them to the TTE prediction setting to promote equitable predictions. These algorithms span a wide range of learning strategies and are categorized into three main groups: 1) pre-processing: \textbf{subgroup rebalancing (SR)}~\cite{kamiran2012data}, 2) in-processing: \textbf{domain independence (DI)}~\cite{wang2020towards}, \textbf{fair representation learning (FRL)}~\cite{wang2021understanding}, \textbf{distributionally robust optimization (DRO)}~\cite{hu2024fairness}, and post-processing: \textbf{controlling for sensitive attributes (CSA)}~\cite{pope2011implementing}. To the best of our knowledge, this is the first study to comprehensively evaluate such a diverse set of fairness algorithms for TTE prediction in medical imaging. %Furthermore, our benchmark is easily extensible, allowing for the inclusion of new algorithms. 
Detailed descriptions of each algorithm are provided in Appendix~\ref{sec:d}.

% \begin{wrapfigure}[18]{r}{0.5\textwidth}
%   \begin{center}
%     \includegraphics[width=0.5\textwidth]{NeurIPS2025/figs/tte_performance.png}
%   \end{center}
%   \vskip -0.2in
%   \caption{Per-group predictive performance ($C^{td}$) of TTE prediction models across various datasets and sensitive attribute combinations. The visualized performances correspond to the best models determined by model selection conducted on the validation sets. The $95\%$ confidence intervals (CIs) are calculated using bootstrapping over test sets.\label{fig:3}}
%   \vskip -0.15in
% \end{wrapfigure}

\textbf{Evaluation Metrics.} Compared to classification tasks, evaluating TTE prediction, particularly in medical contexts, requires careful consideration. The primary challenge in assessing accuracy in TTE prediction is \textbf{censoring}, which complicates the use of standard evaluation metrics. As a result, no single evaluation metric is universally ideal for all TTE prediction scenarios. %Moreover, many commonly used metrics are known to be \textbf{improper}~\cite{gneiting2007strictly}, meaning the true conditional survival distribution does not necessarily achieve the best possible score. 

To address these challenges, we adopt multiple evaluation metrics to ensure a comprehensive assessment of TTE predictive performance. These metrics can be grouped into two categories: 1) ranking-based metrics which evaluate the ranking of patients in the dataset based on their corresponding survival times: \textbf{time-dependent C-index ($C^{td}$)}~\cite{antolini2005time}, \textbf{time-dependent AUC ($AUC^{td}$)}~\cite{uno2007evaluating} and 2) squared error which measures the error between estimated survival times and ground-truth values: \textbf{Integrated Brier score ($IBS$)}~\cite{graf1999assessment}.
For each performance metric, we consider a corresponding fairness metric. A detailed description of all metrics can be found in Appendix~\ref{sec:e}.

% \begin{wrapfigure}[19]{r}{0.5\textwidth}
%   \begin{center}
%     \includegraphics[width=0.5\textwidth]{NeurIPS2025/figs/pretrain_performance.png}
%   \end{center}
%   \vskip -0.2in
%   \caption{Per-group average performance gap ($\Delta C^{td}$) for TTE prediction models using a pre-training strategy compared to training from scratch across various datasets and sensitive attribute combinations. A positive $\Delta C^{td}$ indicates that the pre-training strategy enhances predictive performance relative to training from scratch.\label{fig:4}}
%   \vskip -0.15in
% \end{wrapfigure}

\textbf{Model Selection.} Prioritizing fairness often involves a trade-off with utility, as the model’s objective shifts from utility to balancing both utility and fairness. Consequently, model selection becomes critical in determining the optimal trade-off. In our setting, we perform a hyperparameter search and select the best fair TTE prediction models based on fairness performance on the validation set, allowing for up to a 5\% reduction in predictive performance compared to the base TTE models.

\textbf{Implementation Details.} We use a 2D EfficientNet~\cite{tan2019efficientnet} backbone for the AREDS and MIMIC-CXR datasets, and a 3D ResNet-18 backbone~\cite{tran2018closer} for the ADNI dataset. These lightweight backbones are chosen to mitigate overfitting. %{\color{green} why are those image encoders in such a medical imaging benchmark paper? Are they representative? Reviewers in healthcare will ask.} 
In addition to training models from scratch, we also explore pre-training using weights from models trained on the ImageNet~\cite{deng2009imagenet} and Kinetics~\cite{kay2017kinetics} datasets. To ensure stability across random initializations, we perform a hyperparameter search with 10 random seeds for each combination of dataset, sensitive attribute, algorithm, and evaluation metric. More implementation details can be found in Appendix~\ref{sec:f} and in our code repository.~\footnote{\url{https://github.com/pth1993/FairTTE}}
\section{Experiment and Result}
\begin{wrapfigure}[20]{r}{0.5\textwidth}
\vspace{-0.8cm}
  \begin{center}
    \includegraphics[width=0.5\textwidth]{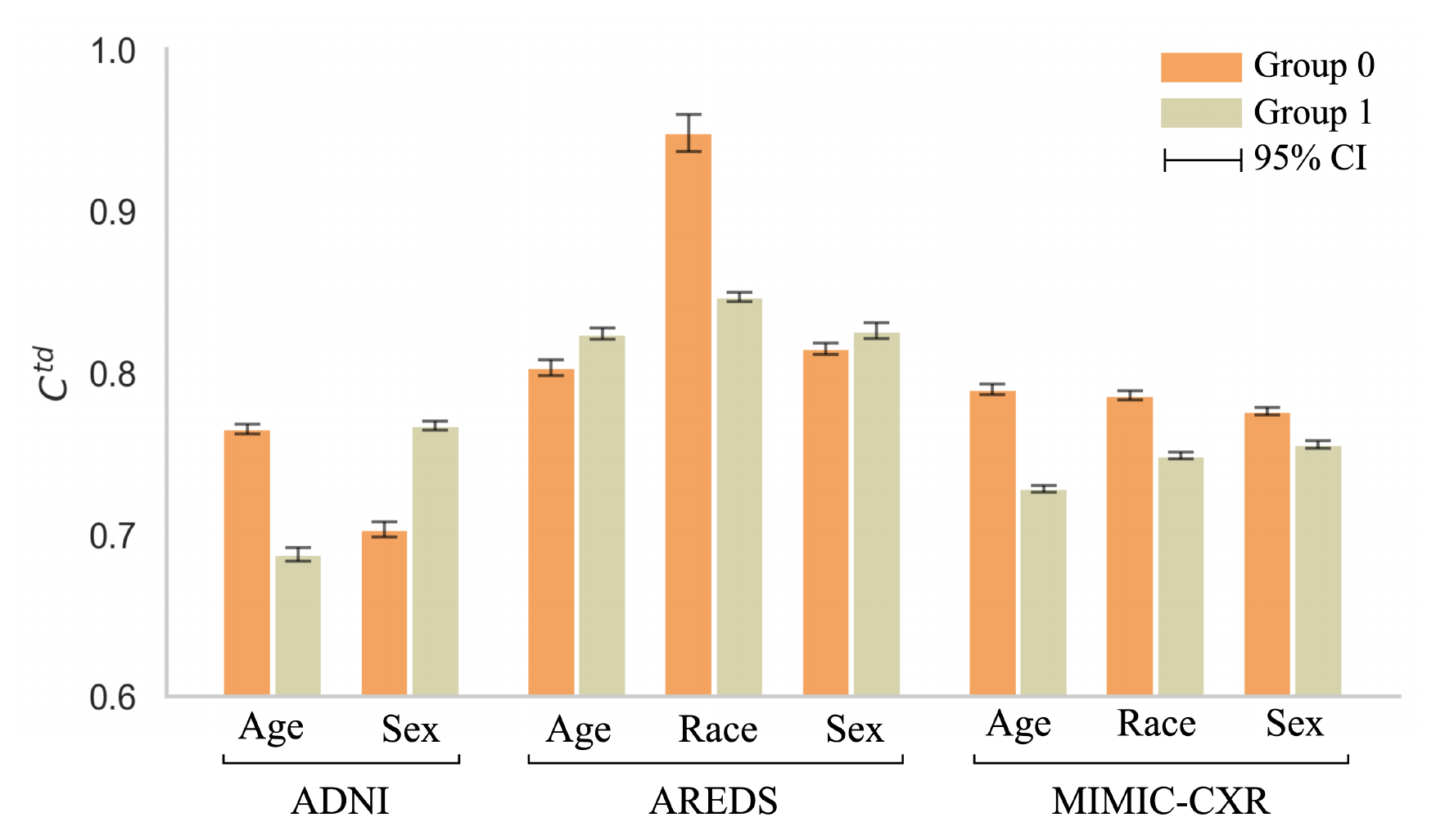}
  \end{center}
% \vspace{-0.3cm}
  \caption{Per-group performance ($C^{td}$) of TTE prediction models across various datasets and sensitive attribute combinations. The visualized performances correspond to the best models determined by model selection conducted on validation sets. The $95\%$ confidence intervals (CIs) are calculated using bootstrapping over test sets. Definitions of groups 0 and 1 are provided in Appendix~\ref{sec:f2}.\label{fig:4}}
  % \vskip -0.15in
\end{wrapfigure}
\textbf{Bias Across TTE Prediction Models in Diverse Imaging Modalities and Outcomes.} 
We first train TTE prediction models, including DeepHit, Nnet-survival, and PMF, on various datasets and sensitive attributes, then select the best models based on their performance on the validation sets. For each dataset and sensitive attribute pair, we report predictive performance using metrics including $C^{td}$, $AUC^{td}$, and $IBS$, along with fairness as the performance gap between the best and worst groups. As shown in Figure~\ref{fig:4}, these performance gaps (measured by $C^{td}$) are prevalent across all datasets and sensitive attributes. 
From this figure, we also observe that these gaps are more pronounced for age and race compared to sex across all datasets. While bias in model predictions has been extensively discussed in the context of medical classification and segmentation, as well as TTE prediction for tabular data, it has not been systematically quantified for TTE prediction in medical imaging. This study provides the first comprehensive analysis across a wide range of imaging modalities, TTE outcomes, and sensitive attributes.

\textbf{Statistical Tests.} We conduct statistical tests to verify the robustness of our findings against the variability in TTE prediction models and hyperparameters. 
Specifically, for each combination of dataset, sensitive attribute, and evaluation metric, we perform two-sided Wilcoxon signed-rank test~\cite{conover1999practical} 
\begin{wrapfigure}[18]{r}{0.5\textwidth}
\vspace{-0.59cm}
  \begin{center}
    \includegraphics[width=0.5\textwidth]{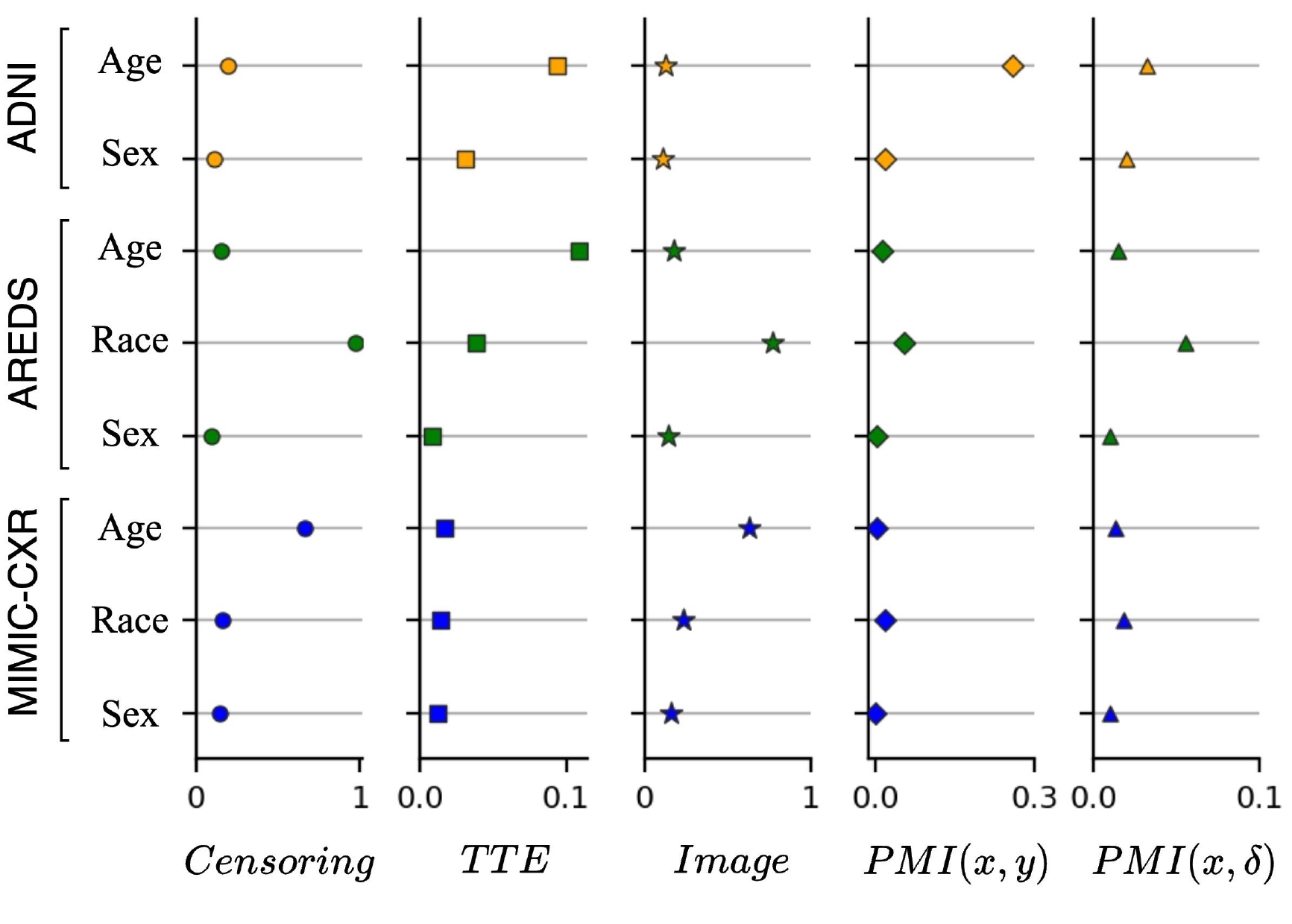}
  \end{center}
\vspace{-0.3cm}
  \caption{Quantification of the degree of various sources of bias across all datasets and sensitive attributes. Bias degrees range from 0 to 1, where 0 indicates no bias and 1 represents maximum bias within the datasets.\label{fig:5}}
\end{wrapfigure}
on the results across all TTE prediction models and hyperparameters to identify significant differences in predictive performance between groups (p-value $< 0.05$). As illustrated in Figure~\ref{fig:a2}, our analysis reveals significant performance disparities between groups across all experimental settings, highlighting the importance of considering fairness in TTE prediction tasks.

\textbf{Quantifying Sources of Bias.} To better understand the unfair behavior of TTE prediction models, we quantify the degree of each source of bias in various datasets and sensitive attribute settings. Specifically, to estimate disparities in PMI between RVs across groups, we compute normalized mutual information scores. To evaluate disparities in image features and TTE outcomes across groups, we calculate the Wasserstein distance~\cite{villani2009wasserstein}. Lastly, disparities in censoring rates are quantified by examining differences in the ratio of censored data across groups. Detailed methods for quantifying sources of bias can be found in Appendix~\ref{sec:f4}. 

Figure~\ref{fig:5} illustrates the degree of bias across these five sources for each dataset and sensitive attribute setting. The results reveal a correlation between the degree of bias and the fairness performance of TTE prediction models. For instance, in the AREDS dataset, the degree of bias in terms of disparities in censoring rates, in PMI between $X_Z$ and $Y$, and in PMI between $X_Z$ and $\Delta$ is significantly higher when considering race as the sensitive attribute compared to sex or age. This finding aligns with Figure~\ref{fig:4}, where the performance gap between racial groups is considerably larger than the gaps between sex or age groups. Additionally, we observe that settings with greater disparities in image features across groups exhibit larger performance gaps. This observation aligns with prior findings in medical image classification tasks, where higher subgroup separability is associated with greater performance disparities~\cite{jones2023role, yang2024limits}.

\begin{wrapfigure}[17]{r}{0.5\textwidth}
\vspace{-0.61cm}
  \begin{center}
    \includegraphics[width=0.5\textwidth]{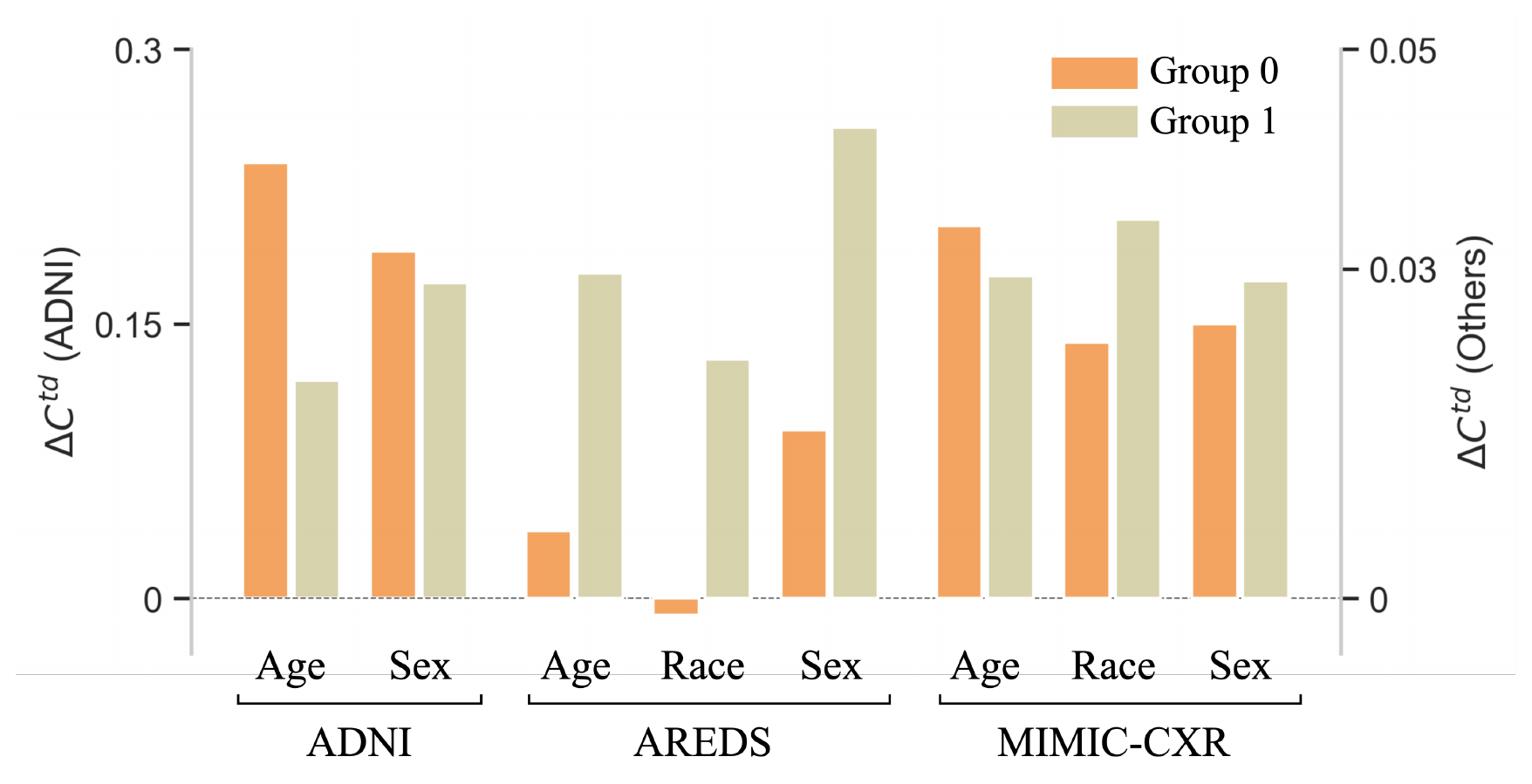}
  \end{center}
\vspace{-0.2cm}
  \caption{Per-group average performance gap ($\Delta C^{td}$) for TTE prediction models using a pre-training strategy compared to training from scratch across various datasets and sensitive attribute combinations. A positive $\Delta C^{td}$ indicates that the pre-training strategy enhances predictive performance relative to training from scratch.\label{fig:6}}
  % \vskip -0.in
\end{wrapfigure}

\textbf{The Role of Pre-Training.} Pre-training on large datasets has been proven effective in many applications~\cite{bommasani2021opportunities}. In this study, we investigate the impact of pre-training on both the accuracy and fairness of TTE prediction models. As shown in Figure~\ref{fig:6}, pre-training outperforms training models from 
scratch in terms of predictive performance. This improvement is particularly notable for the ADNI dataset, which is relatively small and contains more complex data structures (3D images) compared to the AREDS and MIMIC-CXR datasets (2D images). However, in terms of fairness, we do not observe a significant improvement with pre-training compared to training from scratch, as shown in Figure~\ref{fig:a4}. Specifically, the p-values from one-sided Wilcoxon signed-rank tests are larger than 0.05 in 18 out of 24 settings, suggesting that pre-training does not lead to more equitable predictions in most cases. These results imply that while pre-training enhances accuracy, combining it with fairness algorithms may be necessary to achieve both fairer and more accurate TTE prediction.

\begin{table*}[t]
\caption{Predictive and fairness performances of fairness algorithms across all dataset and sensitive attribute combinations. We report the actual predictive performance as percentages for DeepHit. For fair algorithms, relative changes in each metric compared to DeepHit are shown. The reported performances correspond to the best models selected via model selection conducted on the validation sets. \textcolor{blue}{Blue} indicates positive changes, \textcolor{red}{red} indicates negative changes. Metrics with $\downarrow$ are better when lower, and with $\uparrow$ when higher. Best fair algorithm performances are highlighted with \colorbox{gray!20}{gray cells}\label{tab:1}.}
% \caption{Predictive and fairness performances of fairness algorithms across all dataset and sensitive attribute combinations. Percentages show baseline performance (DeepHit), while relative changes for fairness algorithms are reported against DeepHit. Performances reflect the best models selected via validation. \textcolor{blue}{Blue} indicates positive changes, \textcolor{red}{red} indicates negative changes. Metrics with $\downarrow$ are better when lower, and with $\uparrow$ when higher. Best fairness algorithm performances are highlighted with \colorbox{gray!20}{gray cells}\label{tab:1}.}
% \vspace{-0.05cm}
\centering
\resizebox{\linewidth}{!}{
\begin{tabular}{l|ccc|ccc|ccc|ccc}
\toprule
\multicolumn{1}{c|}{Model} & \multicolumn{3}{c|}{Accuracy} & \multicolumn{3}{c|}{Fairness} & \multicolumn{3}{c|}{Accuracy} & \multicolumn{3}{c}{Fairness} \\
 & $AUC^{td} \uparrow$ & $IBS \downarrow$ & $C^{td}  \uparrow$ & $\mathcal{F}_{AUC^{td}} \downarrow$ & $\mathcal{F}_{IBS} \downarrow$ & $\mathcal{F}_{C^{td}} \downarrow$ & $AUC^{td} \uparrow$ & $IBS \downarrow$ & $C^{td}  \uparrow$ & $\mathcal{F}_{AUC^{td}} \downarrow$ & $\mathcal{F}_{IBS} \downarrow$ & $\mathcal{F}_{C^{td}} \downarrow$ \\
\midrule
& \multicolumn{6}{c|}{ADNI - Age}& \multicolumn{6}{c}{ADNI - Sex}\\ \midrule
DeepHit & 82.02 & 24.00 & 74.20 & 14.19 & 16.39 & 7.74 & 82.02 & 24.00 & 73.84 & 7.02 & 2.93 & 15.82 \\
DRO & \cellcolor{gray!20}\textcolor{blue}{1.46\%} & \textcolor{red}{-13.99\%} & \textcolor{red}{-6.40\%} & \textcolor{blue}{0.84\%} & \textcolor{red}{-46.61\%} & \textcolor{red}{-18.26\%} & \textcolor{red}{-2.65\%} & \cellcolor{gray!20}\textcolor{red}{-18.57\%} & \textcolor{red}{-1.69\%} & \textcolor{blue}{23.56\%} & \textcolor{red}{-17.17\%} & \cellcolor{gray!20}\textcolor{red}{-74.88\%} \\
SR & \textcolor{red}{-5.76\%} & \textcolor{blue}{0.90\%} & \cellcolor{gray!20}\textcolor{blue}{6.73\%} & \cellcolor{gray!20}\textcolor{red}{-49.81\%} & \textcolor{red}{-39.84\%} & \textcolor{red}{-26.50\%} & \textcolor{red}{-3.54\%} & \textcolor{blue}{1.99\%} & \cellcolor{gray!20}\textcolor{red}{-1.67\%} & \textcolor{blue}{388.82\%} & \cellcolor{gray!20}\textcolor{red}{-71.74\%} & \textcolor{red}{-24.78\%} \\
FRL & \textcolor{red}{-6.00\%} & \cellcolor{gray!20}\textcolor{red}{-18.63\%} & \textcolor{red}{-2.55\%} & \textcolor{blue}{53.91\%} & \cellcolor{gray!20}\textcolor{red}{-60.63\%} & \textcolor{red}{-35.86\%} & \cellcolor{gray!20}\textcolor{red}{-1.75\%} & \textcolor{red}{-12.58\%} & \textcolor{red}{-1.98\%} & \cellcolor{gray!20}\textcolor{red}{-2.95\%} & \textcolor{blue}{8.74\%} & \textcolor{red}{-35.60\%} \\
DI & \textcolor{blue}{1.00\%} & \textcolor{red}{-14.48\%} & \textcolor{red}{-2.19\%} & \textcolor{blue}{55.54\%} & \textcolor{red}{-37.51\%} & \cellcolor{gray!20}\textcolor{red}{-66.22\%} & \textcolor{red}{-15.81\%} & \textcolor{blue}{5.10\%} & \textcolor{red}{-5.57\%} & \textcolor{blue}{14.09\%} & \textcolor{blue}{193.83\%} & \textcolor{blue}{7.49\%} \\
CSA & \textcolor{red}{-4.18\%} & \textcolor{red}{-11.44\%} & \textcolor{red}{-5.79\%} & \textcolor{red}{-32.44\%} & \textcolor{red}{-42.55\%} & \textcolor{red}{-26.01\%} & \textcolor{red}{-3.89\%} & \textcolor{red}{-4.99\%} & \textcolor{red}{-10.38\%} & \textcolor{blue}{59.28\%} & \textcolor{red}{-0.51\%} & \textcolor{red}{-50.43\%} \\
\midrule
& \multicolumn{6}{c|}{AREDS - Age}& \multicolumn{6}{c}{AREDS - Race}\\ \midrule
DeepHit & 78.41 & 15.37 & 81.30 & 1.58 & 12.56 & 2.20 & 81.78 & 11.74 & 84.53 & 14.00 & 10.14 & 11.09 \\
DRO & \textcolor{blue}{0.88\%} & \textcolor{red}{-4.37\%} & \textcolor{blue}{0.08\%} & \textcolor{blue}{19.19\%} & \textcolor{red}{-20.08\%} & \textcolor{red}{-5.03\%} & \textcolor{blue}{1.78\%} & \textcolor{red}{-2.28\%} & \textcolor{red}{-1.41\%} & \cellcolor{gray!20}\textcolor{red}{-34.89\%} & \textcolor{red}{-10.93\%} & \cellcolor{gray!20}\textcolor{red}{-37.21\%} \\
SR & \textcolor{blue}{1.71\%} & \textcolor{red}{-2.52\%} & \textcolor{blue}{0.32\%} & \textcolor{red}{-1.93\%} & \textcolor{red}{-4.59\%} & \textcolor{blue}{15.44\%} & \textcolor{red}{-0.58\%} & \cellcolor{gray!20}\textcolor{red}{-3.73\%} & \textcolor{red}{-0.69\%} & \textcolor{red}{-24.63\%} & \textcolor{red}{-11.97\%} & \textcolor{red}{-5.65\%} \\
FRL & \textcolor{blue}{0.80\%} & \textcolor{red}{-0.70\%} & \textcolor{blue}{0.20\%} & \cellcolor{gray!20}\textcolor{red}{-25.41\%} & \textcolor{red}{-8.21\%} & \cellcolor{gray!20}\textcolor{red}{-35.01\%} & \textcolor{blue}{1.75\%} & \textcolor{blue}{4.41\%} & \cellcolor{gray!20}\textcolor{red}{-0.21\%} & \textcolor{red}{-33.49\%} & \cellcolor{gray!20}\textcolor{red}{-16.10\%} & \textcolor{red}{-11.51\%} \\
DI & \cellcolor{gray!20}\textcolor{blue}{2.12\%} & \textcolor{red}{-1.74\%} & \cellcolor{gray!20}\textcolor{blue}{0.73\%} & \textcolor{blue}{83.73\%} & \textcolor{red}{-6.53\%} & \textcolor{blue}{9.89\%} & \cellcolor{gray!20}\textcolor{blue}{2.47\%} & \textcolor{red}{-0.67\%} & \textcolor{red}{-0.84\%} & \textcolor{red}{-8.58\%} & \textcolor{red}{-9.64\%} & \textcolor{red}{-26.11\%} \\
CSA & \textcolor{blue}{0.89\%} & \cellcolor{gray!20}\textcolor{red}{-5.94\%} & \textcolor{blue}{0.11\%} & \textcolor{red}{-1.95\%} & \cellcolor{gray!20}\textcolor{red}{-29.56\%} & \textcolor{blue}{4.60\%} & \textcolor{blue}{1.37\%} & \textcolor{red}{-1.32\%} & \textcolor{red}{-0.23\%} & \textcolor{red}{-10.74\%} & \textcolor{red}{-8.14\%} & \textcolor{red}{-21.97\%} \\
\midrule
& \multicolumn{6}{c|}{AREDS - Sex}& \multicolumn{6}{c}{MIMIC-CXR - Age}\\ \midrule
DeepHit & 79.08 & 15.36 & 81.77 & 0.76 & 3.84 & 1.32 & 78.61 & 20.38 & 76.21 & 3.06 & 1.49 & 5.93 \\
DRO & \textcolor{blue}{1.21\%} & \textcolor{red}{-3.62\%} & \textcolor{red}{-0.54\%} & \textcolor{blue}{12.75\%} & \cellcolor{gray!20}\textcolor{red}{-6.52\%} & \cellcolor{gray!20}\textcolor{red}{-95.74\%} & \textcolor{red}{-1.55\%} & \textcolor{blue}{13.50\%} & \textcolor{red}{-1.61\%} & \textcolor{red}{-49.46\%} & \textcolor{blue}{58.14\%} & \textcolor{red}{-17.81\%} \\
SR & \textcolor{red}{-0.07\%} & \textcolor{red}{-4.02\%} & \textcolor{red}{-0.14\%} & \textcolor{red}{-47.46\%} & \textcolor{red}{-4.14\%} & \textcolor{red}{-85.27\%} & \textcolor{red}{-0.96\%} & \textcolor{blue}{2.75\%} & \cellcolor{gray!20}\textcolor{red}{-0.37\%} & \cellcolor{gray!20}\textcolor{red}{-82.53\%} & \textcolor{red}{-8.31\%} & \textcolor{blue}{2.86\%} \\
FRL & \cellcolor{gray!20}\textcolor{blue}{1.76\%} & \textcolor{red}{-5.37\%} & \textcolor{red}{-0.54\%} & \cellcolor{gray!20}\textcolor{red}{-51.07\%} & \textcolor{red}{-1.20\%} & \textcolor{blue}{23.24\%} & \textcolor{red}{-4.28\%} & \textcolor{blue}{5.26\%} & \textcolor{red}{-2.42\%} & \textcolor{red}{-61.41\%} & \cellcolor{gray!20}\textcolor{red}{-60.79\%} & \cellcolor{gray!20}\textcolor{red}{-21.96\%} \\
DI & \textcolor{blue}{0.18\%} & \textcolor{red}{-5.84\%} & \cellcolor{gray!20}\textcolor{blue}{1.31\%} & \textcolor{red}{-13.82\%} & \textcolor{blue}{16.12\%} & \textcolor{red}{-65.70\%} & \cellcolor{gray!20}\textcolor{blue}{0.41\%} & \textcolor{blue}{5.29\%} & \textcolor{red}{-0.63\%} & \textcolor{red}{-52.21\%} & \textcolor{red}{-44.27\%} & \textcolor{red}{-10.24\%} \\
CSA & \textcolor{red}{-0.18\%} & \cellcolor{gray!20}\textcolor{red}{-8.23\%} & \textcolor{blue}{0.46\%} & \textcolor{red}{-28.46\%} & \textcolor{blue}{17.58\%} & \textcolor{red}{-71.42\%} & \textcolor{blue}{0.36\%} & \cellcolor{gray!20}\textcolor{red}{-3.19\%} & \textcolor{red}{-0.97\%} & \textcolor{red}{-66.46\%} & \textcolor{red}{-29.21\%} & \textcolor{red}{-13.72\%} \\
\midrule
& \multicolumn{6}{c|}{MIMIC-CXR - Race}& \multicolumn{6}{c}{MIMIC-CXR - Sex}\\ \midrule
DeepHit & 78.61 & 21.10 & 76.21 & 2.89 & 0.87 & 4.03 & 78.61 & 20.38 & 76.21 & 4.77 & 1.23 & 2.09 \\
DRO & \textcolor{red}{-0.72\%} & \textcolor{blue}{3.13\%} & \textcolor{red}{-0.78\%} & \textcolor{red}{-5.69\%} & \textcolor{blue}{31.34\%} & \cellcolor{gray!20}\textcolor{red}{-4.80\%} & \textcolor{red}{-0.35\%} & \textcolor{blue}{4.51\%} & \textcolor{red}{-1.20\%} & \textcolor{blue}{12.28\%} & \textcolor{blue}{92.27\%} & \textcolor{red}{-9.61\%} \\
SR & \cellcolor{gray!20}\textcolor{red}{-0.11\%} & \textcolor{red}{-0.90\%} & \textcolor{red}{-0.41\%} & \textcolor{blue}{10.28\%} & \cellcolor{gray!20}\textcolor{red}{-74.37\%} & \textcolor{red}{-0.80\%} & \cellcolor{gray!20}\textcolor{red}{-0.12\%} & \textcolor{blue}{2.31\%} & \cellcolor{gray!20}\textcolor{red}{-0.19\%} & \textcolor{blue}{17.02\%} & \textcolor{blue}{25.35\%} & \textcolor{red}{-1.37\%} \\
FRL & \textcolor{red}{-0.61\%} & \textcolor{red}{-2.44\%} & \cellcolor{gray!20}\textcolor{blue}{0.01\%} & \textcolor{blue}{36.89\%} & \textcolor{red}{-19.46\%} & \textcolor{blue}{0.34\%} & \textcolor{red}{-2.24\%} & \textcolor{blue}{0.21\%} & \textcolor{red}{-0.57\%} & \textcolor{blue}{15.69\%} & \textcolor{blue}{10.97\%} & \textcolor{red}{-4.16\%} \\
DI & \textcolor{red}{-0.44\%} & 0.00\% & \textcolor{red}{-0.52\%} & \textcolor{blue}{37.80\%} & \textcolor{blue}{30.77\%} & \textcolor{red}{-0.43\%} & \textcolor{red}{-1.94\%} & \textcolor{blue}{1.21\%} & \textcolor{red}{-0.34\%} & \textcolor{blue}{14.26\%} & \textcolor{red}{-8.22\%} & \cellcolor{gray!20}\textcolor{red}{-13.76\%} \\
CSA & \textcolor{red}{-0.42\%} & \cellcolor{gray!20}\textcolor{red}{-4.61\%} & \textcolor{red}{-0.96\%} & \cellcolor{gray!20}\textcolor{red}{-14.83\%} & \textcolor{blue}{36.14\%} & \textcolor{blue}{0.44\%} & \textcolor{red}{-0.47\%} & \cellcolor{gray!20}\textcolor{red}{-2.58\%} & \textcolor{red}{-0.63\%} & \cellcolor{gray!20}\textcolor{blue}{10.63\%} & \cellcolor{gray!20}\textcolor{red}{-49.08\%} & \textcolor{red}{-0.46\%} \\
\bottomrule
\end{tabular}}
\vspace{-0.5cm}
\end{table*}

\textbf{Performance of Fairness Algorithms in TTE Prediction.} We conduct an experiment to evaluate the effectiveness of fairness algorithms in reducing disparities between groups in TTE prediction. In this experiment, we use DeepHit as the base TTE prediction model and integrate various fairness algorithms to achieve equitable predictions. To ensure a comprehensive evaluation, we select algorithms from pre-processing (i.e., SR), in-processing (i.e., DI, FRL, DRO), and post-processing approaches (i.e., CSA). 
% For each fairness algorithm, we perform hyperparameter search and select the best model based on the strategy outlined in Section~\ref{sec:setup}. 
As shown in Table~\ref{tab:1}, existing fairness methods reduce performance gaps between groups compared to DeepHit in settings where bias sources are substantial (e.g., ADNI with age, AREDS with race, and MIMIC-CXR with age). However, when bias sources are small, enforcing fairness can sometimes exacerbate performance disparities. Moreover, no method consistently outperforms DeepHit across all settings. To verify the significance of our observations, we perform a Friedman test followed by a Nemenyi post-hoc test~\cite{demvsar2006statistical} across all algorithms, datasets, and sensitive attribute settings. As shown in Figure~\ref{fig:a5}, the results confirm that current fairness algorithms do not significantly outperform DeepHit in mitigating bias. Additionally, we observe trade-offs in predictive performance in most cases where fairness interventions reduce group disparities: while certain sources of bias (e.g., disparities in image features) are mitigated, others often persist. These findings highlight the need for developing new fairness algorithms capable of addressing multiple sources of bias without compromising predictive performance.

% \textbf{Fairness-Utility Trade-Off.} 
% Incorporating fairness shifts the objective from pure utility optimization to balancing utility and fairness. To assess this trade-off in fair TTE prediction methods, we compute equity scaling scores~\cite{luo2024harvard} across datasets and sensitive attributes. As shown in Figure~\ref{fig:a6}, different methods exhibit varying trade-offs. However, these patterns are inconsistent across metrics. Notably, DRO performs best for $AUC^{td}$, while CSA and SR excel for $IBS$ and $C^{td}$, respectively. This highlights the need for careful selection of fairness algorithms to achieve an optimal fairness-utility trade-off.

\textbf{Fair TTE Prediction Under Distribution Shift.} 
Fairness algorithms are known to lose effectiveness under distribution shifts, particularly in medical image diagnosis tasks~\cite{yang2024limits,jones2025rethinking}. Building on this insight, we investigate the impact of distribution shifts on fair TTE prediction. As illustrated in Figure~\ref{fig:3}, we define distribution shifts as scenarios where correlations between the sensitive attribute and other random variables in the causal graph exist in the training dataset but disappear in the testing dataset. To simulate such shifts, we manipulate the data generation process by corrupting images, TTE labels, or censoring indicators for one group while keeping the other group unchanged. The real-world motivations and detailed descriptions of these causal distribution shifts are provided in Appendices~\ref{sec:g2} and~\ref{sec:f5} .

\begin{wrapfigure}[26]{r}{0.5\textwidth}
\vspace{-0.7cm}
  \begin{center}
    \includegraphics[width=0.5\textwidth]{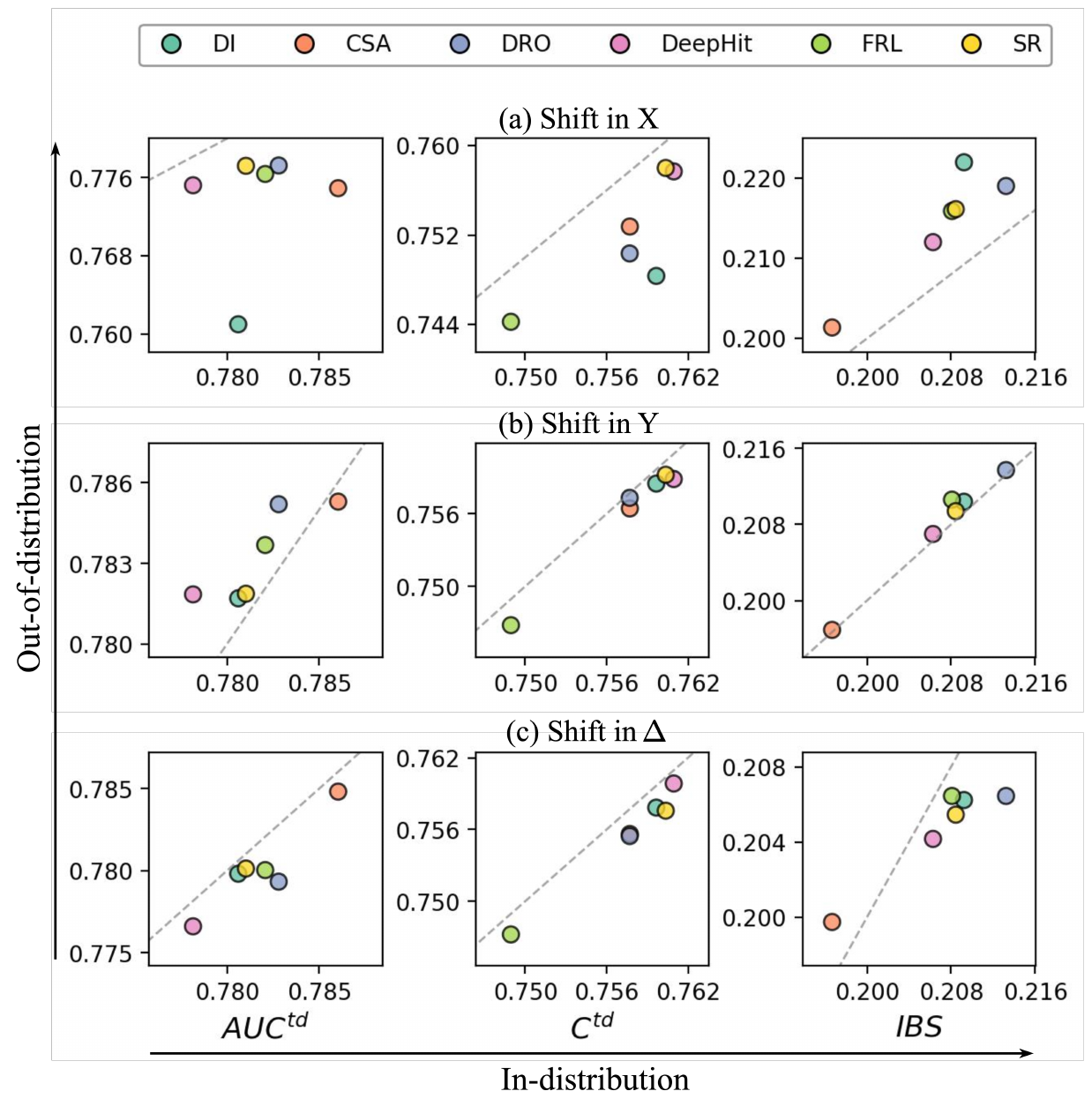}
  \end{center}
  % \vskip -0.22in
  \vspace{-0.2cm}
  % \caption{Predictive performance comparison of fair TTE prediction models in in-distribution (ID) vs. out-of-distribution (OOD) learning scenarios on MIMIC-CXR dataset with sex as sensitive attribute. The displayed results represent average performance across all random seeds. Points on the dashed line indicate equal performance in both scenarios. For $AUC^{td}$ and $C^{td}$, points in lower triangles means model performances in OOD are worse than ID. For $IBS$, points in upper triangles means model performances in OOD are worse than ID.\label{fig:7}}
  \caption{Predictive performance comparison of fair TTE prediction models in ID vs. OOD settings on the MIMIC-CXR dataset with sex as the sensitive attribute. Results are averaged across all random seeds. Points on the dashed line indicate equal performance in both settings. For $AUC^{td}$ and $C^{td}$ (resp. $IBS$), points below (resp. above) the line indicate degraded performance in OOD.\label{fig:7}}
  % \vskip -0.25in
\end{wrapfigure}

Figure~\ref{fig:7} compares the predictive performance of models under in-distribution and distribution shift learning scenarios for the MIMIC-CXR dataset, where sex is considered the sensitive attribute. The complete results for all settings are provided in Appendix~\ref{sec:g5}. Our findings indicate that different types of distribution shifts impact model performance in distinct ways. Specifically, adding noise to TTE labels (shift in $Y$) leads to a significant decline in $IBS$, as the increased label uncertainty makes accurate predictions more challenging. Flipping censoring indicators (shift in $\Delta$), on the other hand, severely degrades ranking-based metrics such as $C^{td}$ and $AUC^{td}$, as it reduces the availability of comparable pairs during training, making it harder for the model to learn an effective ranking function. Additionally, introducing noise to medical images (shift in $X$) negatively affects all performance metrics, as degraded image quality limits the model’s ability to extract meaningful features. These observations align with our expectations—corrupting TTE labels directly impacts metrics measuring error between predicted and ground-truth TTE, while censoring flips disrupt ranking-based evaluations by impairing the model’s ability to differentiate survival times.

Regarding fairness, our observations reveal mixed results. In some cases, distribution shifts negatively impact fairness, while in others, they lead to improvements. Notably, adding noise to the data of one group generally degrades predictive performance on that group. However, we also observe that this noise can similarly affect the other group—sometimes even more severely—thereby reducing the performance gap between groups and unintentionally improving fairness. These findings underscore the complexities of achieving fairness in the presence of distribution shifts and highlight the need for developing TTE algorithms that are both fair and robust in such challenging scenarios.
\section{Limitation and Broader Impact}

While FairTTE represents a significant step forward in advancing fairness research in medical image prognosis by providing the first comprehensive benchmark across multiple datasets and fairness algorithms, we acknowledge several limitations in the current study. Specifically, our analysis focuses on the standard TTE prediction setting, which assumes a single clinical risk and non-informative right censoring. In real-world clinical applications, however, patients often face multiple competing risks and may drop out of studies for reasons that introduce informative censoring, making the fairness landscape considerably more complex. Addressing fairness in such settings remains an important direction for future research. Additionally, although we adopt group fairness definitions (i.e., minimizing performance gaps across subgroups), we recognize that strictly enforcing these criteria can, in some clinical scenarios, degrade overall utility or harm performance for all groups. In such scenarios, alternative fairness notions may be more appropriate, depending on the clinical context and ethical objectives.

Selecting an appropriate fairness criterion in the medical domain requires careful consideration of the clinical context, ethical principles, and statistical validity. An effective fairness notion should align with the model’s intended use, its potential impact on different patient subgroups, and real-world constraints. In our study, we focus on statistical group fairness (i.e., predictive performance gaps across subgroups) which is widely examined in the medical image analysis literature. This choice is grounded in the assumption that any causal pathway from sensitive attributes represents an unfair influence in the causal graph and should be mitigated through fairness constraints. However, we acknowledge that in practical clinical scenarios, such causal pathways may reflect fair and clinically meaningful relationships. Enforcing group fairness in these cases may inadvertently remove relevant information and degrade predictive performance.

Moving forward, we suggest several directions for fairness research in medical image analysis: (1) Identifying the causal nature of bias to distinguish between fair and unfair sources, enabling models to address specific pathways appropriately; (2) Developing fairness metrics and mitigation strategies that preserve clinically relevant (fair) pathways while minimizing the effect of unfair ones; and (3) Collaborating with clinicians and domain experts to define context-specific fairness objectives that are aligned with both clinical utility and ethical standards.
\section{Conclusion}
We systematically investigate fair TTE prediction in medical imaging, introducing a unified framework to define and quantify bias sources and establishing a comprehensive and realistic evaluation. Our framework includes three TTE prediction methods, five fairness algorithms, and three large-scale public datasets spanning diverse modalities and outcomes. Extensive experiments reveal critical insights, including the link between bias sources and model unfairness, the inconsistent effectiveness of existing fairness methods, the role of pre-training strategies, and the challenges of causal distribution shifts. We hope our fine-grained analysis encourages more rigorous evaluations and drives the development of new fairness algorithms for TTE prediction.
% We systematically investigate fair TTE prediction for medical imaging, introduce a unified framework to define and quantify different sources of bias in medical images, and establish a comprehensive benchmark for realistic evaluation. Our benchmark encompasses 3 cutting-edge TTE prediction methods, 5 fairness algorithms, and 3 large-scale publicly available medical imaging datasets, covering a diverse range of imaging modalities and TTE outcomes. Through extensive experiments, we uncover several critical insights into fair TTE prediction with significant implications for future research. These include the correlation between bias sources and model unfairness, the inconsistent performance of existing fairness algorithms, the influence of pre-training strategies, and the challenges posed by causal distribution shifts. We hope that our fine-grained analysis will inspire more realistic and rigorous evaluations and foster the development of novel fairness algorithms for TTE prediction.
\section*{Acknowledgements}

\noindent This work was funded in part by the National Science Foundation under award number IIS-2145625 and by the National Institutes of Health under awards number R01AI188576 and R01CA301579.
\bibliographystyle{plain}
\bibliography{ref}
\addtocontents{toc}{\protect\setcounter{tocdepth}{2}}

\appendix

\newpage

\renewcommand{\contentsname}{Appendix Contents}
\setcounter{tocdepth}{1}
\tableofcontents
\clearpage
% \addcontentsline{toc}{section}{Appendix}

\renewcommand{\thefigure}{A\arabic{figure}}
\renewcommand{\thetable}{A\arabic{table}}
\setcounter{figure}{0}
\setcounter{table}{0}

\section{Related Works}\label{sec:a}
\paragraph{TTE Prediction.} TTE prediction models can generally be classified into two categories: continuous-time and discrete-time models, each with distinct approaches for handling event timing. Continuous-time models treat time as a continuous variable and often extend traditional models like Cox regression. For example, DeepSurv~\cite{katzman2018deepsurv} extends the Cox regression by using a deep neural network with non-linear activation functions in hidden layers. Cox-Time~\cite{kvamme2019time} further builds on DeepSurv, introducing time-dependent predictors that allow for the estimation of time-varying effects. In contrast, discrete-time models treat time as a series of distinct intervals and typically use classification techniques. DeepHit~\cite{lee2018deephit} learns survival times directly without assuming a specific underlying stochastic process, parameterizing the discrete probability mass function. Another method, Nnet-survival~\cite{gensheimer2019scalable}, parametrizes the discrete hazard function using a neural network and optimizes the negative log-likelihood loss.
\paragraph{Fairness in Machine Learning.} Fairness in machine learning has gained significant attention in recent years, with a focus on ensuring models are unbiased and equitable across individuals and groups. \underline{\textit{Fairness metrics.}} Fairness metrics can be broadly categorized into two types: group fairness~\cite{dwork2012fairness,corbett2017algorithmic,zafar2017fairness,hardt2016equality} and individual fairness~\cite{dwork2012fairness,sharifi2019average}. Group fairness ensures that models are fair across different demographic groups, while individual fairness emphasizes that similar individuals should be treated similarly. \underline{\textit{Fairness algorithms}}. To address fairness and bias issues, bias mitigation methods are generally classified into three approaches: pre-processing, which focuses on modifying the input data before model training~\cite{kamiran2012data,calmon2017optimized}; in-processing, which incorporates fairness constraints during model training~\cite{kamishima2012fairness,zhang2018mitigating,madras2018learning,shui2022learning}; and post-processing, which adjusts model outputs to improve fairness~\cite{hardt2016equality,jiang2020wasserstein}.
\paragraph{Fairness in Medical Imaging.} In medical image analysis, machine learning (ML) models have been shown to exhibit systematic biases related to various attributes such as race, gender, and age~\cite{seyyed2021underdiagnosis,larrazabal2020gender}. These biases are prevalent across different medical imaging modalities, including chest X-rays~\cite{seyyed2020chexclusion}, CT scans~\cite{zhou2021radfusion}, and skin dermatology images~\cite{kinyanjui2020fairness}. While several efforts have been made to benchmark fairness algorithms on medical images, existing datasets~\cite{irvin2019chexpert,kovalyk2022papila,tschandl2018ham10000,groh2021evaluating} and benchmarks~\cite{zong2023medfair,tian2024fairseg} primarily focus on diagnostic tasks like image classification and segmentation. Unfortunately, they often overlook the crucial domain of medical prognosis, which involves predicting TTE outcomes.
\paragraph{Fairness in TTE Prediction.} Despite significant advances in TTE prediction, research on fairness in this area remains limited. \underline{\textit{Fairness metrics for TTE prediction.}} Fairness metrics for TTE prediction have only recently been defined. These metrics can be roughly classified into three categories based on their objectives: (i) ensuring similar predicted TTE outcomes for similar data points~\cite{keya2021equitable,rahman2022fair,zhang2023individual,zhang2023censored,wang2024individual}, (ii) ensuring similar predicted outcomes for data points from different groups~\cite{keya2021equitable,rahman2022fair,zhao2023fairness}, and (iii) ensuring similar predictive performance across different groups~\cite{do2023fair,hu2024fairness,zhang2021fair,zhang2022longitudinal}.  
\underline{\textit{Fairness algorithms for TTE prediction.}} Building on these metrics, several methods have been proposed to achieve fair TTE prediction. One approach incorporates fairness as a regularization term during model training~\cite{keya2021equitable,rahman2022fair,do2023fair}, ensuring that the model accounts for fairness constraints throughout its optimization process. Another approach focuses on improving worst-group accuracy by leveraging distributionally robust optimization techniques~\cite{hu2024fairness,hashimoto2018fairness,sagawa2019distributionally}, which aim to enhance performance for underrepresented or disadvantaged groups. In addition to these in-processing methods, recent work has also explored pre- and post-processing strategies to address fairness in TTE prediction~\cite{zhao2023fairness}. However, these efforts are limited to tabular data and fail to consider medical images, which are essential and pervasive in medical prognosis tasks.

\vfill
\section{Missing Proof}\label{sec:b}
\paragraph{Proof for Theorem~\ref{thm:1}}\label{sec:b1}
For any $a,a' \in \mathcal{A}$, we have:
\begin{align*}
    \Er(f_{a'},h,D_{a'}) &\leq \Er(f_{a},h,D_{a}) + \left|\Er(f_{a},h,D_{a}) - \Er(f_{a'},h,D_{a'})\right| \\
    &\overset{(1)}{\leq} \Er(f_{a},h,D_{a}) + \left|\Er(f_{a},h,D_{a}) - \Er(h,h^{\ast},D_{a})\right| \\
    &+ \left|\Er(h,h^{\ast},D_{a}) - \Er(h,h^{\ast},D_{a'})\right| + \left|\Er(h,h^{\ast},D_{a'}) - \Er(f_{a'},h,D_{a'})\right| \\
    &\overset{(2)}{\leq} \Er(f_{a},h,D_{a}) + \Er(f_{a},h^{\ast},D_{a}) + \mathcal{D}(\mathcal{H},D_{a},D_{a'}) + \Er(f_{a'},h^{\ast},D_{a'}) \\
    &\overset{(3)}{=} \Er(f_{a},h,D_{a}) + \eta(\mathcal{H},f_{a},f_{a'}) + \mathcal{D}(\mathcal{H},D_{a},D_{a'})
\end{align*}
where $h^{\ast} = \arg\min_{h' \in \mathcal{H}} \left(\Er(f_a,h',D_a) + \Er(f_{a'},h',D_{a'})\right)$. We have $\overset{(1)}{\leq}$ by using inequality $|a+b| \leq |a| + |b|$; $\overset{(2)}{\leq}$ by using triangle inequality for $\Er$ metric and $\left|\Er(h,h^{\ast},D_{a}) - \Er(h,h^{\ast},D_{a'})\right| \leq \max_{h', h'' \in \mathcal{H}}\left| \Er(h',h'',D_a) - \Er(h',h'',D_{a'}) \right| = \mathcal{D}(\mathcal{H},D_a,D_{a'})$; $\overset{(3)}{=}$ because $\eta(\mathcal{H}, f_a,f_{a'}) = \left(\Er(f_a,h^{\ast},D_a) + \Er(f_{a'},h^{\ast},D_{a'})\right)$ by definition. Subtracting $\Er(f_{a},h,D_{a})$ from both sides and taking $\max$ operator, we have:
\begin{equation*}
    \mathcal{F}_{\Er}(h) = \max_{a, a' \in \mathcal{A}} \left | \Er\left(f_a,h,D_a\right) - \Er\left(f_{a'},h,D_{a'}\right) \right | \leq \max_{a, a' \in \mathcal{A}} \left( \eta(\mathcal{H},f_{a},f_{a'}) + \mathcal{D}(\mathcal{H},D_{a},D_{a'}) \right)
\end{equation*}

\paragraph{Discussion on the assumption of performance metric.}\label{sec:b2}
The proof of Theorem~\ref{thm:1} relies on the assumption that the performance metric $\Er$ satisfies the properties of triangle inequality and symmetry. This assumption is relatively mild and holds for many commonly used performance metrics. For instance,~\cite{shaker2023multi} introduced the symmetric discordance index (SDI), a ranking-based metric that adheres to these properties, demonstrating the practical applicability of this assumption in practice.

\paragraph{Proof of Proposition~\ref{thm:2}}
For $a \in \mathcal{A}$ with $I_a(Z, T) = I_a(X, T)$, we have:

\begin{align}
    \log P(t|x,a) &= \log \left( \int P(t,z|x,a) dz \right) \nonumber \\
    &= \log \left( \int P(t|z,a) P(z|x,a) dz \right) \nonumber \\
    &= \log \left( \mathbb{E}_{P(z|x)} \left[ P(t|z,a) \right] \right) \nonumber \\
    &\overset{(1)}{\geq} \mathbb{E}_{P(z|x)} \left[ \log p(t|z,a) \right] \label{eq:s4}
\end{align}

We have $\overset{(1)}{\geq}$ by using Jensen's inequality. $\forall a' \in \mathcal{A}$, taking expectation w.r.t. $P(x,t|a')$ over both sides, we have:

\begin{align}
    &\mathbb{E}_{P(x,t|a')} \left[ \log P(t|x,a) - \mathbb{E}_{P(z|x)} \left[ \log P(t|z,a) \right] \right]  \nonumber \\
    &= \int \int \left( \log P(t|x,a) - \mathbb{E}_{P(z|x)} \left[ \log P(t|z,a) \right] \right) P(x,t|a') dx dt \nonumber \\
    &= \int \int \left( \log P(t|x,a) - \mathbb{E}_{P(z|x)} \left[ \log P(t|z,a) \right] \right) P(x,t|a) \frac{P(x,t|a')}{P(x,t|a)} dx dt \nonumber \\
    &=\mathbb{E}_{P(x,t|a)} \left[ \left( \log P(t|x,a) - \mathbb{E}_{P(z|x)} \left[ \log P(t|z,a) \right] \right) \frac{P(x,t|a')}{P(x,t|a)} \right]  \nonumber
\end{align}
\vfill
\begin{align}
    &\overset{(1)}{\leq} \left( \max_{x,t}\frac{P(x,t|a')}{P(x,t|a)} \right) \mathbb{E}_{P(x,t|a)} \left[  \log P(t|x,a) - \mathbb{E}_{P(z|x)} \left[ \log P(t|z,a) \right] \right]  \nonumber \\
    &= \left( \max_{x,t}\frac{P(x,t|a')}{P(x,t|a)} \right) \left( \mathbb{E}_{P(x,t|a)} \left[ \log P(t|x,a) \right] - \mathbb{E}_{P(z,t|a)} \left[   \log P(t|z,a) \right] \right)  \nonumber \\
    &= \left( \max_{x,t}\frac{P(x,t|a')}{P(x,t|a)} \right) \left( H_a(T,X) - H_{a'}(T,Z) \right) \nonumber \\
    &= \left( \max_{x,t}\frac{P(x,t|a')}{P(x,t|a)} \right) \left( \left( H_a(T) - H_a(T,Z) \right) - \left( H_a(T) - H_a(T,X) \right) \right) \nonumber \\
    &= \left( \max_{x,t}\frac{P(x,t|a')}{P(x,t|a)} \right) \left( I_a(T,Z) - I_a(T,X) \right) \nonumber \\
    &\overset{(2)}{=} 0 \label{eq:s5}
\end{align}

We have $\overset{(1)}{\leq}$ because $\log P(t|x,a) - \mathbb{E}_{P(z|x)} \left[ \log P(t|z,a) \right] \geq 0$ according to Eq. (\ref{eq:s4}); $\overset{(2)}{=}$ because $I_a(T,Z) = I_a(T,X)$. Based on Eq. (\ref{eq:s5}), we have:

\begin{align}
    \mathbb{E}_{P(x,t|a')} \left[ \log P(t|x,a) \right]  &=\mathbb{E}_{P(x,t|a')} \left[ \mathbb{E}_{P(z|x)} \left[ \log P(t|z,a) \right] \right] \nonumber \\
    &=\mathbb{E}_{P(t,z|a')} \left[ \log P(t|z,a) \right] \label{eq:s6}
\end{align}

We also have:

\begin{align}
    \mathbb{E}_{P(t,z|a')} \left[ \log P(t|z,a') \right] &= - H_{a'}(T|Z) \nonumber \\
    &= I_{a'}(T,Z) - H_{a'}(T) \nonumber \\
    &\overset{(1)}{\leq} I_{a'}(T,X) - H_{a'}(T) \nonumber \\
    &= - H_{a'}(T|X) \nonumber \\
    &= \mathbb{E}_{P(x,t|a')} \left[ \log P(t|x,a') \right] \label{eq:s7}
\end{align}

We have $\overset{(1)}{\leq}$ by using data processing inequality. Finally, we have:

\begin{align}
    & \;\;\;\;\; \mathbb{E}_{P(z|a')} \left[ \mathcal{D}_{KL} \left( P(t|z,a') \parallel P(t|z,a) \right) \right] \nonumber \\
    &\overset{(1)}{=} \mathbb{E}_{P(z|a')} \left[ \mathcal{D}_{KL} \left( P(t|z,a') \parallel P(t|z,a) \right) \right] - \mathbb{E}_{P(x|a')} \left[ \mathcal{D}_{KL} \left( P(t|x,a') \parallel P(t|x,a) \right) \right] \nonumber \\
    &= \mathbb{E}_{P(t,z|a')} \left[ \log P(t|z,a') - \log P(t|z,a) \right] - \mathbb{E}_{P(x,t|a')} \left[ \log P(t|x,a') - \log P(t|x,a) \right] \nonumber \\
    &= \left( \mathbb{E}_{P(t,z|a')} \left[ \log P(t|z,a') \right] - \mathbb{E}_{P(x,t|a')} \left[ \log P(t|x,a') \right] \right) \nonumber \\
    &+ \left( \mathbb{E}_{P(x,t|a')} \left[ \log P(t|x,a) \right] - \mathbb{E}_{P(t,z|a')} \left[ \log P(t|z,a) \right] \right) \nonumber \\
    &\overset{(2)}{=} 0 \label{eq:s8}
\end{align}

We have $\overset{(1)}{=}$ because the shift between two domains w.r.t. input space $\mathcal{X}$ is covariate shift; $\overset{(2)}{=}$ by using Eq. (\ref{eq:s6}) and Eq. (\ref{eq:s7}) and the fact that KL-divergence is non-negative. Note that Eq. (\ref{eq:s8}) implies that the shift between these two domains w.r.t. representation space $\mathcal{Z}$ is also covariate shift (i.e., $P(t|z,a) = P(t|z,a') = P(t|z), \forall a, a' \in \mathcal{A}$).
\begin{table}[t]
\caption{Overview of medical image datasets for fair TTE prediction evaluation.}\label{tab:a1}
\vskip 0.1in
\centering
\resizebox{\textwidth}{!}{
\begin{tabular}{lllllccc}
\toprule
Dataset                    & Prediction Task                                              & Modality                              & Subgroup              & Attribute        & \# images & Censoring rate & Mean TTE \\\midrule
\multirow{7}{*}{AREDS} & \multirow{7}{*}{Late AMD} & \multirow{7}{*}{\makecell[l]{Retinal\\ Fundus}} &                       & Total            & 129708        & 83.9\%        & 4.4 (years)\\\cline{4-8}
                           &                                                   &                                       & \multirow{2}{*}{Age}  & $\le$70   & 44224        & 83.1\%        & 5.1
\\
                           &                                                   &                                       &                       & \textgreater{}70 & 85484        & 84.4\%        & 3.9
\\\cline{4-8}
                           &                                                   &                                       & \multirow{2}{*}{Sex}  & Female           & 71837        & 83.0\%        & 4.4
\\
                           &                                                   &                                       &                       & Male             & 57871        & 85.1\%        & 4.3
\\\cline{4-8}
                           &                                                   &                                       & \multirow{2}{*}{Race} & Non-white        & 4888         & 99.2\%        & 3.1
\\
                           &                                                   &                                       &                       & White            & 124820       & 83.3\%        & 4.4\\\midrule
\multirow{7}{*}{MIMIC-CXR} & \multirow{7}{*}{\makecell[l]{In-hospital\\ Mortality}} & \multirow{7}{*}{\makecell[l]{Chest\\ X-ray}}    &                       & Total            & 269360       & 61.7\%        & 488.6 (days)\\\cline{4-8}
                           &                                                   &                                       & \multirow{2}{*}{Age}  & $\le$60   & 103437       & 77.3\%        & 503.3
\\
                           &                                                   &                                       &                       & \textgreater{}60 & 165923       & 52.0\%        & 484.2
\\\cline{4-8}
                           &                                                   &                                       & \multirow{2}{*}{Sex}  & Female           & 125742       & 63.9\%        & 514.4
\\
                           &                                                   &                                       &                       & Male             & 143618       & 59.7\%        & 468.4
\\\cline{4-8}
                           &                                                   &                                       & \multirow{2}{*}{Race} & Non-white        & 83234        & 66.7\%        & 487.2
\\
                           &                                                   &                                       &                       & White            & 186126       & 59.4\%        & 489.1\\\midrule
\multirow{5}{*}{ADNI}      & \multirow{5}{*}{\makecell[l]{Alzheimer’s\\ Disease}}              & \multirow{5}{*}{\makecell[l]{Brain \\MRI}}      &                       & Total            & 2227

         & 63.2\%        & 35.9 (months)\\\cline{4-8}
                           &                                                   &                                       & \multirow{2}{*}{Age}  & $\le$80& 1597
& 62.2\%& 37.1
\\
                           &                                                   &                                       &                       & \textgreater{}80& 630
& 65.7\%& 32.6
\\\cline{4-8}
                           &                                                   &                                       & \multirow{2}{*}{Sex}  & Female           & 986
& 64.7\%        & 38.5
\\
                           &                                                   &                                       &                       & Male             & 1241& 62.0\%        & 33.9
\\\bottomrule
\end{tabular}}
\end{table}
\section{Dataset Details}\label{sec:c}
\subsection{AREDS}
\subsubsection{Dataset Description}
The Age-Related Eye Disease Study (AREDS)~\cite{ferris2005simplified} was a clinical trial conducted between 1992 and 2001 across 11 retinal specialty clinics in the United States. The primary objective was to study the risk factors for age-related macular degeneration (AMD) and the impact of dietary supplements on AMD progression. The study followed 4,757 participants, aged 55–80 at enrollment, for a median of 6.5 years. Participants were selected with a broad range of AMD severity, from no AMD to late-stage AMD in one eye. At each visit, certified technicians captured color fundus photography images using a standardized imaging protocol, although adherence to the protocol varied, leading to visits at irregular intervals. AMD severity scores were determined by expert graders at the University of Wisconsin Fundus Photograph Reading Center, with late AMD defined as the presence of neovascular AMD or atrophic AMD with geographic atrophy (severity scores from 10 to 12 using severity scale~\cite{davis2005age}). For this study, we include images from both the left and right eyes of each participant and make predictions separately for each eye. In cases where multiple images were captured per eye during a visit, we select only one image for analysis. Demographic information, including age, sex, and race, is available in the dataset and is used as sensitive attributes in our study.
\subsubsection{TTE Outcome Construction}
In our study, TTE outcomes for fundus images are defined as the duration, in years, from the date an image was captured to the first recorded diagnosis of late AMD in the corresponding eye. For eyes without a recorded diagnosis of late AMD, the censoring dates are set to the time of the last imaging visit. To ensure that the model forecasts the future risk of developing late AMD, all images taken during the final visit for a given eye were excluded from the dataset, as this visit was used solely to determine the TTE outcome. By removing the final visit, we ensured that no images were included with late AMD already present at the time of acquisition. This process resulted in a final dataset of 129,708 fundus images, each paired with corresponding TTE information, enabling a robust analysis of the TTE prediction for AMD progression.
\subsubsection{Data Access}
AREDS is a publicly available dataset hosted in the National Center for Biotechnology Information (NCBI) database of Genotypes and Phenotypes (dbGAP) through controlled access. Researchers can request access to this dataset at dbGAP. Once the application is approved, researchers can access data at the following address: \url{https://www.ncbi.nlm.nih.gov/projects/gap/cgi-bin/study.cgi?study_id=phs000001.v3.p1}

\subsection{MIMIC-CXR}
\subsubsection{Dataset Description}
The MIMIC-CXR dataset~\cite{johnson2019mimic} is a comprehensive, publicly available collection of chest X-ray images, along with associated clinical data, from the larger MIMIC-IV~\cite{johnson2023mimic} database. It includes over 370,000 chest X-ray images from more than 65,000 patients, annotated with both structured and unstructured clinical information, such as patient demographics, diagnoses, and other relevant clinical details. For our study, we utilize the MIMIC-CXR-JPG~\cite{johnson2019mimic}, a processed version of the MIMIC-CXR dataset, which provides images in JPG format derived from the original DICOM files. Additionally, we link this dataset to the broader MIMIC database to access patient demographic information, including age, sex, and race, which we incorporate as sensitive attributes in our fairness benchmarking framework.
\subsubsection{TTE outcome construction}
To construct TTE outcomes for chest X-ray images, we extract in-hospital mortality events from the MIMIC-IV database. For patients without a recorded date of mortality, the censoring dates are set as 1 year after their last recorded discharge date. We exclude any images that do not have a matching record in the MIMIC-IV patient table, images taken after the latest discharge date, and images taken after a recorded date of mortality. TTE is then calculated as the number of days from the image study date to either the date of mortality or the censoring date. This process results in a final dataset of 269,360 chest X-ray images, each paired with corresponding TTE information, enabling a robust analysis of the TTE prediction for in-hospital mortality.
\subsubsection{Data Access}
MIMIC-CXR, MIMIC-CXR-JPG, and MIMIC-IV are a publicly available datasets hosted by PhysioNet, which is a platform providing access to medical data. To access the dataset, researchers first need to create an account on PhysioNet and then complete the required training (CITI training). Once researchers have completed the CITI training, they will need to request access to the dataset at the following address:
\begin{itemize}
    \item MIMIC-IV: \url{https://physionet.org/content/mimiciv/3.1/}
    \item MIMIC-CXR-JPG: \url{https://physionet.org/content/mimic-cxr-jpg/2.1.0/}
\end{itemize}

\subsection{ADNI}
\subsubsection{Dataset Description}
The Alzheimer's Disease Neuroimaging Initiative (ADNI)~\cite{petersen2010alzheimer} is a large, longitudinal study aimed at identifying biomarkers for Alzheimer's disease (AD) and tracking the progression of the disease over time. Launched in 2004, ADNI is one of the most comprehensive datasets for studying Alzheimer's disease and other neurodegenerative disorders. It includes a wide range of data types, including clinical assessments, neuroimaging (MRI, PET), genetic data, and fluid biomarkers (e.g., cerebrospinal fluid and blood samples) from over 1,700 participants. These participants are categorized into different diagnostic groups, including cognitively normal individuals, those with mild cognitive impairment (MCI), and individuals with Alzheimer's disease. For our study, we focus on neuroimaging data, specifically MRI scans, and consider age and sex as sensitive attributes in our analysis.
\subsubsection{TTE outcome construction}
In our study, TTE outcomes for brain MRI scans are defined as the duration, measured in 6-month intervals, from the date an MRI scan was captured to the first recorded diagnosis of Alzheimer's disease. For participants without a recorded diagnosis of Alzheimer's disease, censoring dates are set to the time of their last imaging visit. To ensure that the model predicts the future risk of developing Alzheimer's disease, we excluded all MRI scans taken during the final visit, as this visit was used solely for determining the TTE outcome. By removing the final visit, we ensured that no MRI scans were included for participants who had already been diagnosed with Alzheimer's disease at the time of acquisition. This process resulted in a final dataset of 2,227 brain MRI scans, each paired with corresponding TTE information, enabling a comprehensive analysis of TTE prediction for Alzheimer's disease progression.
\subsubsection{Data Access}
ADNI is a publicly available dataset hosted in the Image and Data Archive (IDA), a secure online resource for archiving, exploring and sharing neuroscience data. Access to the ADNI dataset requires that researchers register for an IDA account. Once the account is created and the ADNI Data Use Agreement is completed, they can access data at the following address: \url{https://adni.loni.usc.edu}
\section{Algorithms Details}\label{sec:d}

\subsection{TTE Prediction Models}
\subsubsection{PMF} PMF~\cite{kvamme2021continuous} is a discrete-time model designed for TTE prediction tasks. It represents the survival time PMF $P(t|x)$ using a neural network $f(\cdot; \theta) : \mathcal{X} \rightarrow \left[0, 1\right]^L$ where $\theta$ denotes the model parameters and $L$ represents the number of time intervals. The model parameters $\theta$ are estimated by minimizing the negative log-likelihood, averaged over the training data, as follows:
\begin{equation*}
    \mathcal{L}_{PMF}(\theta) = -\frac{1}{n} \sum_{i=1}^{n} \left\{ \delta_i \log \left( f_{\kappa(y_i)}(x_i; \theta) \right) 
+ (1 - \delta_i) \log \left( \sum_{m=\kappa(y_i)+1}^{L} f_m(x_i; \theta) \right) \right\}
\end{equation*}
where $n$ is the number of training samples, $\kappa(y_i)$ denotes the specific time interval (from $1, 2, \cdots, L$) that corresponds to time $y_i$, and $f_m$ represents the predicted probability that the TTE falls within the $m-th$ interval.
\subsubsection{DeepHit} DeepHit~\cite{lee2018deephit} extends the PMF model by incorporating a ranking loss function alongside the negative log-likelihood. Given a comparable set defined as $\mathcal{E} := \left\{(i,j)\in[n]\times[n]:\delta_i=1,y_i < y_j\right\}$, this ranking loss is calculated over the training data, as follows:
\begin{equation*}
    \mathcal{L}^{rank}_{DeepHit}(\theta) = \sum_{(i,j) \in \mathcal{E}} \exp \left( \frac{-\left(\sum_{m=1}^{\kappa(y_i)} f_m(x_i; \theta) - \sum_{m'=1}^{\kappa(y_j)} f_{m'}(x_j; \theta) \right)}{\sigma} \right)
\end{equation*}
where $n$ is the number of training samples, $\kappa(y_i)$ denotes the specific time interval (from $1, 2, \cdots, L$) that corresponds to time $y_i$, and $f_m$ represents the predicted probability that the TTE falls within the $m-th$ interval.
\subsubsection{Nnet-survival}
Nnet-survival~\cite{gensheimer2019scalable} is another discrete-time model designed for TTE prediction tasks. It represents the hazard function $h(\cdot|x)$ using a neural network $f(\cdot; \theta) : \mathcal{X} \rightarrow \mathbb{R}^L$ where $\theta$ denotes the model parameters and $L$ represents the number of time intervals. Specifically, Nnet-survival sets the hazard function equal to:
\begin{equation*}
    h[\ell | x; \theta] := \frac{1}{1 + e^{-f_{\ell}(x; \theta)}} \quad \text{for } \ell \in [L], x \in \mathcal{X}
\end{equation*}
where $f_m(x; \theta)$ is the $m-th$ output of the neural network. The model parameters $\theta$ are estimated by minimizing the negative log-likelihood, averaged over the training data, as follows:
\begin{align*}
    \mathcal{L}_{Nnet-survival}(\theta) = \frac{1}{n} \sum_{i=1}^{n} &\left\{ 
\delta_i \log \left( 1 + e^{-f_{\kappa(y_i)}(x_i; \theta)} \right)
+ (1 - \delta_i) \log \left( 1 + e^{f_{\kappa(y_i)}(x_i; \theta)} \right) \right.\\
&+ \left. \sum_{m=1}^{\kappa(y_i)-1} \log \left( 1 + e^{f_m(x_i; \theta)} \right) 
\right\}
\end{align*}
where $n$ is the number of training samples and $\kappa(y_i)$ denotes the specific time interval (from $1, 2, \cdots, L$) that corresponds to time $y_i$.

\subsection{Fairness Algorithms}

\subsubsection{Distributional Group Optimization}
Distributional Group Optimization~\cite{sagawa2019distributionally,hu2024fairness} is an optimization technique designed to enhance model robustness by minimizing the worst-case training loss across different groups. Instead of optimizing for the average performance, GroupDRO focuses on the group with the highest loss, ensuring that the model does not disproportionately underperform on any particular group. By incorporating increased regularization, this approach helps mitigate disparities in model performance across diverse subpopulations, making it particularly useful in settings where fairness and reliability across groups are critical.

\subsubsection{Subgroup Rebalancing}
This resampling method~\cite{kamiran2012data} addresses class imbalance by upsampling minority groups, ensuring that all groups have equal representation during training. By increasing the frequency of underrepresented samples, the model is exposed to a more balanced dataset, reducing bias and improving fairness. This approach helps prevent the model from being overly influenced by the majority group, leading to more equitable predictions across all groups.

\subsubsection{Fair Representation Learning}
A common approach to promoting fairness in machine learning is through fair representation learning, which aims to obfuscate sensitive group membership information in the learned representations. By ensuring that the model's latent features do not encode discriminatory patterns, this method helps mitigate bias in downstream predictions. Following \cite{wang2021understanding}, we incorporate fairness constraints into representation learning by leveraging kernel-based distribution matching via Maximum Mean Discrepancy. This technique enforces similarity in feature distributions across different groups, reducing disparities while preserving task-relevant information.

\subsubsection{Domain Independence}
The Domain Independence~\cite{wang2020towards} is a method that trains separate classifiers for different groups while utilizing a shared encoder. This approach allows the model to capture group-specific patterns through distinct classifiers while maintaining a common feature representation across all groups. By leveraging a shared encoder, DomainInd enhances generalization and reduces the risk of overfitting to individual groups, ultimately improving the model’s robustness and fairness in diverse domains.

\subsubsection{Controlled for Sensitive Attribute}
This approach~\cite{zhao2023fairness,pope2011implementing}, similar to Domain Independence, involves training separate models for each group based on the values of a sensitive attribute. By doing so, the method captures group-specific patterns while maintaining model flexibility. During the prediction phase, the outputs of the fitted models are averaged across all groups in the population. This averaging process ensures that no single group disproportionately influences the predictions, promoting fairness and reducing bias in the model's outcomes.

% \subsection{Image Backbones}

% \subsubsection{EfficientNet}

% \subsubsection{ResNet3D}

\subsection{Model Architecture Details}
Each (fair) TTE prediction model comprises two key networks: an image encoder and a classifier. The image encoder transforms images into representation vectors, while the classifier predicts survival time intervals based on these learned representations. In our study, we employ a 2D EfficientNet~\cite{tan2019efficientnet} backbone as the image encoder for the AREDS and MIMIC-CXR datasets, and a 3D ResNet-18 backbone~\cite{tran2018closer} for the ADNI dataset. To adapt these models for our task, we replace the original fully connected layers with new layers that map images into the representation space. The architectural details of our 2D and 3D image encoders, along with the classifier, are presented in Table~\ref{tab:a2}.

\begin{table}[H]
\caption{Architecture details of (fair) TTE prediction models. In our experiments, we set \textbf{n\_channel} = 3 for both 2D and 3D images by duplicating grayscale chest X-ray and brain MRI images to obtain three-channel inputs. \textbf{feature\_dim} is set to 64, and \textbf{hidden\_dim} is set to 16. \textbf{n\_class} is determined using the 'equidistant' discretization method, with values of 14 for AREDS, 28 for ADNI, and 128 for MIMIC-CXR.  \label{tab:a2}}
\vskip 0.1in
\centering
\resizebox{\textwidth}{!}{
\begin{tabular}{l|l|l}
\hline
Networks & \multicolumn{2}{l}{Layers} \\ \hline
\multirow{2}{*}{Image Encoder} & 2D image & \begin{tabular}[c|]{@{}l@{}}
Conv2d(input channel = \textbf{n\_channel}, output channel = 32, kernel = 3)\\ 
MBConv1(input channel = 32, output channel = 16, kernel = 3)\\ 
MBConv6(input channel = 16, output channel = 24, kernel = 3) * 2\\ 
MBConv6(input channel = 24, output channel = 40, kernel = 5) * 2\\ 
MBConv6(input channel = 40, output channel = 80, kernel = 3) * 3\\ 
MBConv6(input channel = 80, output channel = 112, kernel = 5) * 3\\ 
MBConv6(input channel = 112, output channel = 192, kernel = 5) * 4\\ 
MBConv6(input channel = 192, output channel = 320, kernel = 3) \\ 
Conv2d(input channel = 320, output channel = 1280, kernel = 1)\\ 
AdaptiveAvgPool2d(output\_size = 1)\\ 
Linear(input dim = 1280, output dim = feature\_dim)\\ 
Dropout(p=0.5)
\end{tabular} \\ \cline{2-3} 
 & 3D image & \begin{tabular}[c]{@{}l@{}}
 Conv3d(input channel = \textbf{n\_channel}, output channel = 64, kernel = 3$\times$7$\times$7)\\ 
Conv3d(input channel = 64, output channel = 64, kernel = 3$\times$3$\times$3) * 4\\ 
Conv3d(input channel = 64, output channel = 128, kernel = 3$\times$3$\times$3)* 4\\ 
Conv3d(input channel = 128, output channel = 256, kernel = 3$\times$3$\times$3)* 4\\ 
Conv3d(input channel = 256, output channel = 512, kernel = 3$\times$3$\times$3)* 4\\ 
AdaptiveAvgPool3d(output\_size=(1,1,1))\\
Linear(input dim = 512, output dim = \textbf{feature\_dim})\\ 
Dropout(p=0.5)
 \end{tabular} \\ \hline
Classifier & \multicolumn{2}{l}{\begin{tabular}[c]{@{}l@{}}
Linear(input dim = \textbf{feature\_dim}, output dim = \textbf{hidden\_dim})\\ 
Linear(input dim = \textbf{hidden\_dim}, output dim = \textbf{n\_classes})
\end{tabular}} \\ \hline
\end{tabular}}
\end{table}
\section{Evaluation Metrics}\label{sec:e}

\subsection{Performance Metrics}
The performance metrics used in our study are defined based on~\cite{chen2024introduction}.
\subsubsection{Time-dependent concordance index}
Harrell's concordance index (C-index) is one of the most widely used accuracy metrics in TTE prediction. It quantifies the fraction of data point pairs that are correctly ranked by a prediction model among those that can be unambiguously ranked. The C-index values range from 0 to 1, with 1 indicating perfect ranking accuracy. However, a notable limitation of the C-index is its dependence on a risk score function for ranking, which many TTE prediction models do not explicitly learn. To address this limitation,~\cite{antolini2005time} introduced a time-dependent concordance index ($C^{td}$), which leverages predicted survival functions, $\hat{S}(\cdot|x)$, to assess model performance more effectively. The $C^{td}$ is computed as follows.

\begin{definition}
    Suppose that we have a survival function estimate $\hat{S}(\cdot|x)$ for any $x \in \mathcal{X}$ . Then using the set of comparable pairs $\mathcal{E} := \left\{(i,j)\in[n]\times[n]:\delta_i=1,y_i < y_j\right\}$, we define the $C^{td}$ metric as:
    \begin{align*}
        C^{td} := \frac{1}{\mathcal{E}} \sum_{(i,j) \in \mathcal{E}} \mathbbm{1} \left\{\hat{S}(y_i|x_i) < \hat{S}(y_i|x_j)\right\}
    \end{align*}
    which is between 0 and 1. Higher scores are better.
\end{definition}

\subsubsection{Time-dependent AUC}
While the C-index and $C^{td}$ scores provide valuable single-number summaries of predictive accuracy in TTE prediction, they lack the ability to evaluate accuracy at a specific user-defined time, $t$. To address this limitation, \cite{uno2007evaluating} introduced time-dependent AUC scores ($AUC^{td}$), which explicitly depend on the chosen time point $t$. The core idea behind $AUC^{td}$ is to frame a binary classification problem for a fixed time $t$, where the “positive” class consists of data points that experienced the event no later than $t$, and the “negative” class includes those that survived beyond $t$. The survival function $\hat{S}(t|\cdot): \mathcal{X} \rightarrow [0, 1]$ serves as the probabilistic classifier, predicting survival probabilities. A lower predicted survival probability for a given point $x$ implies a higher likelihood of belonging to the positive class. The $AUC^{td}$ score quantifies the classifier's ability to distinguish between these two classes at time $t$, offering a time-specific accuracy assessment of the model's predictions. The $AUC^{td}$ is computed as follows.

\begin{definition}
    Suppose that we have a survival function estimate $\hat{S}(\cdot|x)$ for any $x \in \mathcal{X}$ . Then for any $t>0$, using the set of comparable pairs $\mathcal{E}(t) := \left\{(i,j)\in[n]\times[n]:\delta_i=1,y_i \leq t, y_j > t\right\}$, we define the $AUC^{td}(t)$ (the $AUC^{td}$ at time $t$) as:
    \begin{align*}
        AUC^{td}(t) := \frac{\sum_{(i,j) \in \mathcal{E}(t)}w_i \mathbbm{1} \left\{\hat{S}(t|x_i) < \hat{S}(t|x_j)\right\}}{\sum_{(i,j) \in \mathcal{E}(t)}w_i}
    \end{align*}
    where $w_1,w_2,\cdots,w_n \in [0,\infty)$ are inverse probability of censoring weights to be defined as $w_i := 1/(\hat{S}_{censor}(y_i) \hat{S}_{censor}(t))$, and $\hat{S}_{censor}(t)$ is an estimation of $S_{censor}(t) := P(C > t)$ using Kaplan-Meier estimator~\cite{kaplan1958nonparametric}. $AUC^{td}(t)$ is between 0 and 1 and higher scores are better. Finally, we can get $AUC^{td}$ as
    \begin{align*}
        AUC^{td} := \frac{1}{t_{\max} - t_{\min}} \int_{t_{\min}}^{t_{\max}} AUC^{td}(u) du
    \end{align*}
where $t_{\min}$ and $t_{\min}$ are user-specified lower and upper limits of integration.
\end{definition}

\subsubsection{Integrated Brier Score}
The Integrated Brier Score ($IBS$) is a performance metric that directly evaluates the error of an estimated survival function $\hat{S}(\cdot|x)$ without relying on ranking. The $IBS$ is calculated as follows.
\begin{definition}
    Suppose that we have a survival function estimate $\hat{S}(\cdot|x)$ for any $x \in \mathcal{X}$ . Then for any $t>0$, we define the $BS(t)$ (the $IBS$ at time $t$) as:
    \begin{align*}
        BS(t) := \frac{1}{N} \sum_{i=1}^{n} \left( \frac{\hat{S}(t|x_i)^2 \delta_i \mathbbm{1} \left\{ y_i \leq t \right\}}{\hat{S}_{censor}(y_i)} + \frac{\left(1-\hat{S}(t|x_i)^2\right) \mathbbm{1} \left\{ y_i > t \right\}}{\hat{S}_{censor}(t)} \right)
    \end{align*}
    which is nonnegative. Lower scores are better. Finally, we can get $IBS$ as
    \begin{align*}
        IBS := \frac{1}{t_{\max} - t_{\min}} \int_{t_{\min}}^{t_{\max}} BS(u) du
    \end{align*}
where $t_{\min}$ and $t_{\min}$ are user-specified lower and upper limits of integration.
\end{definition}

\subsection{Fairness Metrics}\label{sec:f}

In this study, we define fairness metrics as the predictive performance gaps between groups. This kind of metric is used ensure that the model maintains equal predictive performance across different groups. In particular, given a performance metric $\Er$ for TTE prediction task, we define fairness metric $\mathcal{F}_{\Er}$ as follow:
\begin{align*}
    \mathcal{F}_{\Er}(h) = \max_{a, a' \in \mathcal{A}} \left | \Er_a - \Er_{a'} \right |
\end{align*}
where $\mathcal{A}$ is the set of groups considered in TTE prediction task, $\Er_a$ and $\Er_{a'}$ are the
predictive performance metrics calculated from subsets containing data from groups $a$ and $a'$, respectively. For each predictive performance metric defined above, we have a corresponding fairness metric as follows.
\begin{align*}
    \mathcal{F}_{C^{td}} &= \max_{a, a' \in \mathcal{A}} \left | C^{td}_a - C^{td}_{a'} \right | \\
    \mathcal{F}_{AUC^{td}} &= \max_{a, a' \in \mathcal{A}} \left | AUC^{td}_a - AUC^{td}_{a'} \right | \\
    \mathcal{F}_{IBS} &= \max_{a, a' \in \mathcal{A}} \left | IBS_a - IBS_{a'} \right |
\end{align*}
where $C^{td}_a$, $AUC^{td}_a$, $IBS_a$ are predictive performance metrics calculated from the subset containing data from group $a$, and $C^{td}_{a'}$, $AUC^{td}_{a'}$, $IBS_{a'}$ are predictive performance metrics calculated from the subset containing data from group $a'$.

\subsection{Fairness-Utility Trade-Off Metric}

The fairness metrics mentioned above do not capture the fairness-utility trade-off while in medical context, it is essential to balance fairness and utility to ensure that the model is not only fair but also accurate and effective for all groups. To handle this issue, we leverage the equity-scaling metric ($ES$)~\cite{luo2024harvard} that takes both utility and fairness into account for evaluation. Similar to fairness metric, for each predictive performance metric, we have a corresponding fairness-utility trade-off metric as follows.
\begin{align*}
    ES_{C^{td}} &= \frac{C^{td}_D}{1 + \sum_{a \in \mathcal{A}} \left| C^{td}_{D} - C^{td}_{D_a} \right|} \\
    ES_{AUC^{td}} &= \frac{AUC^{td}_D}{1 + \sum_{a \in \mathcal{A}} \left| AUC^{td}_{D} - AUC^{td}_{D_a} \right|} \\
    ES_{IBS} &= \frac{1 - IBS_D}{1 + \sum_{a \in \mathcal{A}} \left| IBS_{D} - IBS_{D_a} \right|}
\end{align*}
The advantage of the equity-scaling metric lies in its intuitive interpretability. Specifically, a higher equity-scaling score indicates that the model is both more accurate and more equitable simultaneously.
\section{Experimental Setup Details}

\subsection{Implementation Details}

\subsubsection{Hardware Usage}
The experiments were conducted at a supercomputing center utilizing multiple compute nodes. Each node was equipped with an NVIDIA Volta V100 GPU with 16 GB of memory, an Intel Xeon CPU, and 32 GB of RAM, ensuring the computational resources necessary for large-scale experiments. In total, we trained over 20,000 models, requiring approximately 4.56 GPU years of computational effort, highlighting the extensive scale of our study.

\subsubsection{Package Usage}
The FairTTE benchmark is implemented using Python 3, with PyTorch~\cite{paszke2019pytorch} serving as the framework for deep learning computations. The implementation of TTE models is built on the pycox~\cite{kvamme2019time} package, while the evaluation metrics for TTE prediction leverage pycox, scikit-survival~\cite{sksurv}, and SurvivalEVAL~\cite{qi2024survivaleval}. Additionally, the training and evaluation pipeline for TTE prediction models is adapted from the demo code provided in~\cite{chen2024introduction}, ensuring a robust and standardized framework for benchmarking.

\subsection{Data Split and Pre-processing}\label{sec:f2}
\paragraph{Data Split.} 
Each dataset in our study was divided into training, validation, and testing sets using a 60\%:20\%:20\% split ratio. Models were trained on the training sets, evaluated on the testing sets, and the validation sets were used for model selection. Since a single patient may have multiple medical records, we took precautions to prevent data leakage during model training. Specifically, the data was split by patient, ensuring that no patient appearing in the testing set had any records in the training or validation sets. This approach maintains the integrity of the evaluation process and ensures that model performance is assessed on entirely unseen patient data.
\paragraph{Data Pre-processing.}
Before being fed into the TTE prediction models, chest X-ray and color fundus images are resized to $224 \times 224$ pixels, while brain MRI scans are resized to $128 \times 128 \times 96$. Additionally, all pixel values are normalized to a range of 0 to 1 to ensure stability during training and improve model performance. 

We consider binary group setting in our experiment. These groups were constructed according to the following criteria:
\begin{itemize}
    \item Race: 'Non-White' (Group 0), 'White' (Group 1)
    \item Sex: 'Female' (Group 0), 'Male' (Group 1)
    \item Age: 
    \begin{itemize}
        \item MIMIC-CXR: '$\leq 60$' (Group 0), '$> 60$' (Group 1)
        \item AREDS: '$\leq 70$' (Group 0), '$> 70$' (Group 1)
        \item ADNI: '$\leq 80$' (Group 0), '$> 80$' (Group 1)
    \end{itemize}
\end{itemize}

\subsection{Hyperparameter Search}
To ensure a fair comparison, we perform a grid-based hyperparameter search using 10 random seeds. The details of the hyperparameter search for the methods used in our experiments are provided below. 

\begin{itemize}
    \item TTE prediction models
    \begin{itemize}
        \item Learning rate: $10^x$ where $x \sim Uniform(-4,-3)$
        \item Decay rate: $10^x$ where $x \sim Uniform(-6,-4)$
    \end{itemize}
    \item Fair TTE prediction models
    \begin{itemize}
        \item $\eta: 10^x$ where $x \sim Uniform(-3,-1)$ (DRO)
        \item $\lambda: 10^x$ where $x \sim Uniform(-5,2)$ (FRL)
    \end{itemize}
\end{itemize}

For standard TTE prediction models, we select the best models based on their predictive performance metrics calculated on the validation sets. In contrast, for fair TTE prediction models, we prioritize fairness metrics when selecting the best models, allowing for up to a 5\% reduction in accuracy compared to the baseline TTE models. This approach ensures a balanced trade-off between fairness and predictive performance.

\subsection{Quantifying Source of Bias}\label{sec:f4}
In order to quantify the degree of bias sources in each dataset and sensitive attribute setting, we use several metrics as follows.

\paragraph{Disparity in mutual information between $X_Z$ and $Y$ across groups.}
We quantify the disparity in mutual information between $X_Z$ (i.e., image representation generated from the vision backbones) and $Y$ across groups by computing the maximum difference in their normalized mutual information values across all groups, as defined below.
\begin{align*}
    Bias_{MI(X_Z,Y)} = \max_{a,a' \in \mathcal{A}} &\left | \frac{2 I(X_Z, Y|A=a,\Delta=1)}{H(X_Z|A=a,\Delta=1) + H(Y|A=a,\Delta=1)} \right. \\
    &- \left. \frac{2 I(X_Z, Y|A=a',\Delta=1)}{H(X_Z|A=a',\Delta=1) + H(Y|A=a',\Delta=1)} \right |
\end{align*}
where $I(\cdot,\cdot|A=a,\Delta=1)$ represented the mutual information conditioned on $A=a$ and $\Delta=1$ and $H(\cdot|A=a)$ denotes the entropy conditioned on $A=a$ and $\Delta=1$.

\paragraph{Disparity in mutual information between $X_Z$ and $\Delta$ across groups.}
Similarly, we quantify the disparity in mutual information between $X_Z$ (i.e., image representation generated from the vision backbones) and $\Delta$ across groups by computing the maximum difference in their normalized mutual information values across all groups, as defined below.
\begin{equation*}
    Bias_{MI(X_Z,\Delta)} = \max_{a,a' \in \mathcal{A}} \left | \frac{2 I(X_Z, \Delta|A=a)}{H(X_Z|A=a) + H(\Delta|A=a)} - \frac{2 I(X_Z, \Delta|A=a')}{H(X_Z|A=a') + H(\Delta|A=a')} \right |
\end{equation*}

\paragraph{Disparity in TTE distribution across groups.} We measure the disparity in TTE distributions across groups by calculating the maximum Wasserstein distance~\cite{thorpe2018introduction}, normalized by the range of TTE, between the TTE distributions of each group. This is defined as follows:
\begin{equation*}
    Bias_{TTE} = \max_{a,a' \in \mathcal{A}} \left | \frac{\mathcal{W}\left( P(Y|A=a,\Delta=1) , P(Y|A=a',\Delta=1) \right)}{\max_{y \in \mathcal{Y}}y}  \right |
\end{equation*}
where $\mathcal{W}\left(\cdot,\cdot \right)$ denotes the Wasserstein-1 distance between the two distributions.

\paragraph{Disparity in image distribution across groups.} We measure the disparity in image distributions across groups by calculating the maximum Wasserstein distance~\cite{thorpe2018introduction}, normalized by the range of image feature values, between the image distributions of each group. This is defined as follows:
\begin{equation*}
    Bias_{Image} = \max_{a,a' \in \mathcal{A}} \left | \frac{\mathcal{W}\left( P(X_Z|A=a) , P(X_Z|A=a') \right)}{\max_{y \in \mathcal{Y}}y}  \right |
\end{equation*}
where $\mathcal{W}\left(\cdot,\cdot \right)$ denotes the Wasserstein-1 distance between the two distributions. Due to the high dimensionality of image representations, we implement sliced Wasserstein distance~\cite{bonneel2015sliced}, a variant of the Wasserstein distance that approximates the full Wasserstein distance between high-dimensional distributions by projecting them onto one-dimensional subspaces and averaging the resulting 1D Wasserstein distances.

\paragraph{Disparity in censoring rate across groups.} We quantify the disparity in censoring rates across groups by calculating the maximum normalized difference between the means of the censoring distributions for each group, as defined below.
\begin{equation*}
    Bias_{Censoring} = \max_{a,a' \in \mathcal{A}} \left | \frac{\mathbb{E}\left[ \Delta | A = a \right] - \mathbb{E}\left[ \Delta | A = a' \right]}{\mathbb{E}\left[ \Delta \right]}  \right |
\end{equation*}

\subsection{Constructing Causal Distribution Shift}\label{sec:f5}
To construct distribution shift between training and testing data, we modify the training data by introducing correlations between the sensitive attribute and other RVs in the causal graph (Figure~\ref{fig:2}). This adjustment simulates real-world scenarios where biases in data collection or underlying relationships may lead to disparities across groups. The details of this process, including the specific modifications applied to establish these correlations, are outlined below.
\begin{itemize}
    \item \textbf{Distribution shift on $X$:} Images from disadvantaged groups are degraded using a Gaussian blur filter to simulate lower-quality data.
    \item \textbf{Distribution shift on $Y$:} TTE labels for disadvantaged groups are corrupted by adding noise sampled from a uniform distribution.
    \item \textbf{Distribution shift on $\Delta$:} To simulate biased censoring, we flip the censoring indicators for 90\% of uncensored samples within disadvantaged groups.
\end{itemize}

We note that, although these distribution shifts are synthetic, they mimic real-world scenarios, as described below.
\begin{itemize}
    \item Adding noise to images mimics real-world scenarios such as patients in different geographic locations are scans with different equipment. This causes the medical image to appear systematically different for groups in each location.
    \item Adding noise to TTE labels mimics real-world scenarios such as delayed or inaccurate event recording in EHR system.
    \item Flipping censoring indices mimics real-world scenarios in which certain groups experience less consistent access to care due to financial or geographic. Thus, these groups are more likely to drop out of care, resulting in a higher censoring rate.
\end{itemize}

\section{Causal Graphs for Fairness in TTE Prediction}
\subsection{Causal Graphs for Biased and Unbiased Settings}\label{sec:g1}
\begin{figure}[H]
\centering
\includegraphics[width=\textwidth]{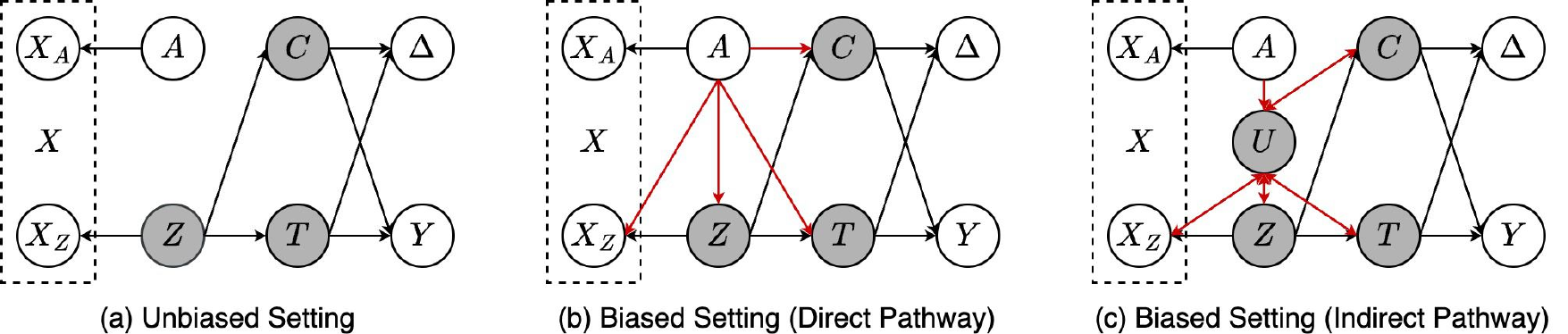}
    \caption{Causal structure in TTE prediction. Gray circles denote unobserved random variables. (a) Unbiased setting, where the sensitive attribute $A$ influences only $X_A$. (b) Biased setting with direct causal pathways, where $A$ is directly associated (red arrows) with other variables in the graph. (c) Biased setting with indirect causal pathways, where $A$ influences (red arrows) other variables through unobserved variables $U$.}\label{fig:s1}
\end{figure}

Figure~\ref{fig:s1} presents the causal graphs for the unbiased and biased scenarios. In the unbiased scenario (Figure~\ref{fig:s1}a), the sensitive attribute $A$ is unrelated to the TTE outcome and influences only $X_A$, with no effect on other variables in the graph. In contrast, in the biased scenarios (Figure~\ref{fig:s1}b and Figure~\ref{fig:s1}c), $A$ also affects additional variables, resulting in dependencies between $A$ and the TTE outcome. These causal pathways may be direct (Figure~\ref{fig:s1}b), mediated through unobserved variables $U$ (Figure~\ref{fig:s1}c), or both.

\subsection{Real-world Causal Graph Examples for Fairness in TTE Prediction}\label{sec:g2}
In this section, we present causal graphs illustrating real-world scenarios in time-to-event (TTE) prediction using medical imaging. Many of these examples are adapted from diagnostic settings in prior work~\cite{jones2024causal}. We describe four scenarios in which the sensitive attribute $A$ influences other variables in the causal graph—namely, the medical image $X$, the underlying condition $Z$, the time-to-event $T$, and the censoring time $C$—leading to disparities in group-specific data distributions. For each scenario, we include two examples: one where the causal pathway from $A$ is valid (\textcolor{red}{red arrows}), appearing in both training and testing data, and one where the pathway is spurious (\textcolor{blue}{blue arrows}), representing bias present only in the training data and absent in the testing data.

\begin{figure}[H]
\centering
\includegraphics[width=\textwidth]{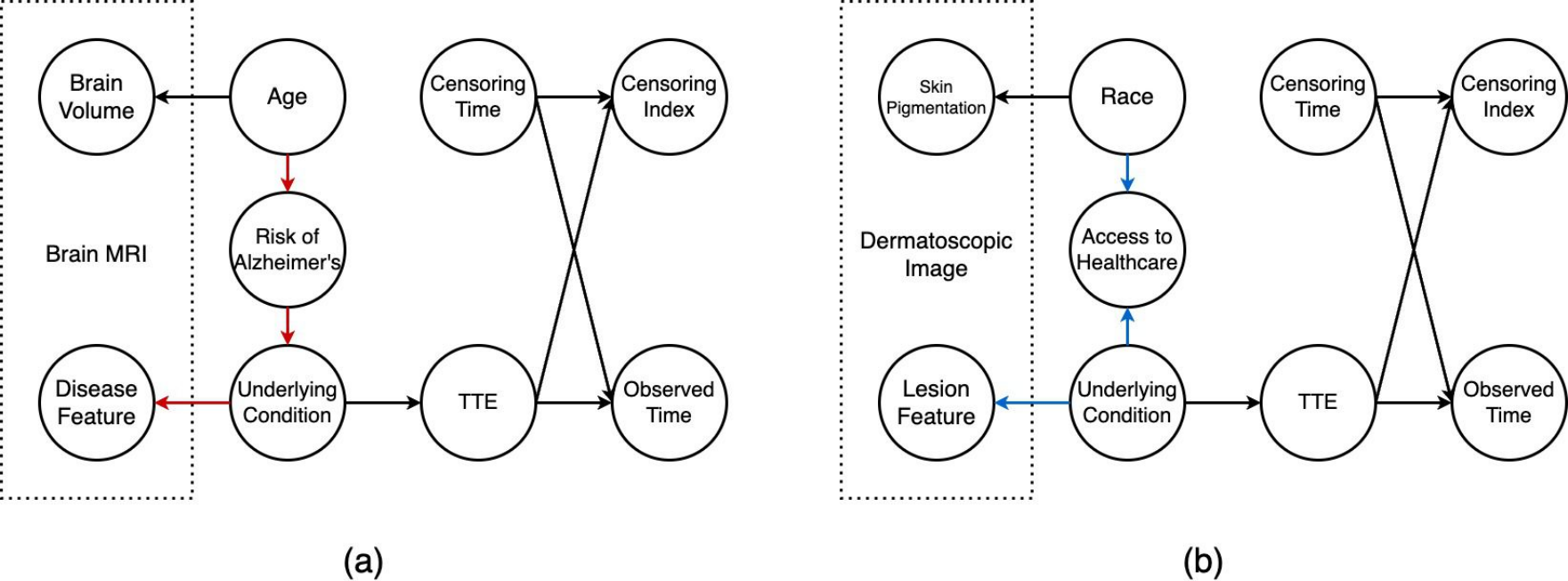}
    \caption{Causal graphs illustrating scenarios where the sensitive attribute $A$ affects the underlying condition $Z$. (a) Valid pathway: age is a known clinical risk factor for Alzheimer’s disease. (b) Invalid pathway: race appears spuriously correlated with $Z$ due to disparities in healthcare access.}\label{fig:s2}
\end{figure}

% \newpage

\begin{figure}[H]
\centering
\includegraphics[width=\textwidth]{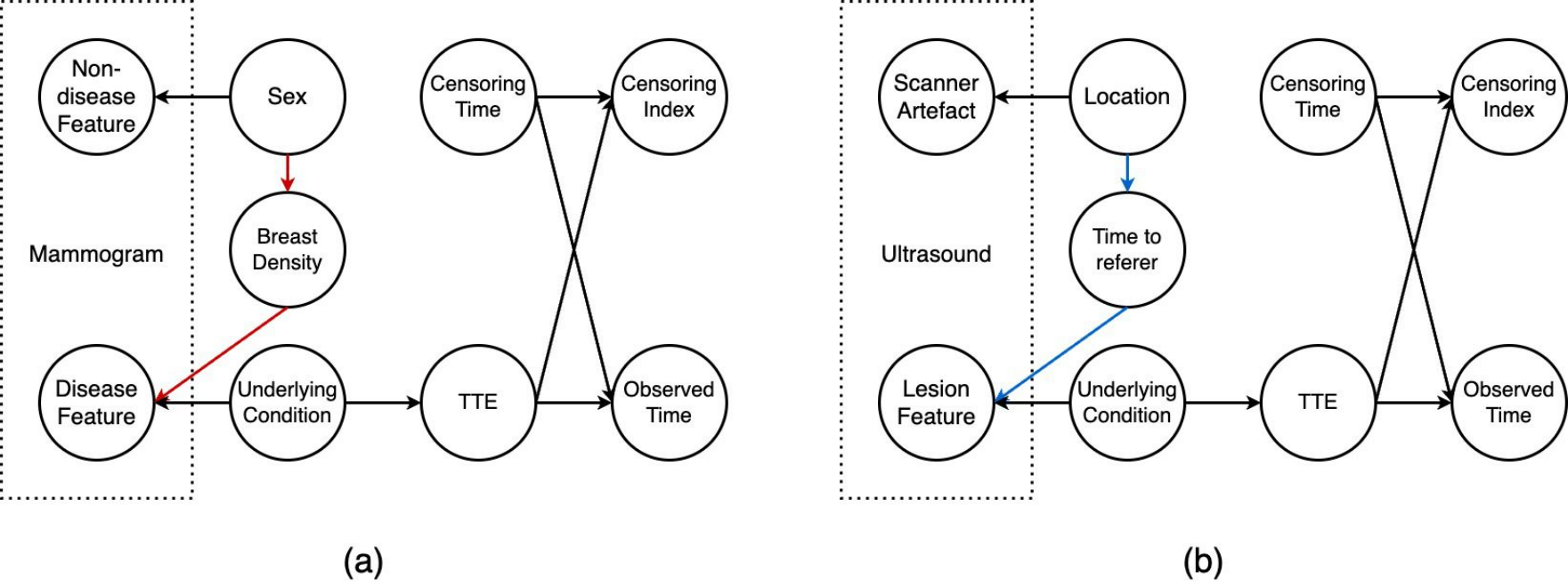}
    \caption{Causal graphs illustrating scenarios where the sensitive attribute $A$ influences the medical image $X$. (a) Valid pathway: breast cancer presents differently in men and women due to inherent differences in breast tissue. (b) Invalid pathway: spurious correlation arises when patients in different locations are imaged at varying disease stages due to inconsistent ultrasound referral policies.}\label{fig:s3}
\end{figure}

\begin{figure}[H]
\centering
\includegraphics[width=\textwidth]{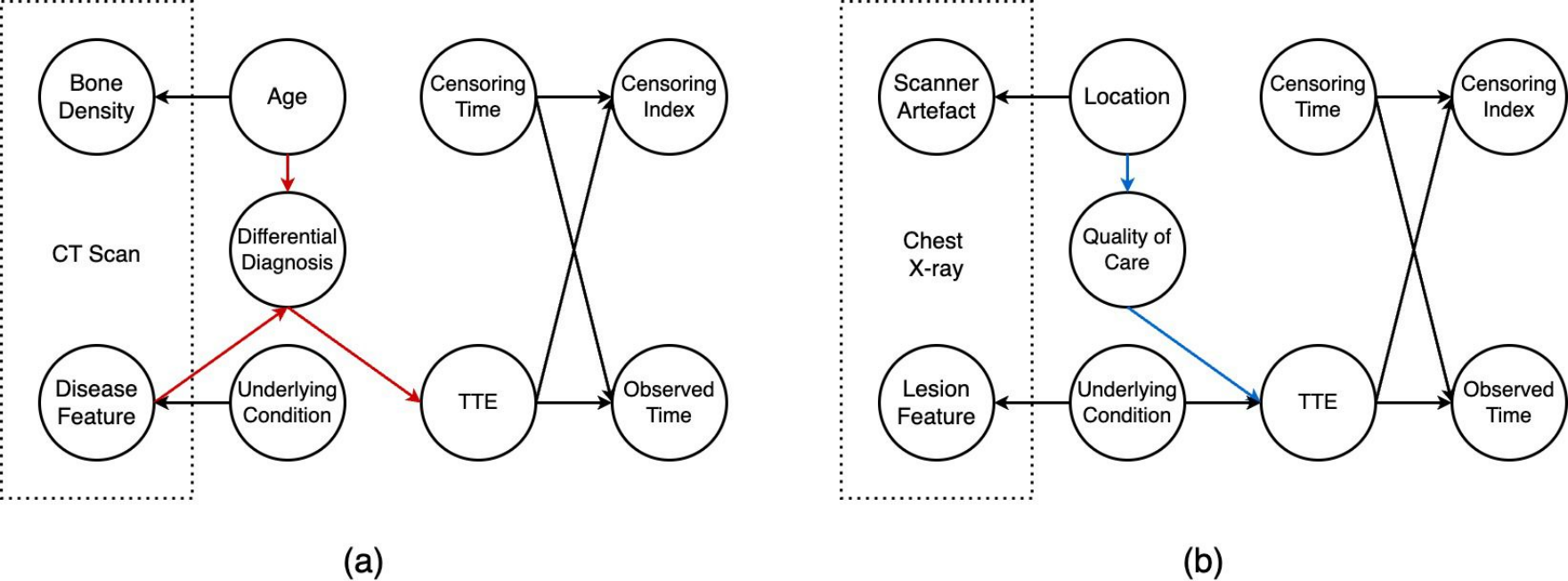}
    \caption{Causal graphs illustrating scenarios where the sensitive attribute $A$ influences the time-to-event outcome $T$. (a) Valid pathway: age contributes to differential diagnosis in epidemiology and legitimately affects disease progression. (b) Invalid pathway: a spurious correlation arises when patients from different locations receive healthcare services of varying quality, impacting $T$ in a non-causal manner.}\label{fig:s4}
\end{figure}

\begin{figure}[H]
\centering
\includegraphics[width=\textwidth]{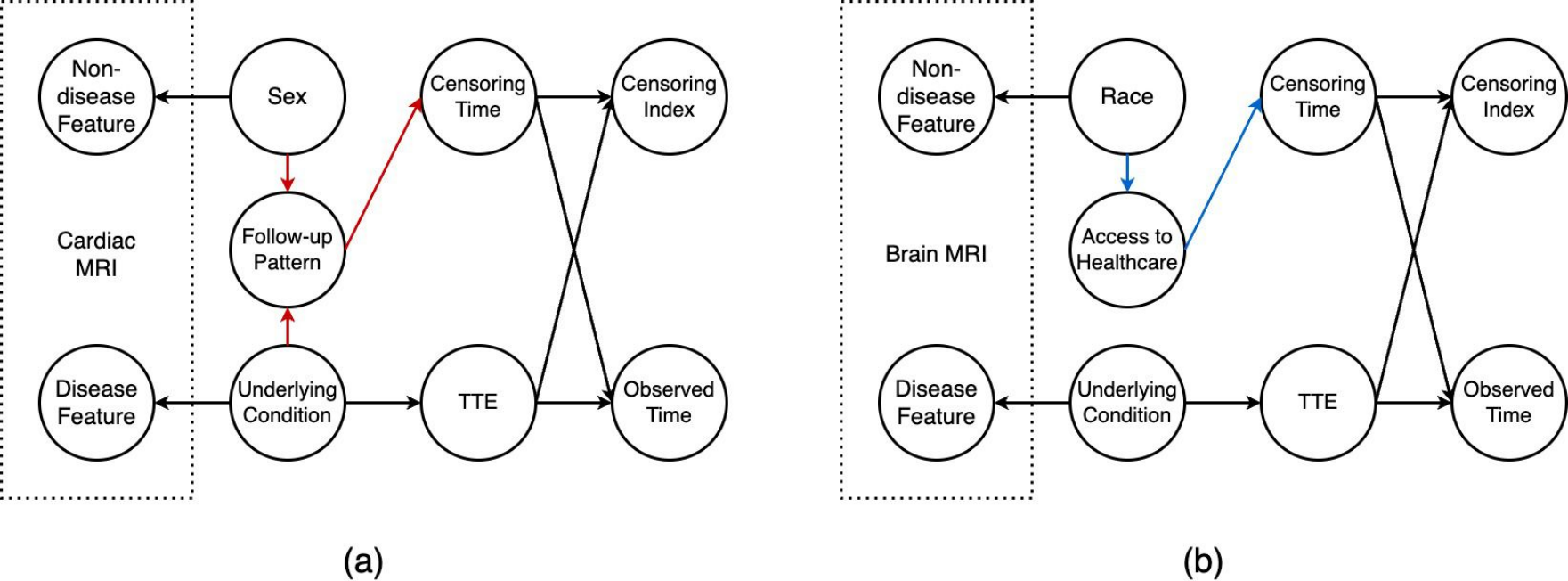}
    \caption{Causal graphs illustrating scenarios where the sensitive attribute $A$ affects the censoring time $C$. (a) Valid pathway: Women may be less likely to receive aggressive follow-up or diagnostic imaging for cardiac conditions, resulting in higher censoring for female patients. (b) Invalid pathway: race appears spuriously correlated with censoring time due to disparities in healthcare access.}\label{fig:s5}
\end{figure}

\newpage
\section{Additional Results}\label{sec:g}

\subsection{Predictive Performance and Fairness in TTE Prediction Models}

\begin{figure}[H]
    \centering
    \begin{subfigure}[t]{\linewidth}
        \centering
        \includegraphics[width=0.97\textwidth]{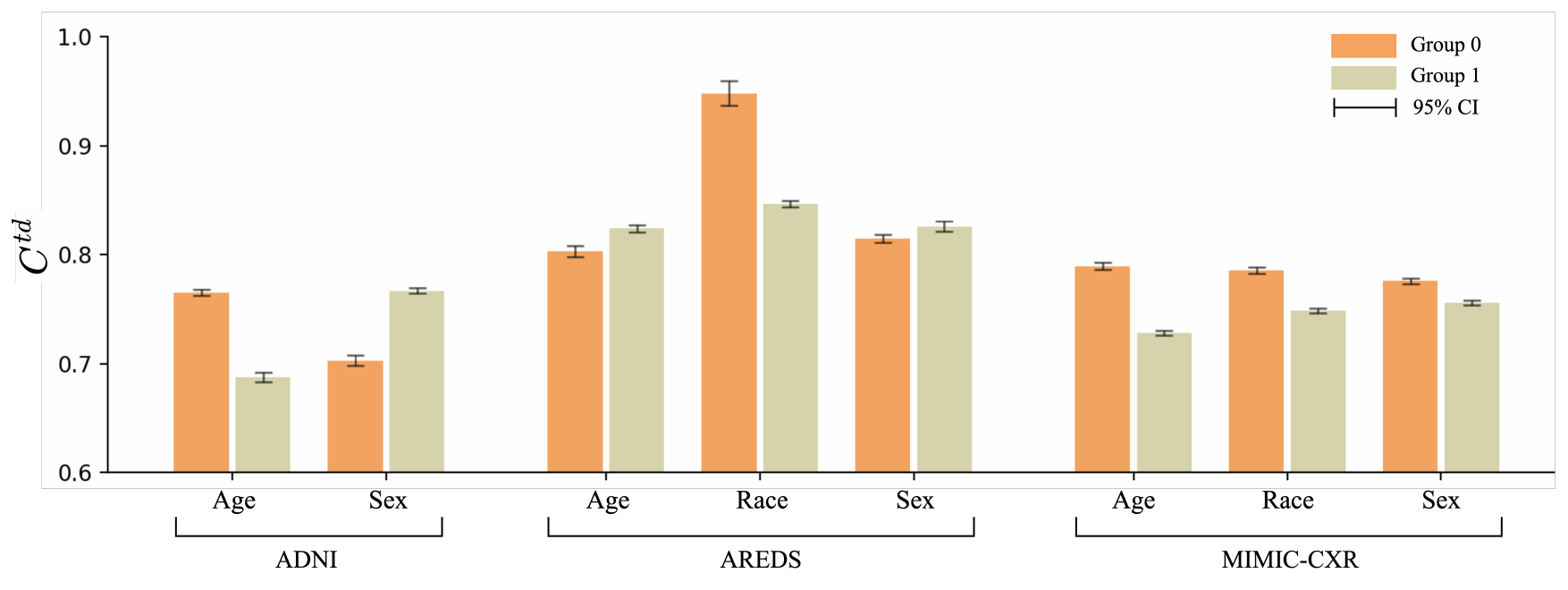}
        \caption{$C^{td}$}
    \end{subfigure}
    \begin{subfigure}[t]{\linewidth}
        \centering
        \includegraphics[width=0.97\textwidth]{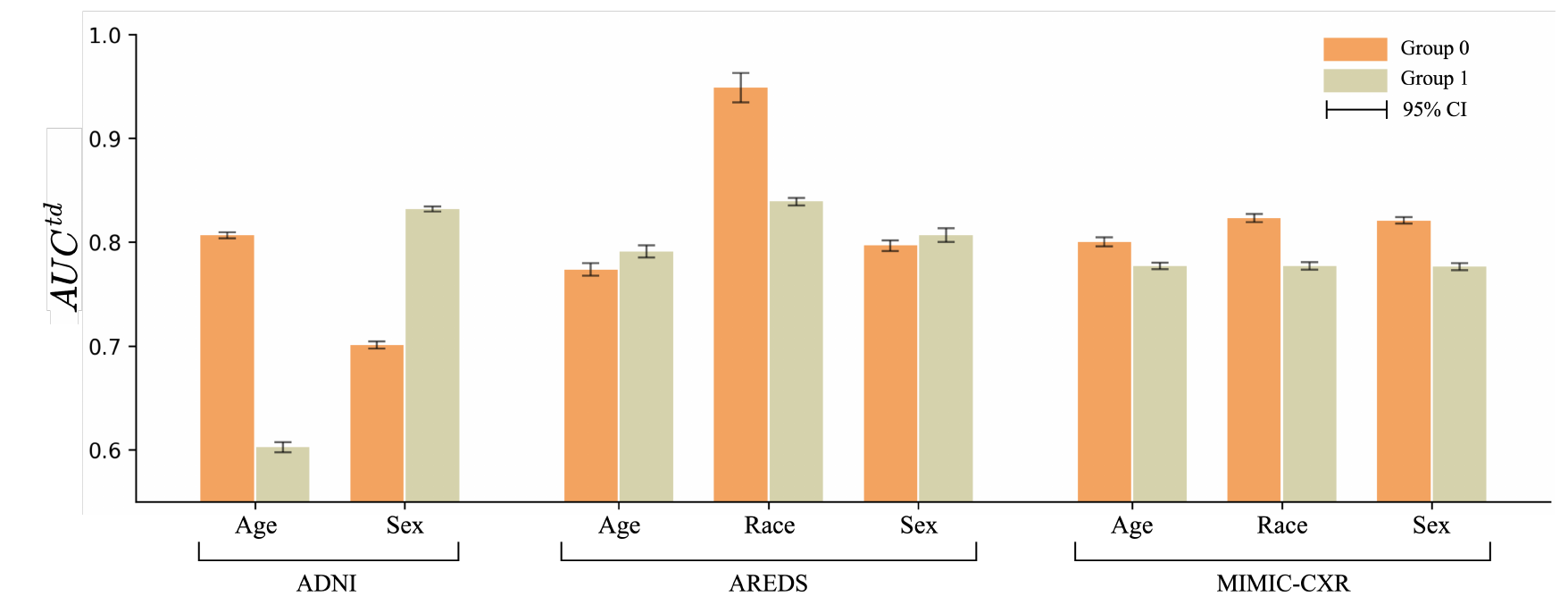}
        \caption{$AUC^{td}$}
    \end{subfigure}
    \begin{subfigure}[t]{\linewidth}
        \centering
        \includegraphics[width=0.97\textwidth]{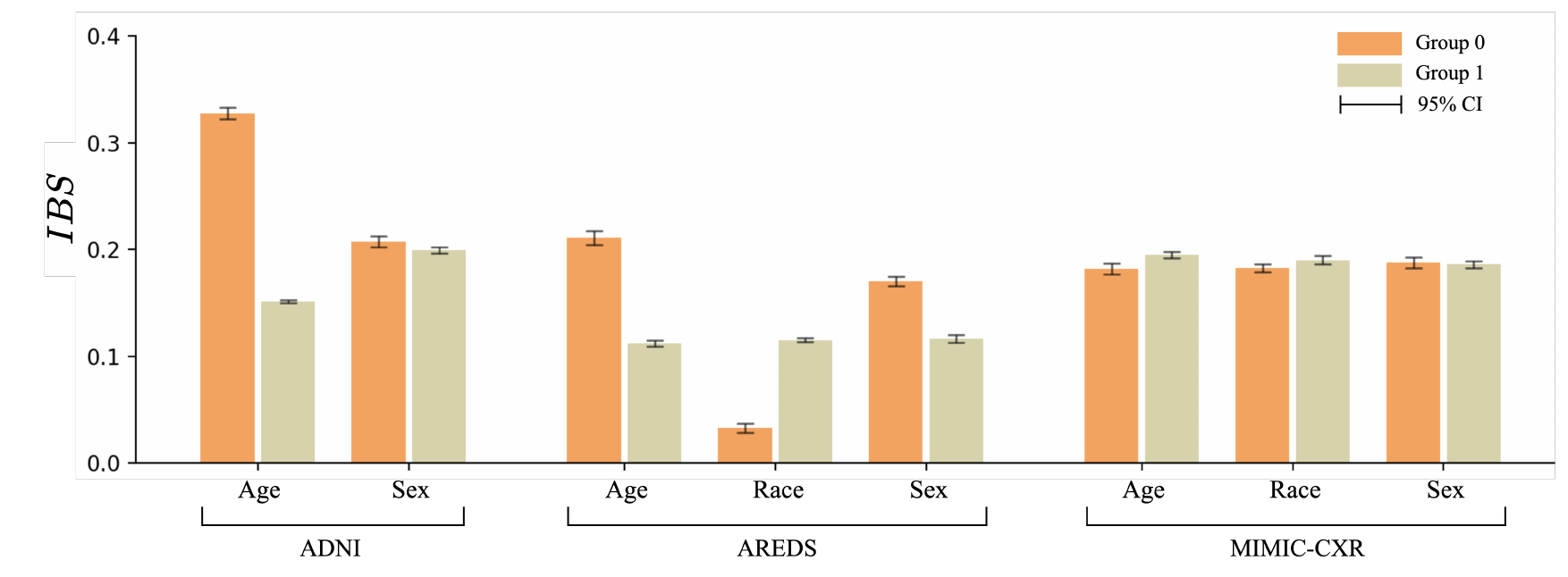}
        \caption{$IBS$}
    \end{subfigure}

    \caption{Per-group predictive performances of TTE prediction models across various datasets and sensitive attribute combinations. The visualized performances correspond to the best models determined by model selection conducted on the validation sets. The 95\% confidence intervals (CIs) are calculated using bootstrapping over the test sets. a) Results measured by $C^{td}$; b) Results measured by $C^{td}$; c) Results measured by $IBS$.\label{fig:a1}}
\end{figure}

\begin{figure}[H]
\centering
\includegraphics[width=\textwidth]{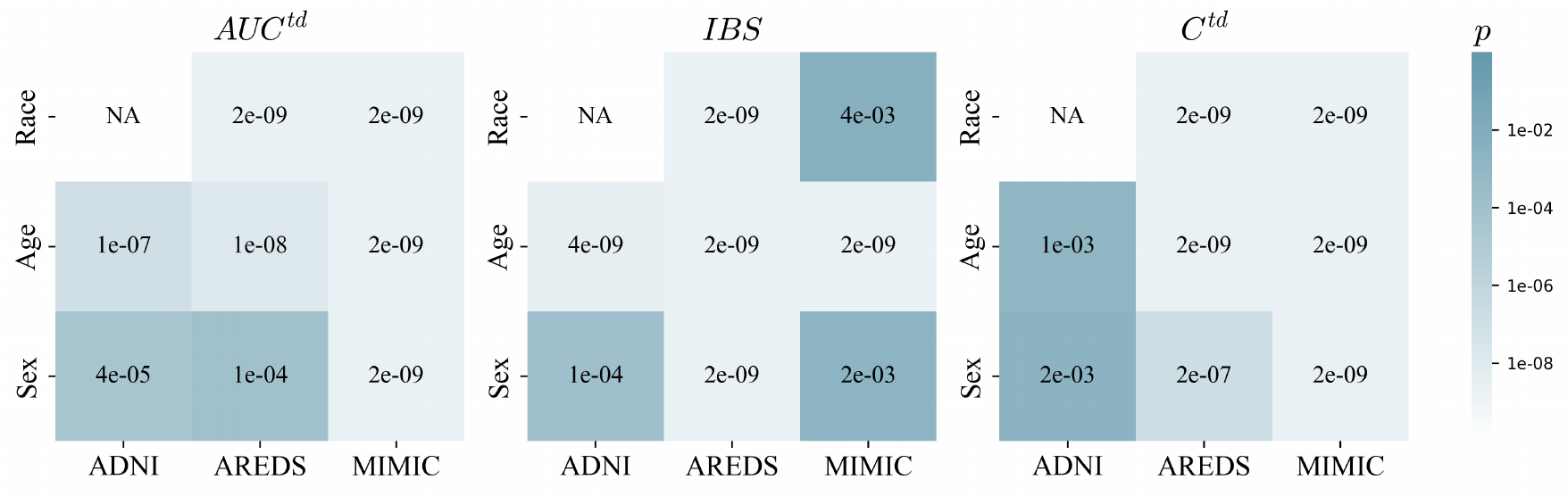}
    \caption{P-values from the two-sided Wilcoxon signed-rank test computed across all TTE prediction models and random seeds. A p-value $< 0.05$ indicates that there is significant differences in predictive performance between groups.}
    \label{fig:a2}
\end{figure}

Figure~\ref{fig:a1} presents the complete per-group performance results of TTE prediction models—DeepHit, Nnet-Survival, and PMF—across all dataset, sensitive attribute, and metric combinations, while Figure~\ref{fig:a2} reports the corresponding significance tests using the two-sided Wilcoxon signed-rank test. As shown, performance gaps between groups are observed across all settings.

\newpage
\subsection{Comparison between Pre-Training and Training from Scratch Strategies for TTE Prediction Models}

\subsubsection{Comparison in Predictive Performance}
\begin{figure}[H]
    \centering
    \begin{subfigure}[t]{\linewidth}
        \centering
        \includegraphics[width=\textwidth]{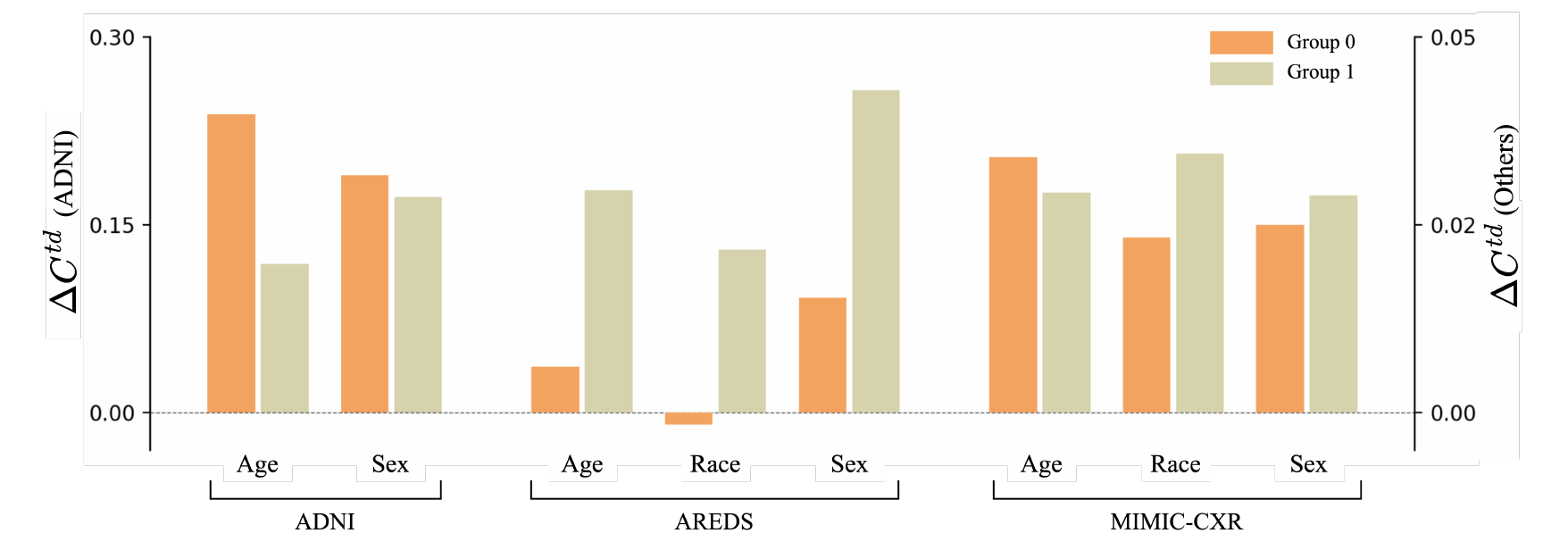}
        \caption{$C^{td}$}
    \end{subfigure}

    \begin{subfigure}[t]{\linewidth}
        \centering
        \includegraphics[width=\textwidth]{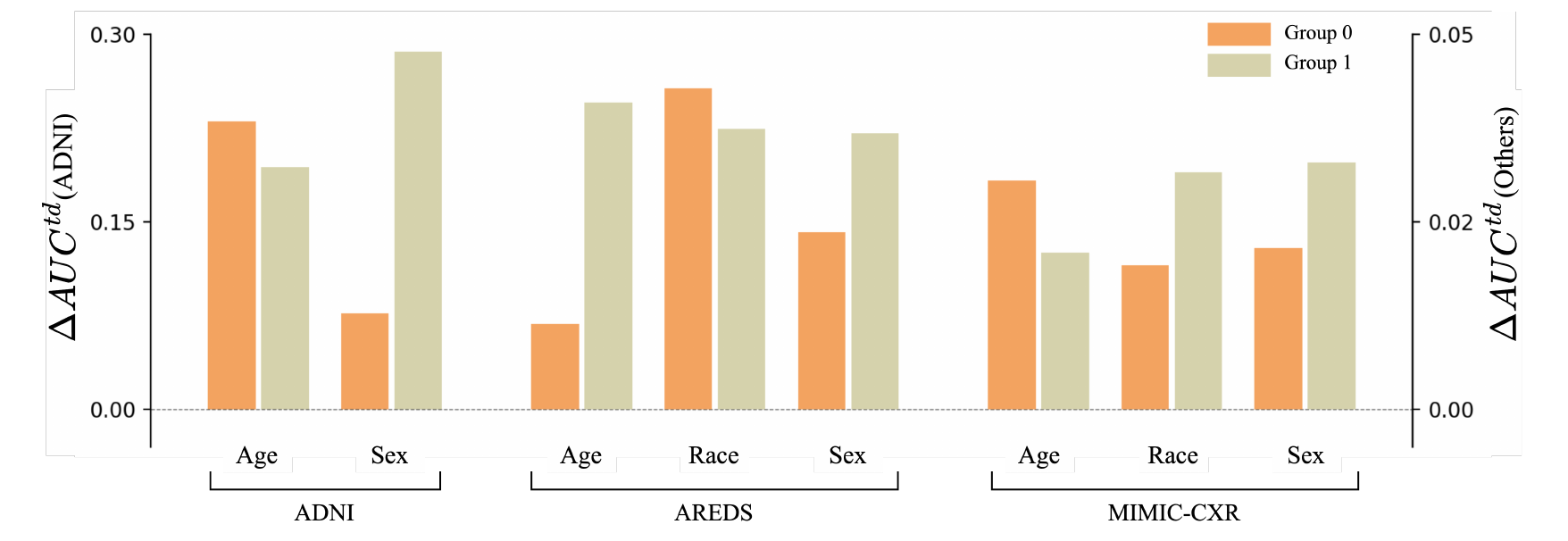}
        \caption{$AUC^{td}$}
    \end{subfigure}

    \begin{subfigure}[t]{\linewidth}
        \centering
        \includegraphics[width=\textwidth]{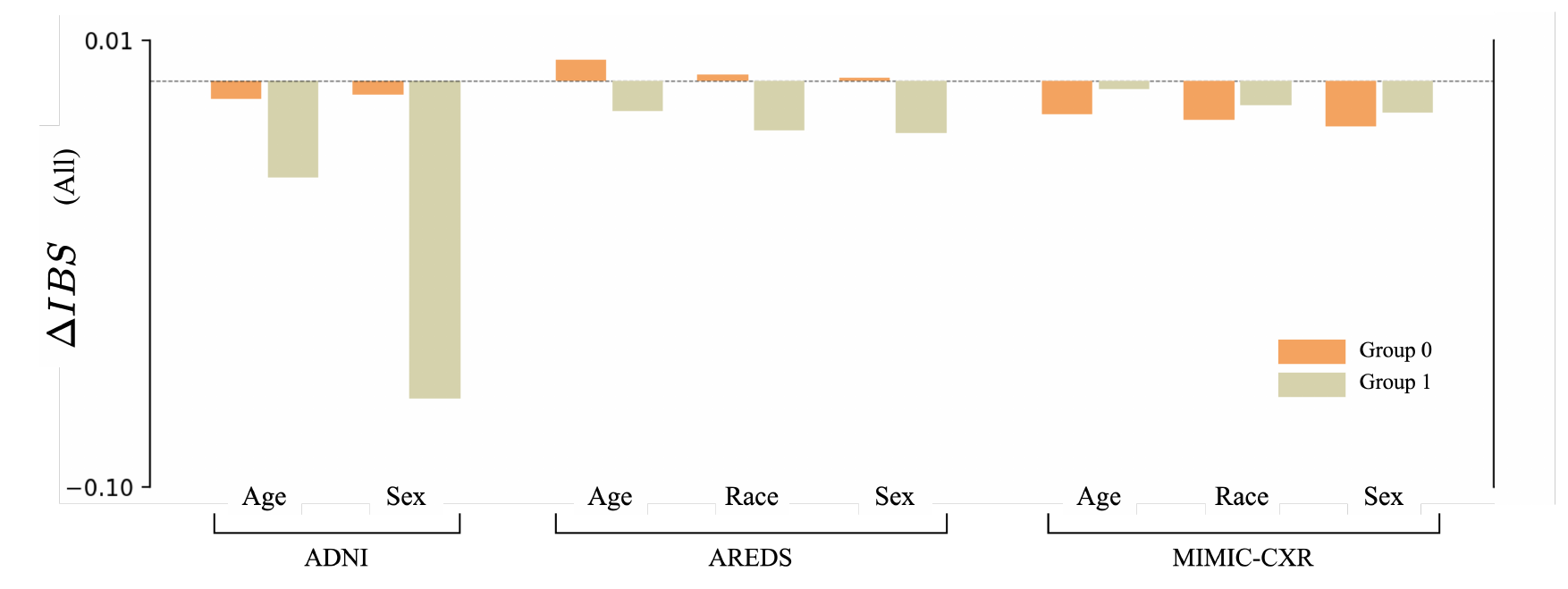}
        \caption{$IBS$}
    \end{subfigure}
    \caption{Per-group average performance gap for TTE prediction models using a pre-training strategy compared to training from scratch across various datasets and sensitive attribute combinations. Positive $\Delta C^{td}$ and $\Delta AUC^{td}$ and negative $\Delta IBS$ values indicate that the pre-training strategy enhances predictive performance relative to training from scratch. a) Results measured by $C^{td}$; b) Results measured by $C^{td}$; c) Results measured by $IBS$.\label{fig:a3}}
\end{figure}

Figure~\ref{fig:a3} presents the complete per-group predictive performance gap between pre-training and training from scratch approaches across all dataset, sensitive attribute, and metric combinations.  As shown, pre-training consistently improves the predictive performance of TTE models across most settings.

\subsubsection{Comparison in Fairness}
\begin{figure}[H]
\centering
\includegraphics[width=\textwidth]{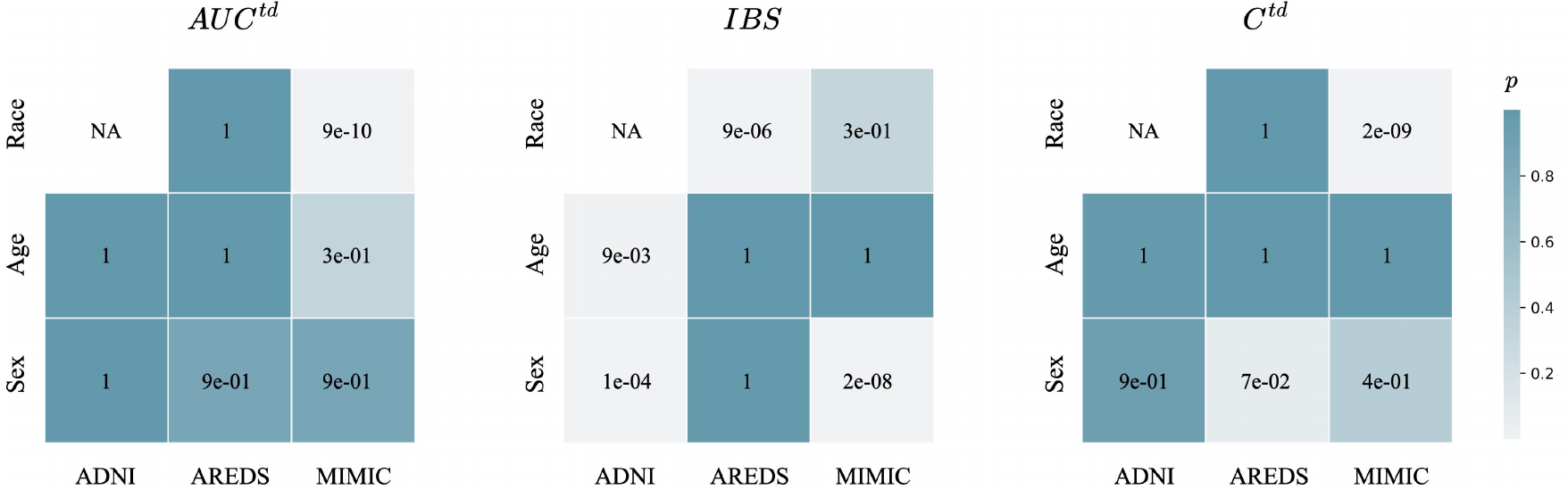}
    \caption{P-values from the one-sided Wilcoxon signed-rank test computed across all TTE prediction models and random seeds. A p-value $> 0.05$ suggests that pre-training does not result in significantly more equitable predictions compared to training from scratch.}
    \label{fig:a4}
\end{figure}

Figure~\ref{fig:a4} presents the significant differences in terms of fairness between pre-training and training from scratch strategies. As shown, we do not observe a significant improvement with pre-training compared to training from scratch. Specifically, the p-values from one-sided Wilcoxon signed-rank tests are larger than 0.05 in 18 out of 24 settings, suggesting that pre-training does not lead to more equitable predictions in most cases.

\subsection{Comparison with Advanced Image Backbones and Medical Pre-training for TTE Prediction Models}
To investigate whether more advanced vision backbones pretrained on medical imaging data can enhance the predictive performance and fairness of TTE prediction models, we conduct an additional experiment using the late AMD progression prediction task on the AREDS dataset. Specifically, we adopt RETFound—a widely recognized eye-specific foundation model based on the Vision Transformer (ViT) architecture~\cite{dosovitskiy2021image} and pretrained on millions of retinal images—as the image backbone for our TTE models. As shown in Table~\ref{tab:a3}, RETFound with ViT does not demonstrate any improvement over EfficientNet pretrained on ImageNet for this task, suggesting that general-purpose backbones may remain competitive despite the availability of domain-specific pretraining.

\begin{table}[H]
\centering
\caption{Predictive and fairness performances of DeepHit using EfficientNet and Vision Transformer as vision backbones on AREDS dataset. EfficientNet is pretrained on ImageNet dataset while Vision Transformer is initialized with pretrained weights provided by RETFound.}
\vskip 0.1in
\label{tab:a3}
\begin{tabular}{ll|ccc|ccc}
\hline
\multicolumn{2}{c|}{} & \multicolumn{3}{c|}{\textbf{EfficientNet}} & \multicolumn{3}{c}{\textbf{Vision Transformer}} \\
\multicolumn{2}{c|}{\textbf{Sensitive Attribute}} & \( AUC^{td} \uparrow \) & \( IBS \downarrow \) & \( C^{td} \uparrow \) & \( AUC^{td} \uparrow \) & \( IBS \downarrow \) & \( C^{td} \uparrow \) \\
\hline
\multirow{3}{*}{\textbf{Accuracy}} 
& Age  & 78.41 & 15.37 & 81.30 & 78.04 & 15.22 & 80.65 \\
& Sex  & 79.08 & 15.36 & 81.77 & 78.01 & 14.51 & 80.72 \\
& Race & 81.78 & 11.74 & 84.53 & 80.99 & 11.99 & 83.91 \\
\hline
\multicolumn{2}{c|}{\textbf{Sensitive Attribute}} & \( \mathcal{F}_{AUC^{td}} \downarrow \) & \( \mathcal{F}_{IBS} \downarrow \) & \( \mathcal{F}_{C^{td}} \downarrow \) & \( \mathcal{F}_{AUC^{td}} \downarrow \) & \( \mathcal{F}_{IBS} \downarrow \) & \( \mathcal{F}_{C^{td}} \downarrow \) \\
\hline
\multirow{3}{*}{\textbf{Fairness}} 
& Age  & 1.58  & 12.56 & 2.20  & 0.35  & 9.50  & 1.44 \\
& Sex  & 0.76  & 3.84  & 1.32  & 0.85  & 4.26  & 0.44 \\
& Race & 14.00 & 10.14 & 11.09 & 9.32  & 10.47 & 10.27 \\
\hline
\end{tabular}
\end{table}

\vfill

\subsection{Fairness in Fair TTE Prediction Models}

\begin{figure}[H]
    \begin{center}
    \includegraphics[width=\textwidth]{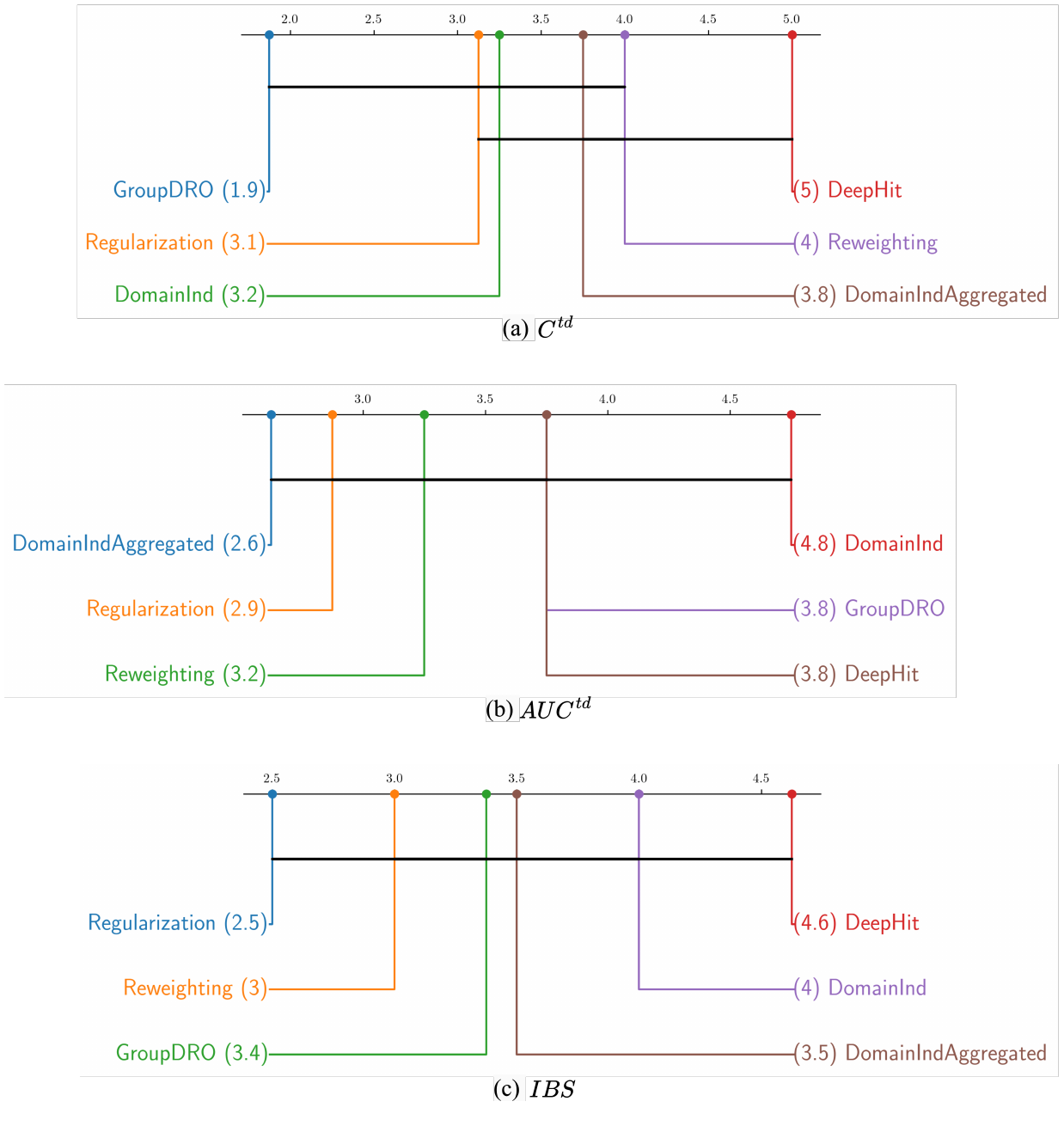}
    \end{center}
    \caption{Critical Difference diagrams for all methods calculated from all dataset and sensitive attribute combinations. Although fairness algorithms are generally ranked higher than DeepHit in all settings, there is no significant difference in terms of fairness as indicated by the connections between fairness algorithms and DeepHit in the diagrams. a) Diagram for $C^{td}$; b) Diagram for $AUC^{td}$; c) Diagram for $IBS$.}
    \label{fig:a5}
\end{figure}

Table~\ref{tab:a4} and Figure~\ref{fig:a5} present the results of statistical significance testing for fair TTE prediction models, conducted using the Friedman test followed by the Nemenyi post-hoc test.

\begin{table}[H]
\caption{P-values from the Friedman test followed by a Nemenyi post-hoc test computed across all dataset and sensitive attribute combinations. A p-value $< 0.05$ indicates that the significant difference in terms of fairness between the two corresponding methods.\label{tab:a4}}
\vskip 0.1in
\centering
\begin{tabular}{llllllll} \toprule
 Metrics& Models & \multicolumn{1}{c}{DI} & \multicolumn{1}{c}{CSA} & \multicolumn{1}{c}{DRO} & \multicolumn{1}{c}{DeepHit} & \multicolumn{1}{c}{FRL} & \multicolumn{1}{c}{SR} \\ \hline
\multirow{6}{*}{$C^{td}$} 
 & DI & 1.000 & 0.995 & 0.684 & 0.420 & 1.000 & 0.967 \\
 & CSA & 0.995 & 1.000 & 0.340 & 0.765 & 0.985 & 1.000 \\
 & DRO & 0.684 & 0.340 & 1.000 & \textbf{0.011} & 0.765 & 0.206 \\
 & DeepHit & 0.420 & 0.765 & \textbf{0.011} & 1.000 & 0.340 & 0.894 \\
 & FRL & 1.000 & 0.985 & 0.765 & 0.340 & 1.000 & 0.937 \\
 & SR & 0.967 & 1.000 & 0.206 & 0.894 & 0.937 & 1.000 \\ \hline
\multirow{6}{*}{$AUC^{td}$} & DI & 1.000 & 0.206 & 0.894 & 0.894 & 0.340 & 0.596 \\
 & CSA & 0.206 & 1.000 & 0.836 & 0.836 & 1.000 & 0.985 \\
 & DRO & 0.894 & 0.836 & 1.000 & 1.000 & 0.937 & 0.995 \\
 & DeepHit & 0.894 & 0.836 & 1.000 & 1.000 & 0.937 & 0.995 \\
 & FRL & 0.340 & 1.000 & 0.937 & 0.937 & 1.000 & 0.999 \\
 & SR & 0.596 & 0.985 & 0.995 & 0.995 & 0.999 & 1.000 \\ \hline
\multirow{6}{*}{$IBS$} & DI & 1.000 & 0.995 & 0.985 & 0.985 & 0.596 & 0.894 \\
 & CSA & 0.995 & 1.000 & 1.000 & 0.836 & 0.894 & 0.995 \\
 & DRO & 0.985 & 1.000 & 1.000 & 0.765 & 0.937 & 0.999 \\
 & DeepHit & 0.985 & 0.836 & 0.765 & 1.000 & 0.206 & 0.507 \\
 & FRL & 0.596 & 0.894 & 0.937 & 0.206 & 1.000 & 0.995 \\
 & SR & 0.894 & 0.995 & 0.999 & 0.507 & 0.995 & 1.000 \\ \bottomrule
\end{tabular}
\end{table}

\vfill
\newpage

\vfill
\subsection{Fairness-Utility Trade-Off Results}
\begin{figure}[H]
  \begin{center}
    \includegraphics[width=0.65\textwidth]{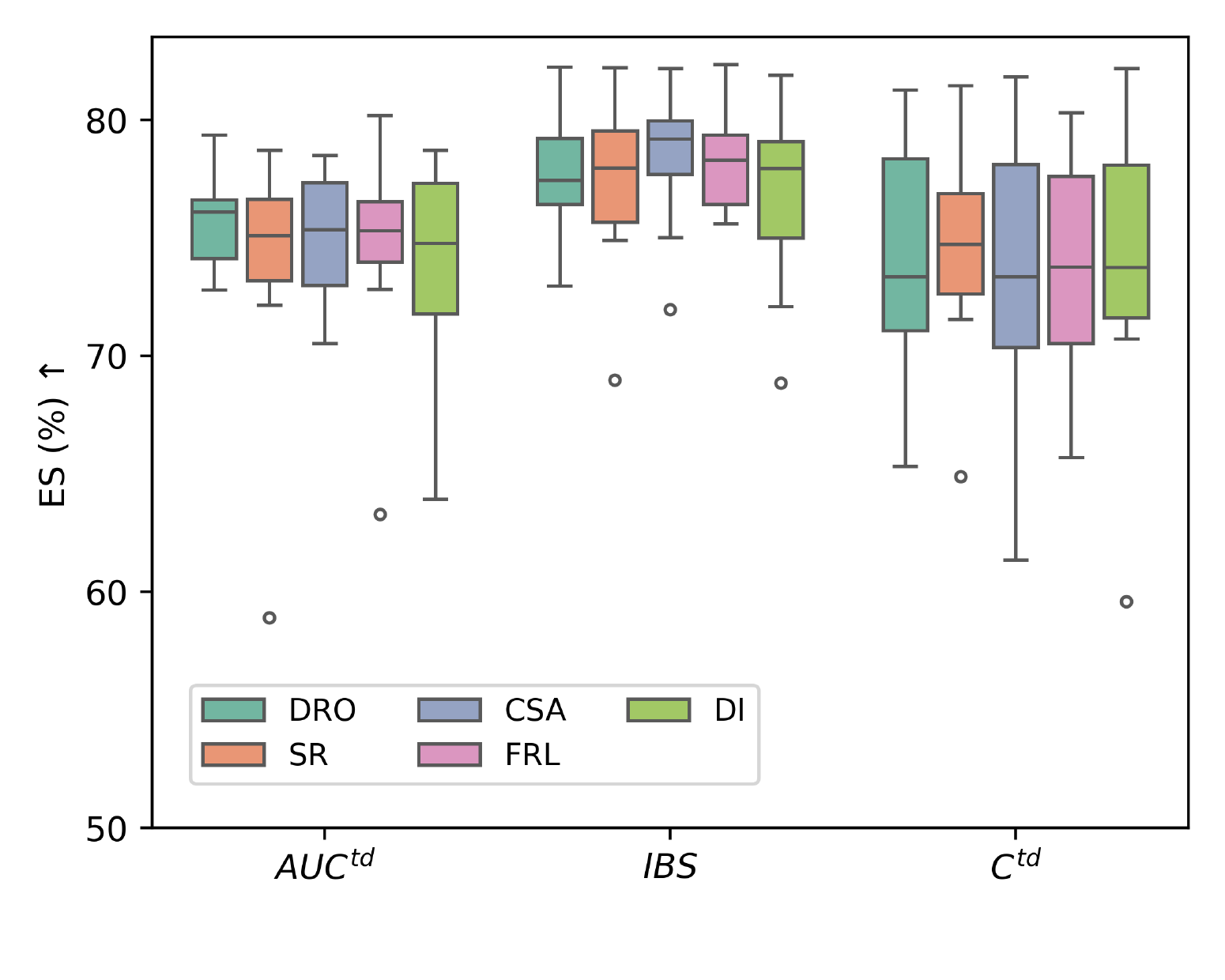}
  \end{center}
  \caption{Fairness-utility trade-offs of fairness algorithms for TTE prediction across various utility metrics. For each metric, we compute the corresponding equity scaling score as a measure of the trade-off. The results for each fairness algorithm are aggregated across all dataset and sensitive attribute combinations.\label{fig:a6}}
\end{figure}

\begin{figure}[H]
  \begin{center}
    \includegraphics[width=0.65\textwidth]{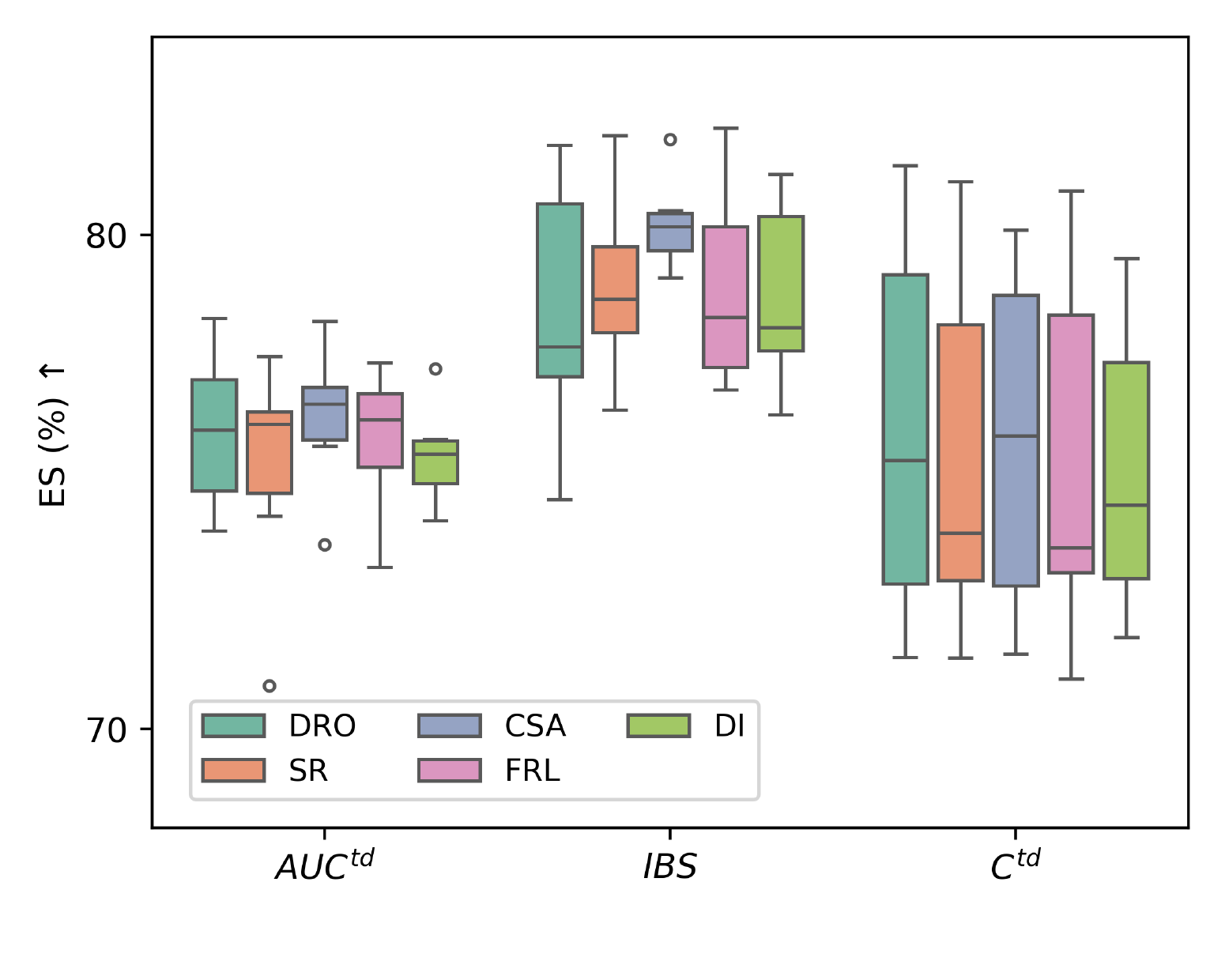}
  \end{center}
  \caption{Fairness-utility trade-offs of fairness algorithms for TTE prediction across various utility metrics under distribution shift created by flipping censoring indices (shift on $X$). For each metric, we compute the corresponding equity scaling score as a measure of the trade-off. The results for each fairness algorithm are aggregated across all dataset and sensitive attribute combinations.\label{fig:a7}}
\end{figure}

\begin{figure}[H]
  \begin{center}
    \includegraphics[width=0.65\textwidth]{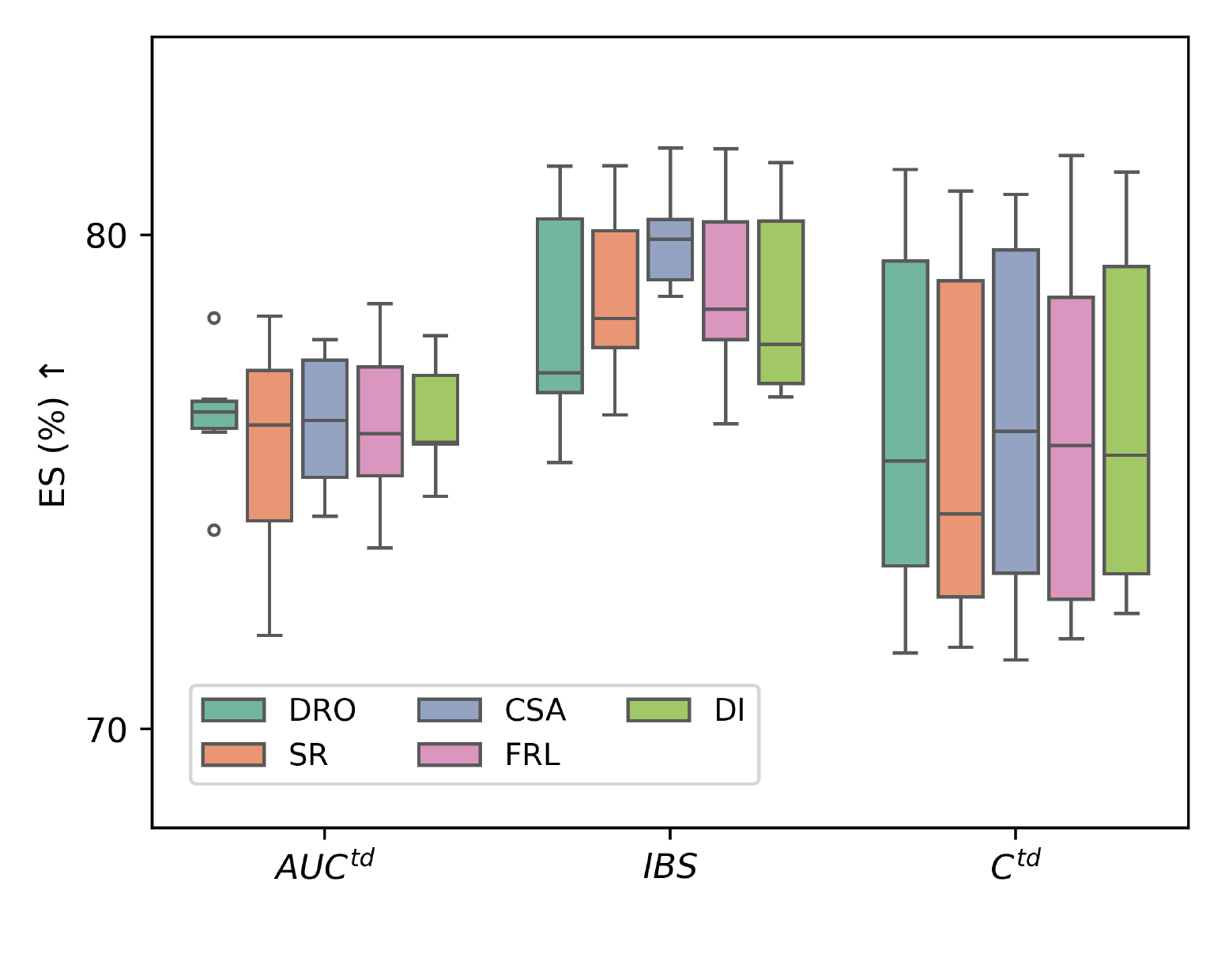}
  \end{center}
  \caption{Fairness-utility trade-offs of fairness algorithms for TTE prediction across various utility metrics under distribution shift created by flipping censoring indices (shift on $Y$). For each metric, we compute the corresponding equity scaling score as a measure of the trade-off. The results for each fairness algorithm are aggregated across all dataset and sensitive attribute combinations.\label{fig:a8}}
\end{figure}

\begin{figure}[H]
  \begin{center}
    \includegraphics[width=0.65\textwidth]{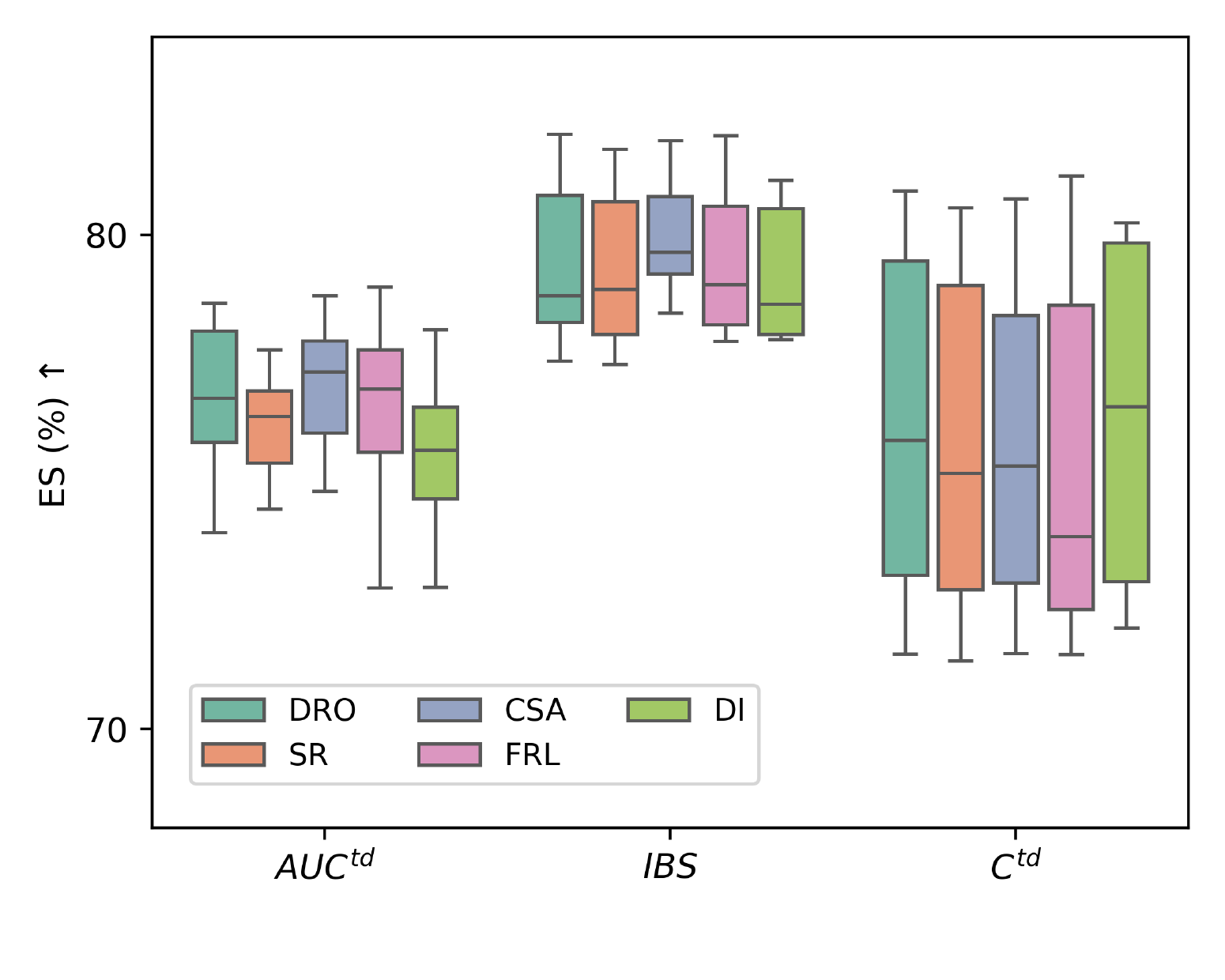}
  \end{center}
  \caption{Fairness-utility trade-offs of fairness algorithms for TTE prediction across various utility metrics under distribution shift created by flipping censoring indices (shift on $\Delta$). For each metric, we compute the corresponding equity scaling score as a measure of the trade-off. The results for each fairness algorithm are aggregated across all dataset and sensitive attribute combinations.\label{fig:a9}}
\end{figure}

Incorporating fairness shifts the objective from pure utility optimization to balancing utility and fairness. To assess this trade-off in fair TTE prediction methods, we compute equity scaling scores~\cite{luo2024harvard} across datasets and sensitive attributes under both in-distribution and distribution shift scenarios. As shown in Figures~\ref{fig:a6}–\ref{fig:a9}, different methods exhibit varying fairness-utility trade-offs, with CSA achieving the most favorable balance in most settings.
% However, these patterns are inconsistent across metrics. Notably, DRO performs best for $AUC^{td}$, while CSA and SR excel for $IBS$ and $C^{td}$, respectively. This highlights the need for careful selection of fairness algorithms to achieve an optimal fairness-utility trade-off.

\vfill
\newpage

\subsection{Additional Results for Predictive Performance and Fairness in Fair TTE Prediction Models under Distribution Shift}\label{sec:g5}

This section presents the complete results for fair TTE prediction under distribution shift scenarios. As illustrated in Figure~\ref{fig:3}, we define distribution shift as a setting where correlations between the sensitive attribute and other variables in the causal graph are present in the training data but absent in the testing data. To simulate such shifts, we intervene on one group (the intervened group) by corrupting specific aspects of the data—namely, the images ($X$), TTE labels ($Y$), or censoring indicators ($\Delta$)—while leaving the other group unchanged. Detailed procedures for generating these shifts are provided in Appendix~\ref{sec:f5}.

\subsubsection{Results for Distribution Shift in \texorpdfstring{$X$}{}}

\begin{figure}[H]
    \begin{center}
    \includegraphics[width=0.97\textwidth]{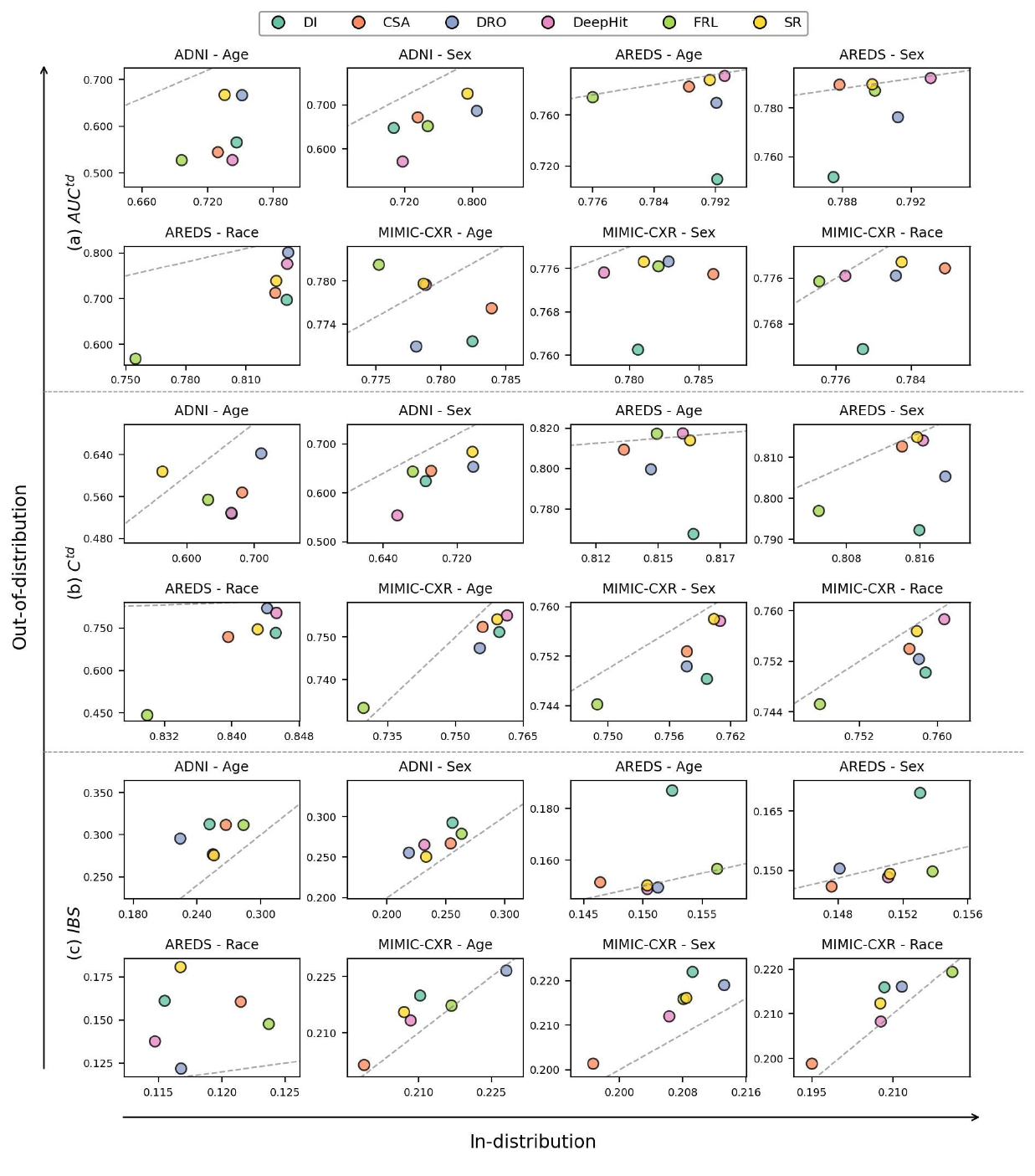}
    \end{center}
    \caption{Comparison of predictive performance for (fair) TTE prediction models in in-distribution vs. out-of-distribution (i.e., shift in $X$) learning scenarios, evaluated across all dataset and sensitive attribute combinations. The displayed results represent the average performance across all random seeds. Points on the dashed line indicate equal performance in both scenarios. a) Results for $AUC^{td}$; b) Results for $C^{td}$; c) Results for $IBS$.}
\end{figure}

\vfill
\newpage

\begin{figure}[H]
    \begin{center}
    \includegraphics[width=\textwidth]{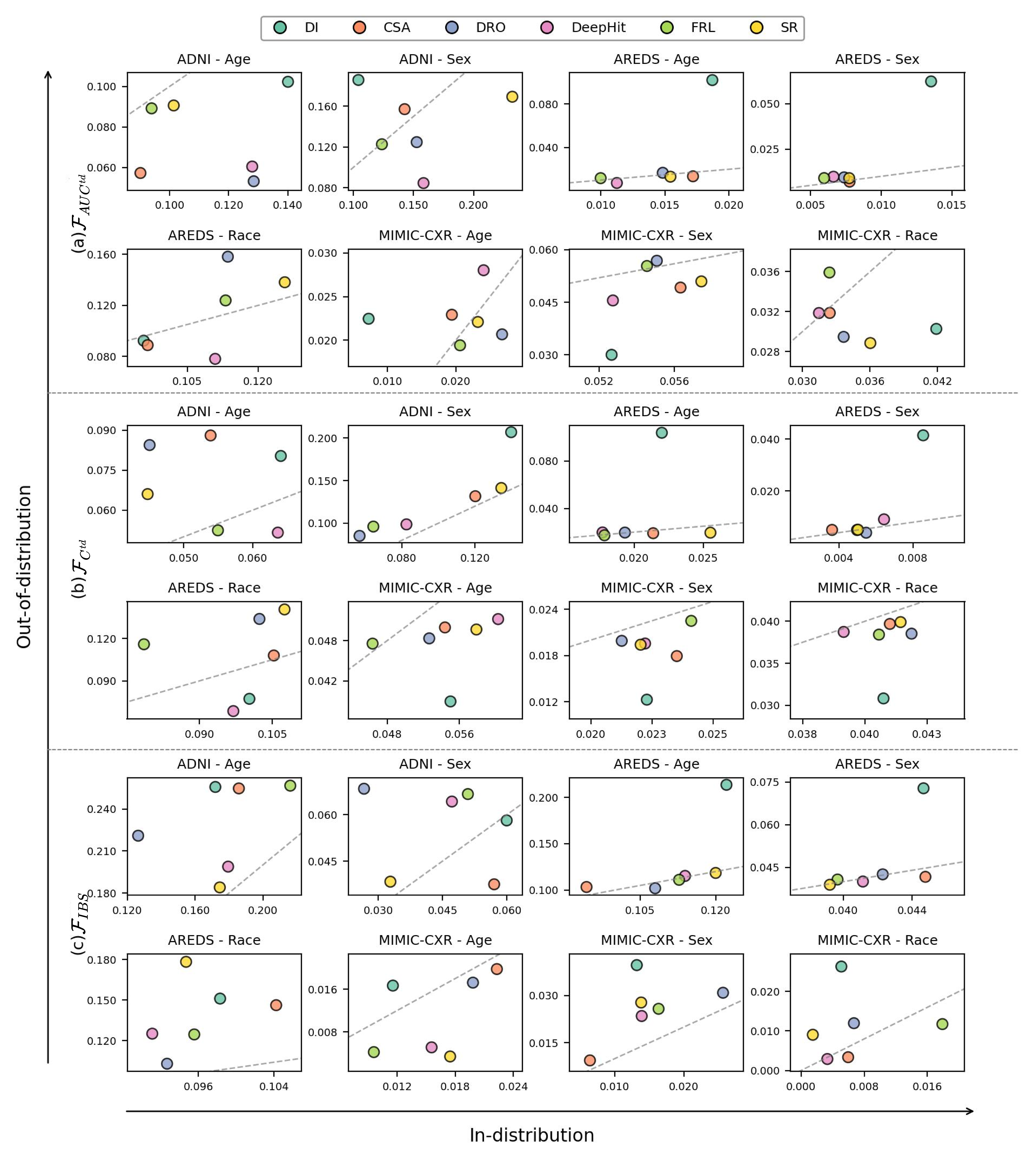}
    \end{center}
    \caption{Comparison of fairness for (fair) TTE prediction models in in-distribution vs. out-of-distribution (i.e., shift in $X$) learning scenarios, evaluated across all dataset and sensitive attribute combinations. The displayed results represent the average performance across all random seeds. Points on the dashed line indicate equal performance in both scenarios. a) Results for $\mathcal{F}_{AUC^{td}}$; b) Results for $\mathcal{F}_{C^{td}}$; c) Results for $\mathcal{F}_{IBS}$.}
\end{figure}

\begin{figure}[H]
    \begin{center}
    \includegraphics[width=\textwidth]{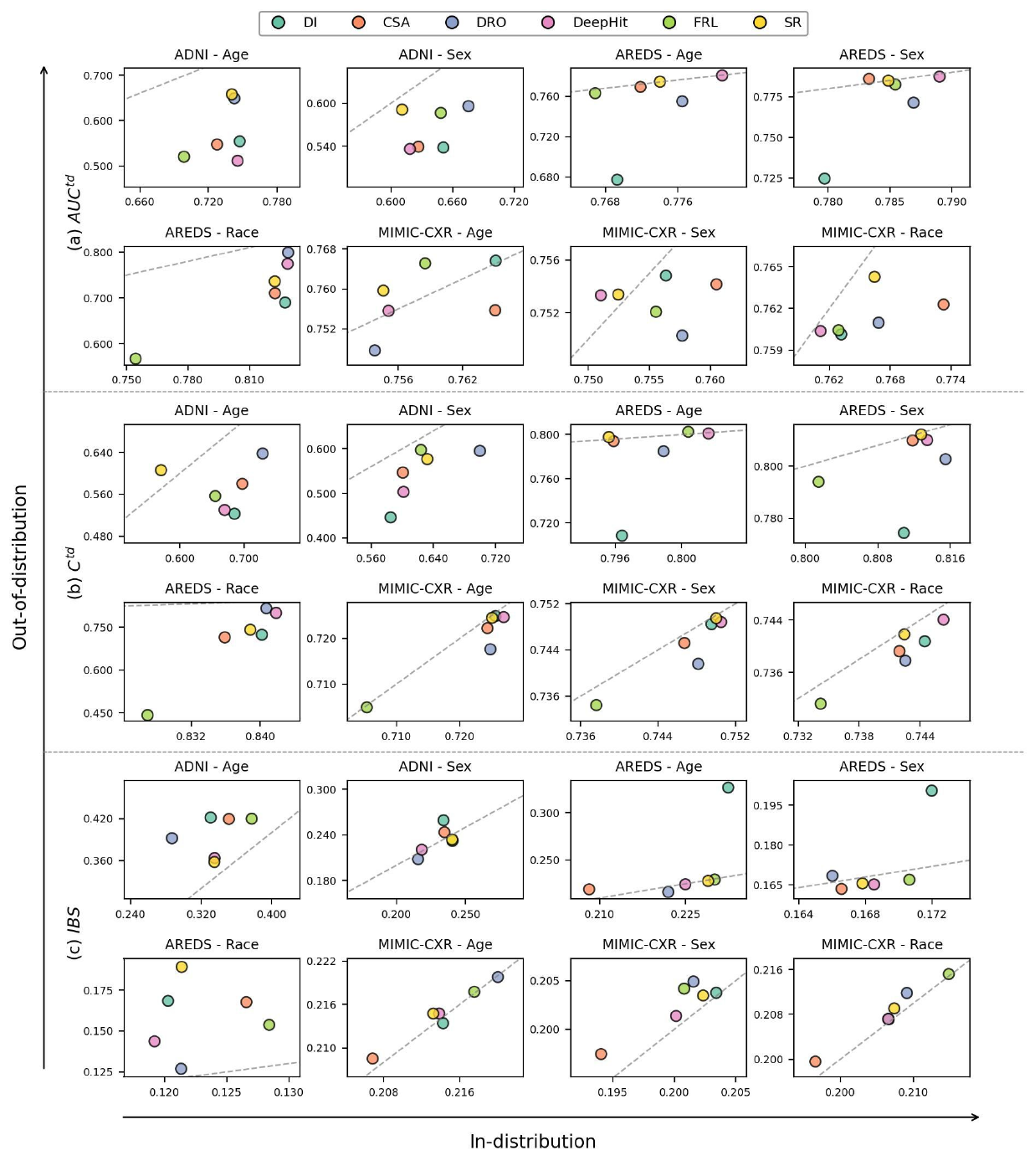}
    \end{center}
    \caption{Comparison of predictive performance on the intervened group for (fair) TTE prediction models in in-distribution vs. out-of-distribution (i.e., shift in $X$) learning scenarios across all dataset and sensitive attribute combinations. The displayed results represent the average performance across all random seeds. Points on the dashed line indicate equal performance in both scenarios. a) Results for $AUC^{td}$; b) Results for $C^{td}$; c) Results for $IBS$.}
\end{figure}

\begin{figure}[H]
    \begin{center}
    \includegraphics[width=\textwidth]{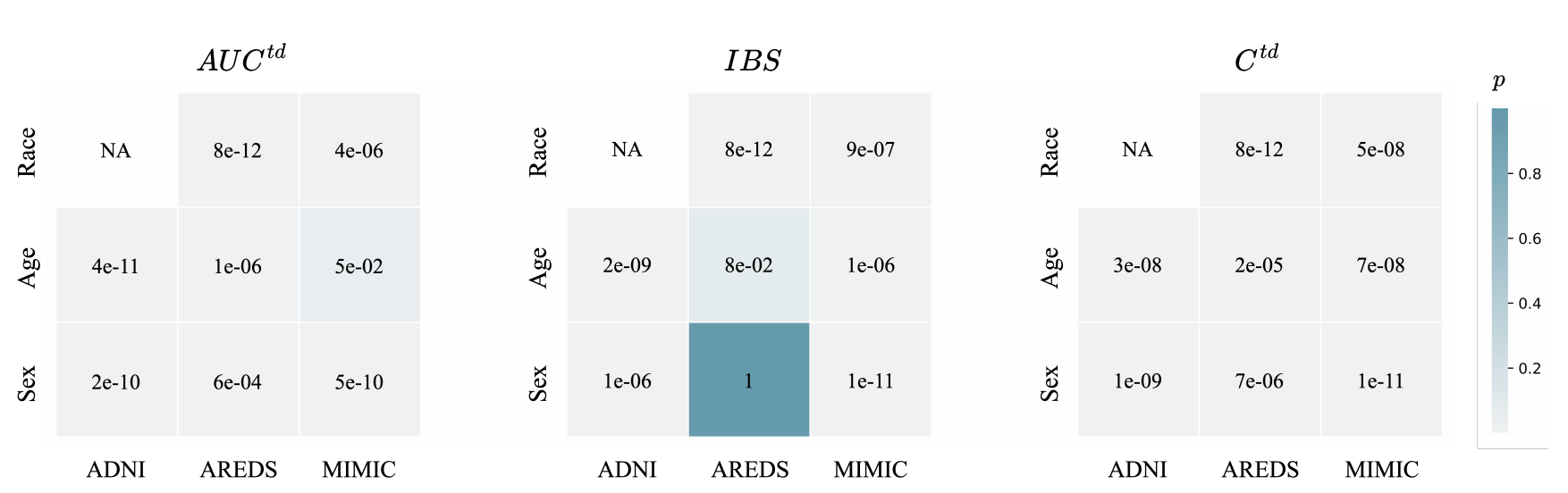}
    \end{center}
    \caption{P-values from the one-sided Wilcoxon signed-rank test computed across all fair TTE prediction models and random seeds. A p-value < 0.05 suggests distribution shift on $X$ significantly degrades TTE predictive performance compared no distribution shift.}
\end{figure}

\subsubsection{Results for Distribution Shift in \texorpdfstring{$Y$}{}}

\begin{figure}[H]
    \begin{center}
    \includegraphics[width=\textwidth]{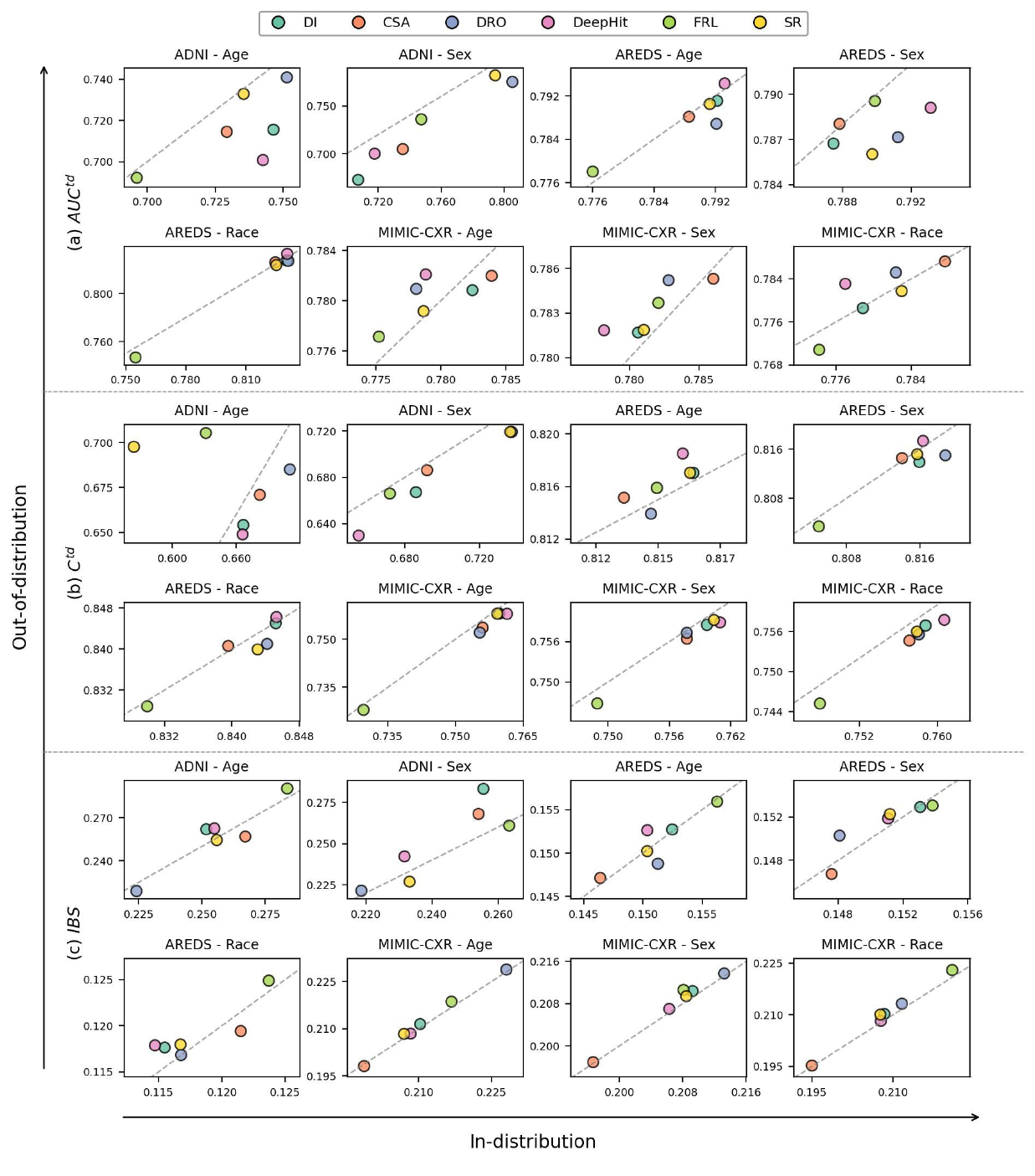}
    \end{center}
    \caption{Comparison of predictive performance for (fair) TTE prediction models in in-distribution vs. out-of-distribution (i.e., shift in $Y$) learning scenarios, evaluated across all dataset and sensitive attribute combinations. The displayed results represent the average performance across all random seeds. Points on the dashed line indicate equal performance in both scenarios. a) Results for $AUC^{td}$; b) Results for $C^{td}$; c) Results for $IBS$.}
\end{figure}

\vfill
\newpage

\begin{figure}[H]
\centering
\includegraphics[width=\textwidth]{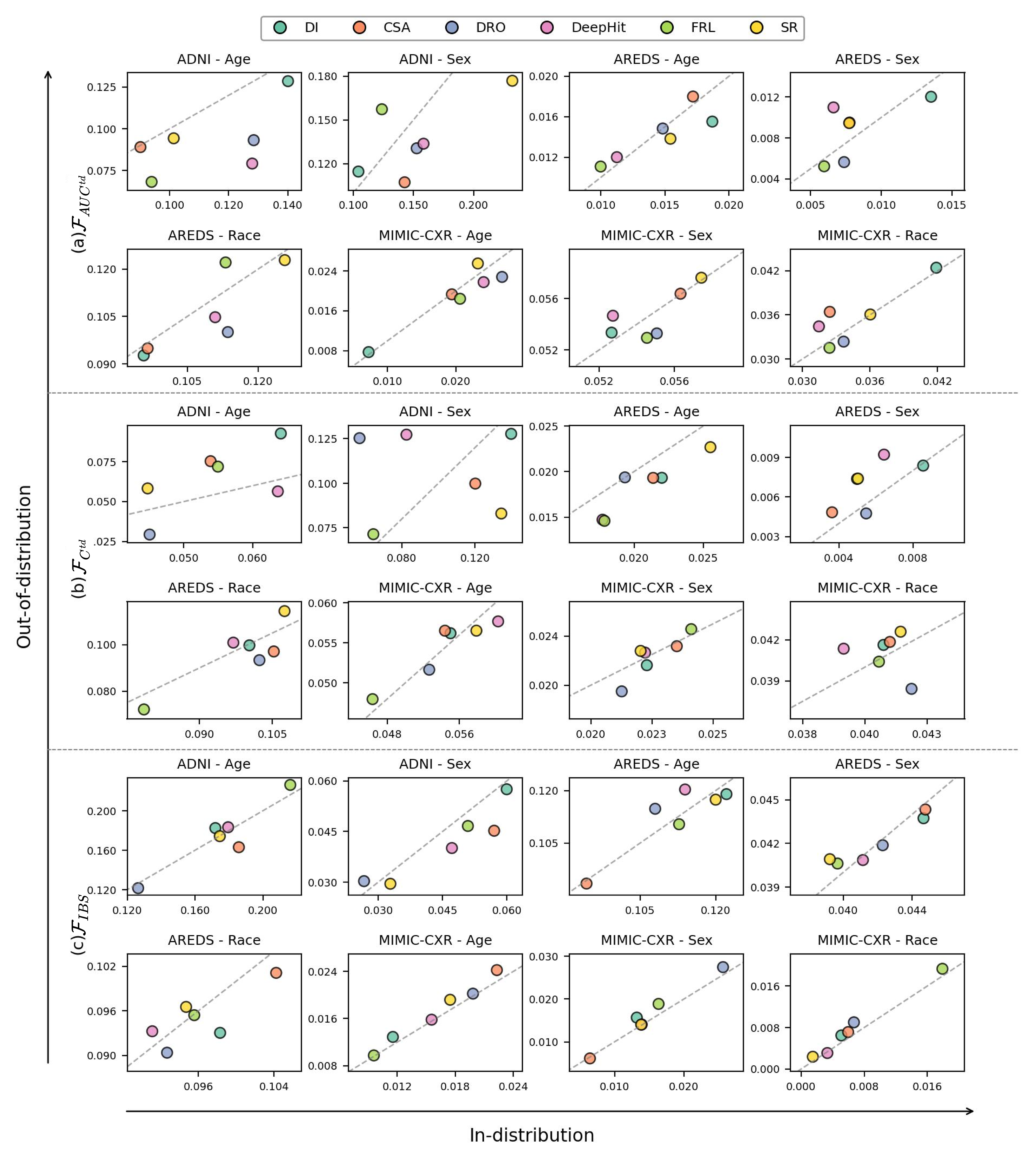}
    \caption{Comparison of fairness for (fair) TTE prediction models in in-distribution vs. out-of-distribution (i.e., shift in $Y$) learning scenarios, evaluated across all dataset and sensitive attribute combinations. The displayed results represent the average performance across all random seeds. Points on the dashed line indicate equal performance in both scenarios. a) Results for $\mathcal{F}_{AUC^{td}}$; b) Results for $\mathcal{F}_{C^{td}}$; c) Results for $\mathcal{F}_{IBS}$.}
\end{figure}

\begin{figure}[H]
    \begin{center}
    \includegraphics[width=\textwidth]{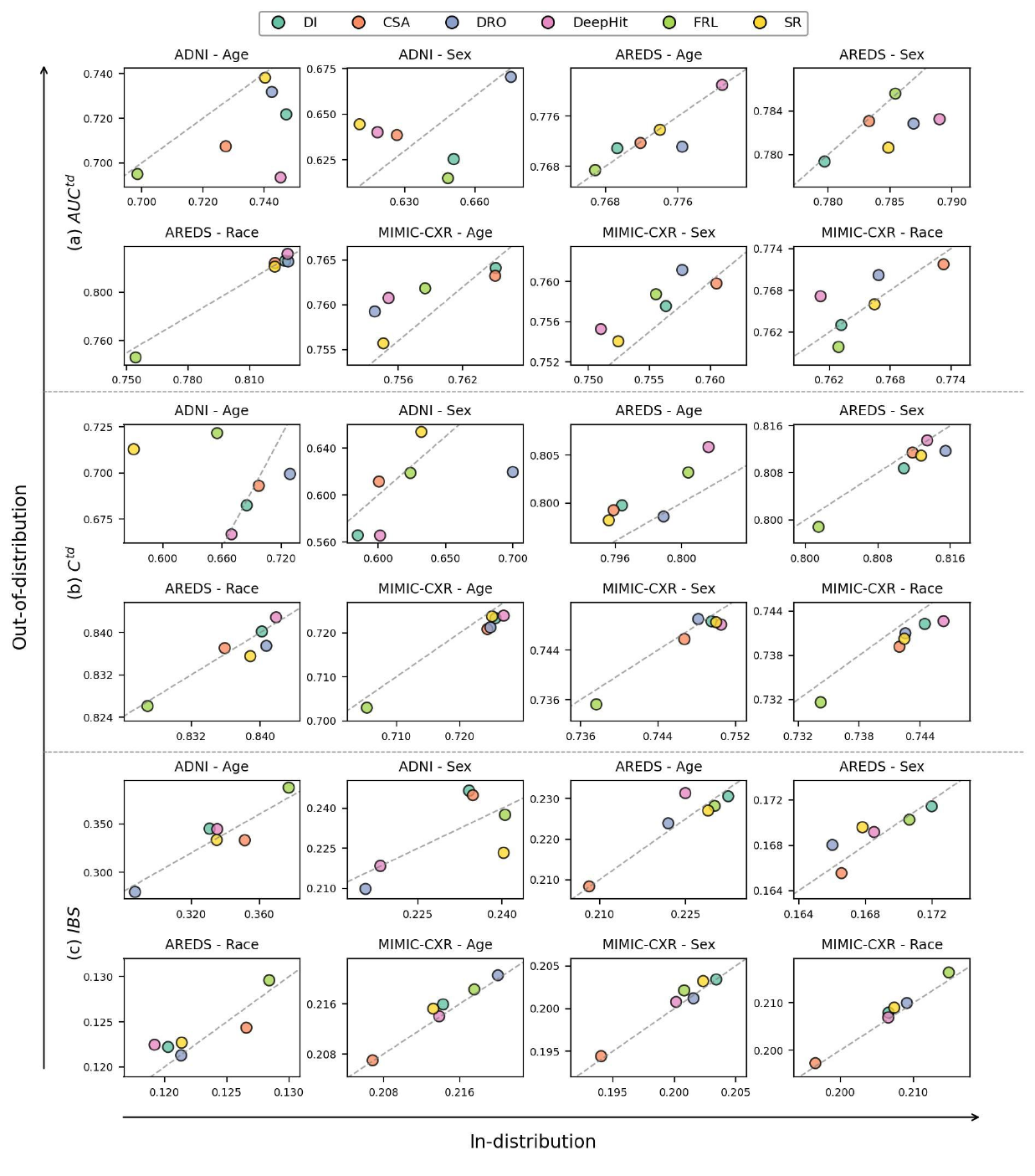}
    \end{center}
    \caption{Comparison of predictive performance on the intervened group for (fair) TTE prediction models in in-distribution vs. out-of-distribution (i.e., shift in $Y$) learning scenarios across all dataset and sensitive attribute combinations. The displayed results represent the average performance across all random seeds. Points on the dashed line indicate equal performance in both scenarios. a) Results for $AUC^{td}$; b) Results for $C^{td}$; c) Results for $IBS$.}
\end{figure}

\begin{figure}[H]
    \begin{center}
    \includegraphics[width=\textwidth]{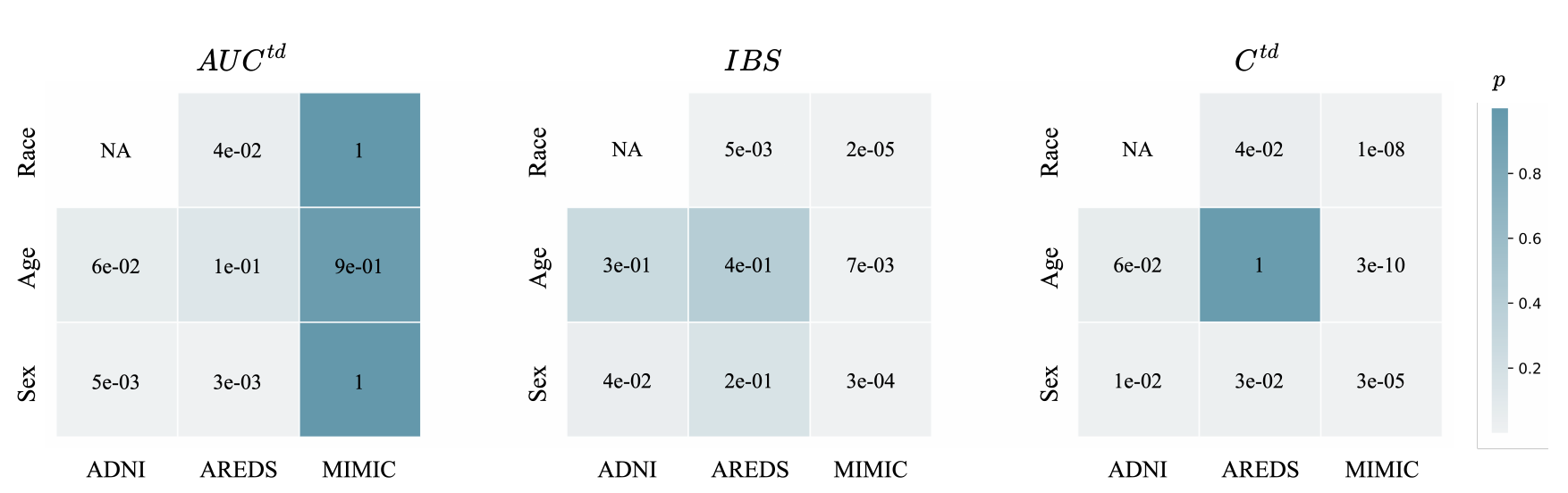}
    \end{center}
    \caption{P-values from the one-sided Wilcoxon signed-rank test computed across all fair TTE prediction models and random seeds. A p-value < 0.05 suggests distribution shift on $Y$ significantly degrades TTE predictive performance compared no distribution shift.}
\end{figure}

\subsubsection{Results for Distribution Shift in \texorpdfstring{$\Delta$}{}}

\begin{figure}[H]
    \begin{center}
    \includegraphics[width=\textwidth]{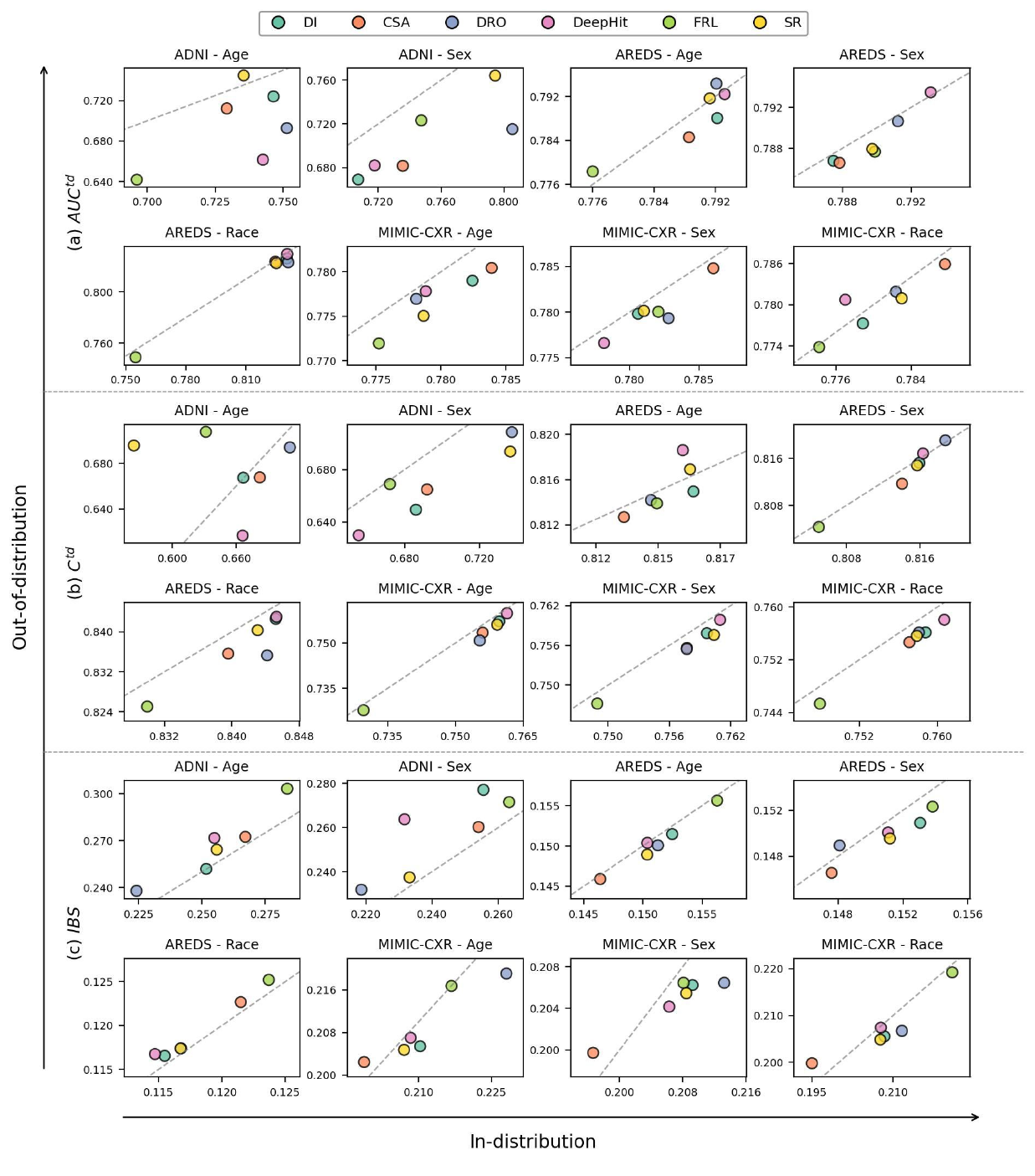}
    \end{center}
    \caption{Comparison of predictive performance for (fair) TTE prediction models in in-distribution vs. out-of-distribution (i.e., shift in $\Delta$) learning scenarios, evaluated across all dataset and sensitive attribute combinations. The displayed results represent the average performance across all random seeds. Points on the dashed line indicate equal performance in both scenarios. a) Results for $AUC^{td}$; b) Results for $C^{td}$; c) Results for $IBS$.}
\end{figure}

\vfill
\newpage

\begin{figure}[H]
    \begin{center}
    \includegraphics[width=\textwidth]{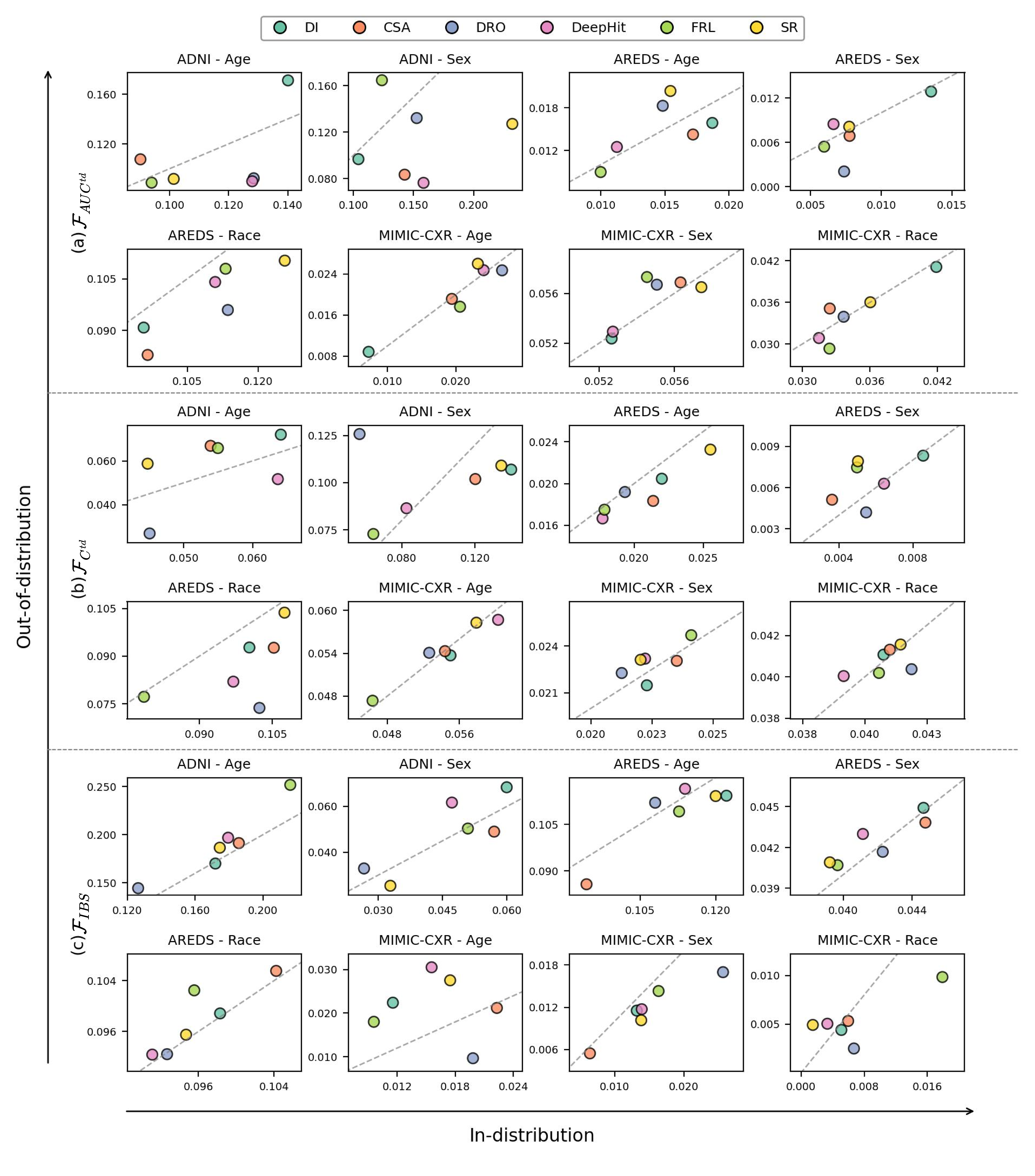}
    \end{center}
    \caption{Comparison of fairness for (fair) TTE prediction models in in-distribution vs. out-of-distribution (i.e., shift in $\Delta$) learning scenarios, evaluated across all dataset and sensitive attribute combinations. The displayed results represent the average performance across all random seeds. Points on the dashed line indicate equal performance in both scenarios. a) Results for $\mathcal{F}_{AUC^{td}}$; b) Results for $\mathcal{F}_{C^{td}}$; c) Results for $\mathcal{F}_{IBS}$.}
\end{figure}

\begin{figure}[H]
    \begin{center}
    \includegraphics[width=\textwidth]{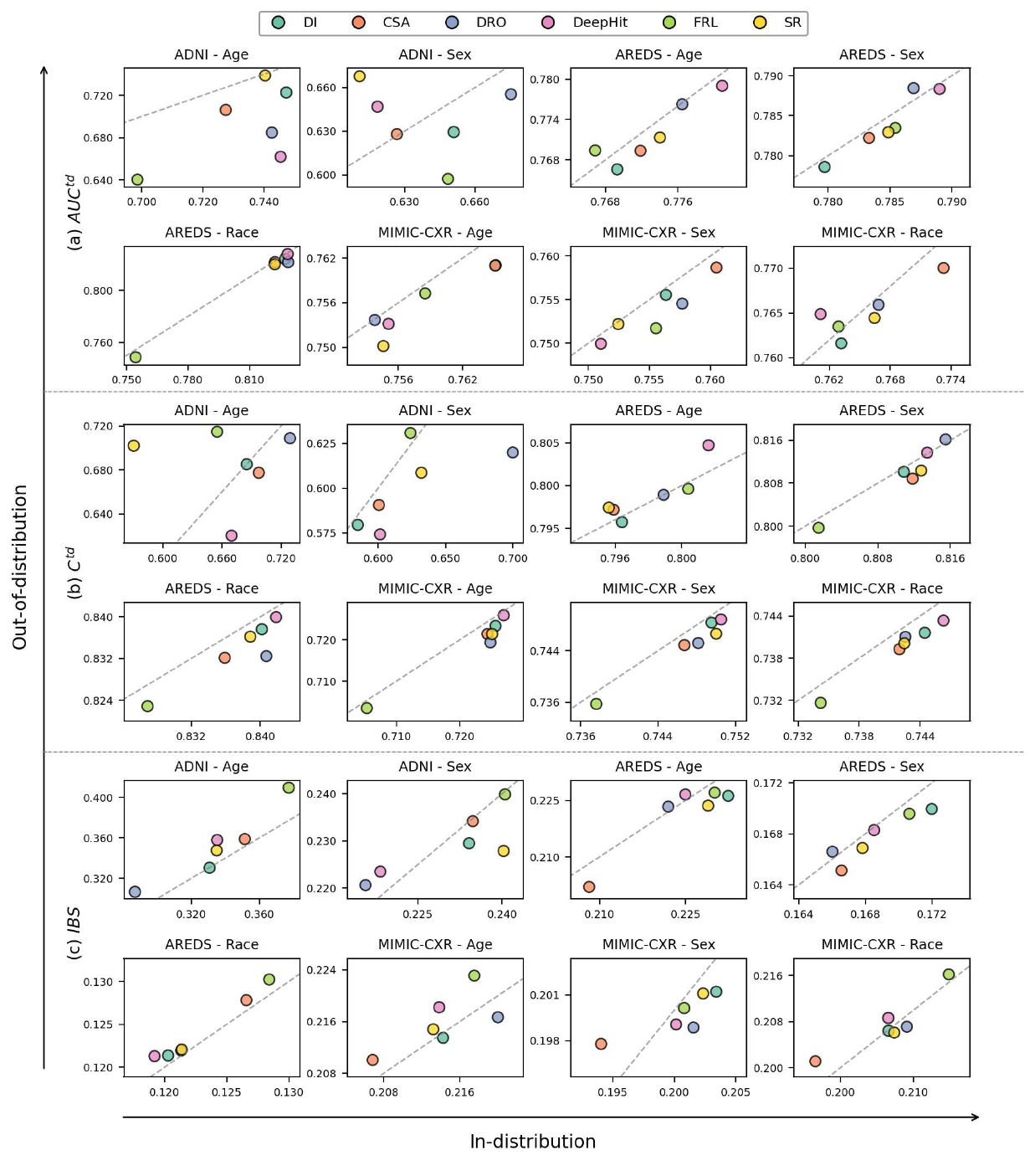}
    \end{center}
    \caption{Comparison of predictive performance on the intervened group for (fair) TTE prediction models in in-distribution vs. out-of-distribution (i.e., shift in $\Delta$) learning scenarios across all dataset and sensitive attribute combinations. The displayed results represent the average performance across all random seeds. Points on the dashed line indicate equal performance in both scenarios. a) Results for $AUC^{td}$; b) Results for $C^{td}$; c) Results for $IBS$.}
\end{figure}

\begin{figure}[H]
    \begin{center}
    \includegraphics[width=\textwidth]{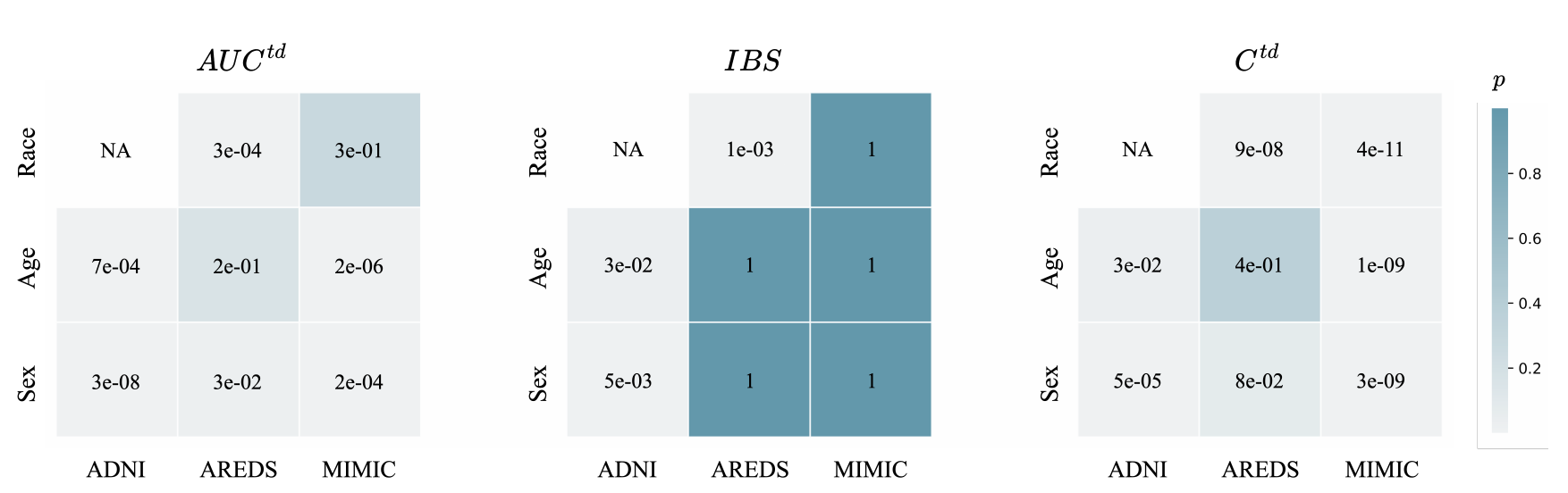}
    \end{center}
    \caption{P-values from the one-sided Wilcoxon signed-rank test computed across all fair TTE prediction models and random seeds. A p-value < 0.05 suggests distribution shift on $\Delta$ significantly degrades TTE predictive performance compared no distribution shift.}
\end{figure}

\end{document}